\documentclass[10pt,twocolumn,letterpaper]{article}

\usepackage[pagenumbers]{cvpr} % To force page numbers, e.g. for an arXiv version

\usepackage{titletoc}
\definecolor{cvprblue}{rgb}{0.21,0.49,0.74}
\usepackage[pagebackref,breaklinks,colorlinks,allcolors=cvprblue]{hyperref}

\usepackage[ruled,vlined]{algorithm2e}
\usepackage{url}            % simple URL typesetting
\usepackage{booktabs}       % professional-quality tables
\usepackage{amsfonts}       % blackboard math symbols
\usepackage{nicefrac}       % compact symbols for 1/2, etc.
\usepackage{microtype}      % microtypography
\usepackage[table]{xcolor}
\usepackage{acronym}
\usepackage{xspace}
\usepackage{amsmath}
\usepackage{graphicx}
\usepackage{enumitem}
\usepackage{tcolorbox}
\usepackage{multirow}
\usepackage{float}
\usepackage{titletoc}
\usepackage{makecell}
\tcbuselibrary{skins}

\definecolor{bestcolor}{rgb}{0.53, 0.66, 0.45}
\definecolor{secondbestcolor}{rgb}{0.63, 0.72, 0.66}
\newcommand{\bone}{\cellcolor{bestcolor}}
\newcommand{\btwo}{\cellcolor{secondbestcolor}}
\definecolor{worldgreen}{RGB}{119, 147, 61}
\definecolor{agentorange}{RGB}{228, 108, 11}
\newcommand{\world}[1]{\textcolor{worldgreen}{\textbf{#1}}}
\newcommand{\agent}[1]{\textcolor{agentorange}{\textbf{#1}}}

\usepackage[capitalize]{cleveref}
\crefname{section}{Sec.}{Secs.}
\Crefname{section}{Section}{Sections}
\Crefname{table}{Table}{Tables}
\crefname{table}{Tab.}{Tabs.}

\usepackage{pifont}
\newcommand{\cmark}{\text{\ding{51}}}
\newcommand{\xmark}{\text{\ding{55}}}

\def\paperID{} % *** Enter the Paper ID here
\def\confName{CVPR}
\def\confYear{2026}

\title{Visually-grounded Humanoid Agents}

\author{Hang Ye$^{1}$ \quad Xiaoxuan Ma$^2$ \quad Fan Lu$^3$ \quad Wayne Wu$^4$ \quad Kwan-Yee Lin$^5$ \quad Yizhou Wang$^1$ \\
$^1$Peking University \quad $^2$Carnegie Mellon University \quad  $^3$Tongji University \\
$^4$University of California, Los Angeles  \quad $^5$University of Michigan
}

\begin{document}

\twocolumn[{%
    \renewcommand\twocolumn[1][]{#1}%
    \setlength{\tabcolsep}{0.0mm} %0
    \newcommand{\sz}{0.125}  % 0.125 0.11
    \maketitle
    \vspace{-0.7cm}
    \begin{center}
        \newcommand{\teaserwidth}{\textwidth}
        \includegraphics[width=.95\linewidth]{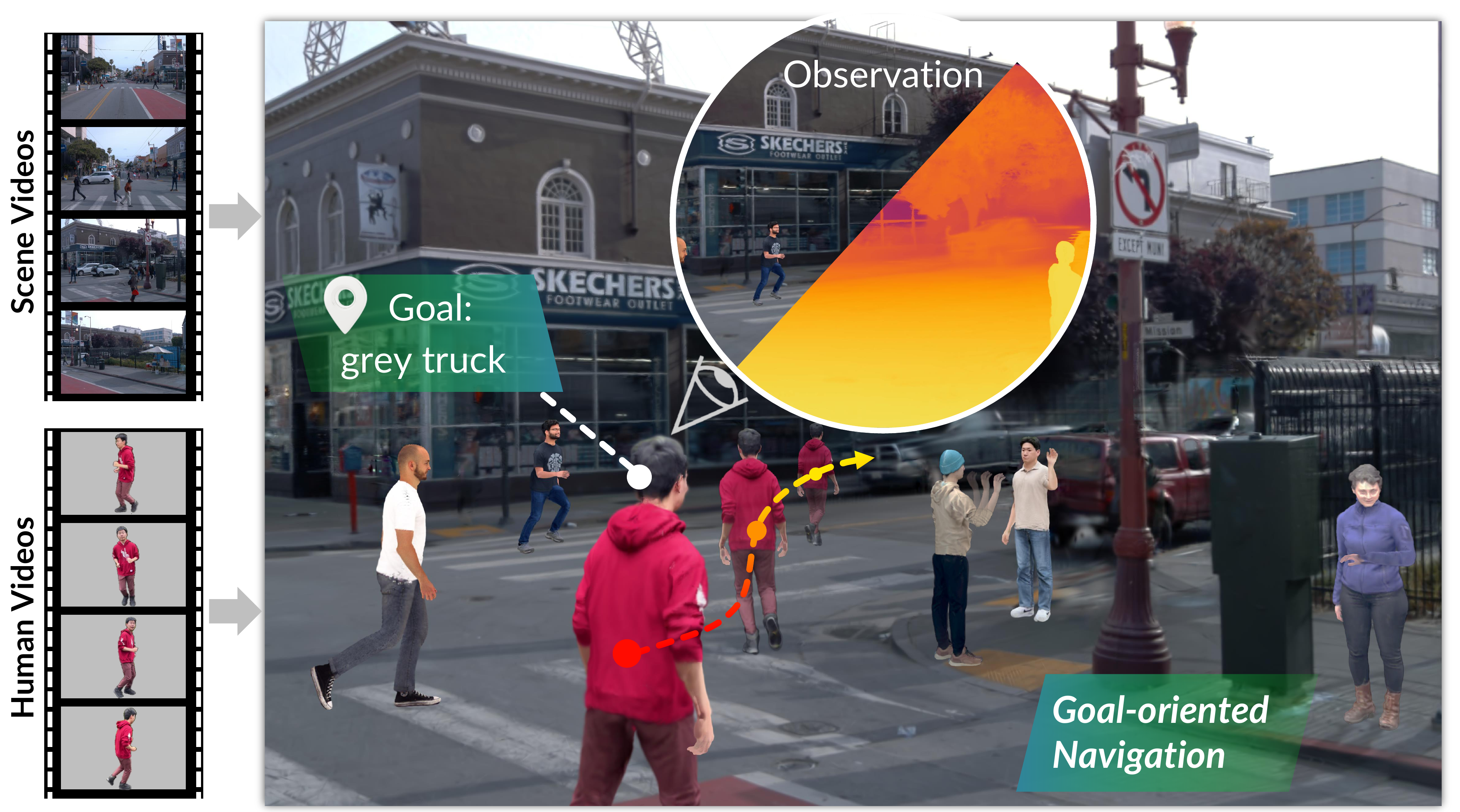}
    \vspace{-1ex}
    \captionof{figure}{
    \textbf{Visually-grounded Virtual Agents in Realistic 3D Scenes.} From monocular videos, our framework reconstructs a high-fidelity 3D environment with rich semantics and instantiates high-fidelity humanoid agents aligned with the scene. Each agent perceives the world through its own egocentric view and acts autonomously, enabling realistic and purposeful behaviors within the reconstructed environment.}
    \vspace{1ex}
    \label{fig:teaser}
    \end{center}%
}]

% misscite
% \newcommand{\misscite}{\textcolor{red}{[C]~}}

\makeatletter
\DeclareRobustCommand\onedot{\futurelet\@let@token\@onedot}
\def\@onedot{\ifx\@let@token.\else.\null\fi\xspace}
\def\eg{\emph{e.g}\onedot} 
\def\Eg{\emph{E.g}\onedot}
\def\ie{\emph{i.e}\onedot} 
\def\Ie{\emph{I.e}\onedot}
\def\cf{\emph{c.f}\onedot} 
\def\Cf{\emph{C.f}\onedot}
\def\etc{\emph{etc}\onedot} 
\def\vs{\emph{vs}\onedot}
\def\aka{a.k.a\onedot}
\def\wrt{w.r.t\onedot} 
\def\dof{d.o.f\onedot}
\def\etal{\emph{et al}\onedot}
\makeatother

% acronym
\acrodef{sota}[SOTA]{State-of-the-Art}

% notation
\newcommand{\image}{\mathbf{I}}
\newcommand{\loss}{\mathcal{L}}
\newcommand{\egorgb}{\boldsymbol{I}_{ego}}
\newcommand{\motion}{\boldsymbol{x}}
\newcommand{\control}{\boldsymbol{c}}
\newcommand{\prompt}{\boldsymbol{p}}
\newcommand{\singlemotion}{\boldsymbol{x}^{i}}%
% \enlargethispage{\baselineskip}
\begin{abstract}
Digital human generation has been studied for decades and supports a wide range of real-world applications. However, most existing systems are \textbf{passively} animated, relying on privileged state or scripted control, which limits scalability to novel environments.
We instead ask: how can digital humans \textbf{actively} behave using only \textit{visual observations} and \textit{specified goals} in novel scenes? Achieving this would enable populating any 3D environments with any digital humans, at scale, that exhibit spontaneous, natural, goal-directed behaviors.
To this end, we introduce \textbf{Visually-grounded Humanoid Agents}, a coupled two-layer (world-agent) paradigm that replicates humans at multiple levels: they \textit{look, perceive, reason, and behave} like real people in real-world 3D scenes. The World Layer provides a structured substrate for interaction, by reconstructing semantically rich 3D Gaussian scenes from real-world videos via an occlusion-aware semantic scene reconstruction pipeline, and accommodating animatable Gaussian-based human avatars within them.
The Agent Layer transforms these avatars into autonomous humanoid agents, equipping them with first-person RGB-D perception and enabling them to perform accurate, embodied planning with spatial-awareness and iterative reasoning, which is then executed at the low level as full-body actions to drive their behaviors in the scene.
We further introduce a comprehensive benchmark to evaluate humanoid–scene interaction within diverse reconstructed 3D environments. Extensive experimental results demonstrate that our agents achieve robust autonomous behavior through effective planning and action execution, yielding higher task success rates and fewer collisions compared to both ablations and state-of-the-art planning methods.
This work offers a new perspective on populating scenes with digital humans in an active manner, enabling more research opportunities for the community and fostering human-centric embodied AI. Data, code, and models will be open-sourced. Project page: \url{https://alvinyh.github.io/VGHuman/}.
\end{abstract}%

\section{Introduction} \label{sec:intro}
Digital humans have become indispensable in AR/VR~\cite{subramanyam2020comparing}, telepresence~\cite{lawrence2024project}, and robotic training~\cite{tsoi2020sean}, where they serve as avatars for interaction and simulation. However, the majority of existing models are trained from \textbf{third-person} data (\textit{e.g.,} motion capture systems~\cite{cheng2023dna} or videos~\cite{alldieck2018peoplesnapshot}) that capture appearance and kinematics but neglect the underlying \textit{decision-making context}. As a result, these systems remain {\textbf{passively controlled}}: they replay scripted motions or follow trajectory planners {\textit{without autonomy}}. Such passivity prevents digital humans from adapting to novel environments, posing a fundamental challenge for scaling up the utility of digital humans to generalize across diverse environments.

To move beyond such limitations, we argue that digital human modeling should also be framed around \textbf{active agents} that replicate humans at multiple levels: {\textbf{look, perceive, move}}, and {\textbf{reason}} like real humans in realistic 3D worlds. This requires a digital human 1) perceiving and acting from a first-person perspective using egocentric sensory input, 2) behaving adaptively under autonomous decision-making, and 3) interacting with realistic environments in a goal-directed manner.
Real humans rely on their own visual observations and short-term goal positions to make context-aware choices~\cite{warren1988direction}, such as navigating cluttered sidewalks and plazas in cities. Embedding this context-aware perception–action loop is essential for developing digital humans that generalize to novel environments and exhibit purposeful behavior. While similar principles have driven progress in robotics~\cite{anderson2018evaluation,xie2025vid2sim}, the complexity of human embodiment and animation has left these questions underexplored in the digital human domain.

With large language models (LLMs)~\cite{kaplan2020scaling} and vision-language models (VLMs)~\cite{radford2021learning}, it has become straightforward to utilize these methods to simulate aspects of digital agents such as high-level reasoning~\cite{yang2024virl} and dialogue~\cite{park2023generativeagents} in complex scenes. However, such systems largely remain disembodied: they are typically constrained to symbolic reasoning~\cite{shridhar2020alfworld} or scripted scenarios~\cite{puig2018virtualhome} and often lack visual grounding, real-world perception–action coupling, and context-aware adaptability.
Some efforts integrate VLMs with visual inputs for the motion planning of agents~\cite{cheng2024navila}, but relying solely on VLM makes it challenging to operate effectively in complex environments.
Due to these limitations, no prior work has been able to span the spectrum from semantic reasoning to embodied digital humans with perception, decision, and action fused into a continuous cycle, allowing them to adapt and act autonomously within complex, real-world 3D environments.

\begin{figure*}[t]
  \centering
  \includegraphics[width=\linewidth]{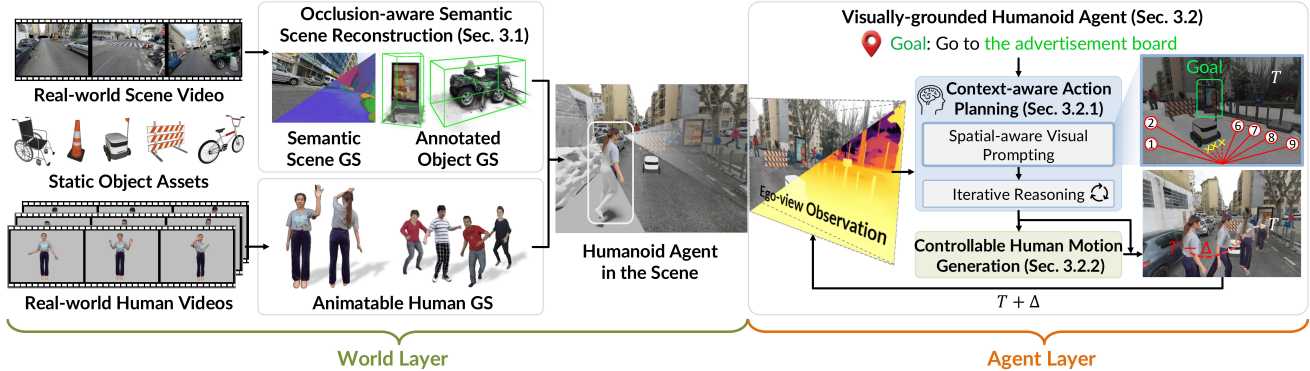}
  \vspace{-4ex}
  \caption{\textbf{Framework Overview.} Our framework consists of two layers.
  The \world{World Layer} processes real-world data (scene videos, object assets, human videos) to build large-scale, semantically detailed environments via occlusion-aware reconstruction, and populates them with GS-based animatable human avatars~(\cref{sec:scene_recon_layer}). Then the \agent{Agent Layer} drives these avatars for human-scene interaction via a perception-action loop, where visually-grounded agents plan actions from egocentric observations~(\cref{sec:method_human}).} 
  \label{fig:pipeline}
   \vspace{-0.5ex}
\end{figure*}

To this end, we introduce \textbf{Visually-grounded Humanoid Agents}, a coupled two-layer paradigm for embodied digital humans (\cref{fig:pipeline}) that requires only real-world source videos of scenes and humans.
\emph 1) The base is~\textbf{World Layer}, which reconstructs large-scale, semantically enriched, and compositional 3D environments from real-world captured videos and accommodates animatable human avatars within them. This layer provides the physical and semantic substrate, a realistic stage on which agents can be embodied.
\emph 2) On top, an \textbf{Agent Layer} equips the avatars with first-person visual perception and goal-driven planning capabilities, enabling them to act autonomously and adaptively in complex 3D environments. This layer provides a perception–action loop, coupling observation, decision, and motor control into a unified cycle. The technical methods behind the two-layer paradigm address key challenges in building embodied digital humans:
\begin{enumerate}[leftmargin=*,itemsep=2pt,topsep=0pt]
\item For the scenes in World Layer, agents require environments that are simultaneously photorealistic, semantically structured, and compositional, yet existing large-scale 3D reconstructions often suffer from occlusions and incomplete semantics. We tackle this with an \textbf{occlusion-aware semantic scene reconstruction} pipeline. It augments 3DGS~\cite{kerbl20233dgs} with semantic features~\cite{qin2024langsplat, wu2024opengaussian} via incorporating occlusion-aware masks and view selection for feature learning. This automatic workflow boosts the accuracy of 3D instance segmentation and can segment 3D instances and annotate them with semantic descriptions in large scenes, creating a semantically rich environment.
\item For the humanoids in Agent Layer, it must couple perception, planning, and action in a unified loop, while directly relying on VLMs for decision-making lacks grounding, suffers from limited memory, and is detached from action. We overcome these limitations within the agent layer through a new \textbf{spatially-aware visual prompting} and \textbf{iterative reasoning} scheme, enabling memory-enhanced, context-aware planning that remains grounded in first-person perception. Coupled with diffusion-based motion generation for realistic full-body execution, this design closes the perception–action loop. Together with the World Layer, it yields the first paradigm for embodied digital humans that can perceive, decide, and act autonomously in complex 3D environments.
\end{enumerate}

Finally, we introduce a new benchmark to evaluate how well the embodied digital humans interact with reconstructed 3D scenes. 
Extensive experiments demonstrate that humanoid agents can perform adaptive, reliable, and complex reasoning in 3D real-world environments, validating the effectiveness of our two-layer paradigm.
We believe this work will establish the foundation for the systematic, large-scale creation and evaluation of future embodied human systems.

\section{Related Work} \label{sec:related}

\textbf{Embodied Agent.} Driving embodied agents to perceive, reason, and interact with the environment has been widely studied. Most prior efforts focus on general, non-human embodiments, like quadruped robots, using either end-to-end Vision-and-Language Navigation (VLN) models~\cite{liu2024citywalker, roth2024viplanner} or VLM-based hierarchical frameworks~\cite{zhang2024navid,goetting2024vlmnav, zhang2024uninavid}.
Nevertheless, embodying humanoid agents poses substantially greater challenges due to their richer action spaces and elevated viewpoints, which complicate long-range perception and decision-making.
While existing works succeed in populating scenes with humans and achieving plausible motion, their agents either lack active scene perception and interaction~\cite{zhang2022gamma, puig2021watchandhelp, puig2023nopa, wu2024metaurban, liu2024pedgen, wei2024chatdyn, chen2024omnire} or are not true 3D embodiments~\cite{yang2024virl, park2023generativeagents}. 
Furthermore, they rely on scripted behaviors or privileged scene information, which hinders generalization to unseen real-world environments.

\noindent\textbf{3D Semantic Scene Reconstruction.} Advances in differentiable rendering, particularly NeRF~\cite{mildenhall2021nerf} and 3D Gaussian Splatting (3DGS)~\cite{kerbl20233dgs}, have revolutionized scene reconstruction. 
Recent work has scaled 3DGS to large scenes via divide-and-conquer strategies~\cite{kerbl2024hiergs, liu2024citygaussian, liu2024citygaussianv2, lin2024vastgaussian} or Level-of-Detail modeling~\cite{lu2024scaffold, ren2024octree, jiang2025horizon}, while other efforts have extended it to capture moving objects in 4D dynamic urban scenes~\cite{zhou2024drivinggaussian, lu2024drivingrecon, chen2024omnire, fischer2024dynamic, fischer2024multi}. In parallel, progress has been made in the semantic understanding of reconstructed scenes. 
LeRF~\cite{kerr2023lerf} embeds CLIP features~\cite{radford2021clip} into neural fields for text-based querying. 
ConceptGraph~\cite{gu2024conceptgraphs} integrates Grounded-SAM~\cite{ren2024groundedsam} to build 3D semantic graphs from point clouds, while more recent studies~\cite{wu2024opengaussian, qin2024langsplat, shi2024legs, zhou2024featuregs} augment 3DGS with semantic features for open-vocabulary scene understanding.
To improve efficiency, optimization-free methods~\cite{cheng2024occam, marrie2025ludvig, joseph2025gradient} directly project 2D semantic features into 3DGS, while SceneSplat~\cite{li2025scenesplat} introduces a generalizable, feed-forward model for scene understanding.
However, these methods are constrained to indoor environments or small-scale outdoor scenarios evaluated on limited benchmarks~\cite{ma2025scenesplat++}, whereas our work is the first to explore semantic understanding on large-scale outdoor 3DGS street scenes on both dataset, benchmark, and method levels. 

\vspace{0.5ex}
\noindent\textbf{Human Avatar Modeling.} 
The creation of realistic 3D human avatars relies on two fundamental pillars: photorealistic appearance and plausible motion.
On the appearance front, advancing beyond point-based representations~\cite{ma2021pop, lin2022fite, zhang2023closet, ye2025freecloth}, recent 3DGS-based methods excel at reconstructing high-fidelity humans from videos~\cite{qian20243dgs, jiang2023instantavatar, hu2024gaussianavatar, lei2024gart, moon2024exavatar, li2024animatable, xiao2025rogs} or single images~\cite{qiu2025lhm, pang2025disco4d, sim2025persona, zhuang2025idol, qiu2025anigs}.
A concurrent work~\cite{mir2025aha} reconstructs scenes with animatable 3D humans via a unified 3DGS representation, but limited to fitting pre-defined trajectory and motions.
On the motion front, prior studies employ diffusion models to synthesize human motions~\cite{tevet2022mdm,chen2023mld}. To achieve scene-aware motion generation, some approaches incorporate scene context as a conditioning signal~\cite{jiang2024scaling, hassan2021samp,mir2024generating,zhang2024scenic,zhao2022coins,cong2024laserhuman}, or learn an RL-based policy~\cite{zhao2023dimos, zhang2022gamma}. However, it is difficult to extend this paradigm to outdoor human–scene interactions due to the scarcity of paired datasets.

\begin{figure*}[t]
  \centering
  \includegraphics[width=\linewidth]{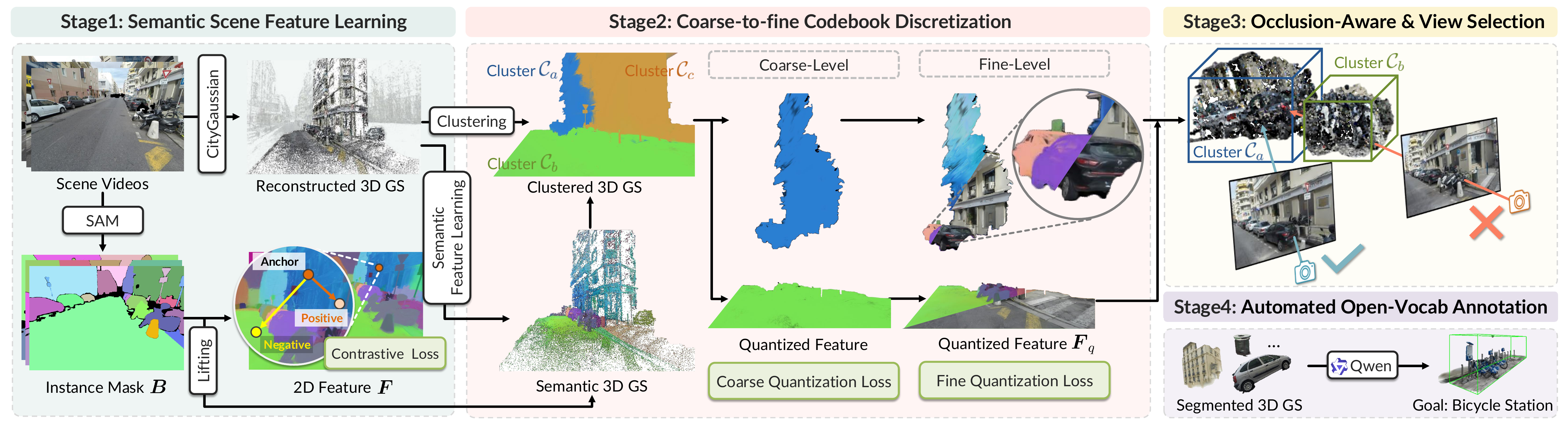}
  \vspace{-4ex}
  \caption{\textbf{Overview of the Occlusion-Aware Semantic Scene Reconstruction.} 
  We first reconstruct 3D Gaussians from scene videos utilizing CityGaussian~\cite{liu2024citygaussian, liu2024citygaussianv2}. To augment 3DGS with instance-level semantics, we extract 2D masks $\boldsymbol{B}$ based on SAM~\cite{kirillov2023sam}, lift them to 3D via contrastive learning, and then segment 3D instances using coarse-to-fine quantization. 
  We introduce occlusion-aware masks and view selection to boost segmentation accuracy in large-scale, occluded outdoor scenes. Finally, each instance is annotated via context-aware visual prompting with a VLM~\cite{bai2025qwen2}, yielding a semantically rich environment with spatially annotated landmarks, ready for human-scene interaction. 
  } 
  \vspace{-1.6ex}
  \label{fig:scene_recon}
\end{figure*}

\section{Method} \label{sec:method}

As illustrated in~\cref{fig:pipeline}, our framework consists of two layers. Given real-world input sources (scene videos, object assets, and human videos), the \world{World Layer} (\cref{sec:scene_recon_layer}) reconstructs: 1) large-scale, semantically enriched environments via a {\textit{occlusion-aware semantic scene reconstruction}} pipeline (\cref{sec:scene_recon}), which augments the 3DGS scenes with robust semantic annotations and spatial landmarks; and 2) \textit{animatable Gaussian-based human avatars}~(\cref{sec:scene_recon_avatar}), which are randomly placed into the reconstructed scenes to enable scalable human–scene interaction. On top, the \agent{Agent Layer} (\cref{sec:method_human}) governs these avatars through a perception–action loop: First, a \textit{high-level, context-aware planning} (\cref{sec:high_level_vlm}) utilizes a VLM in a zero-shot manner to interpret first-person observations through spatial visual prompting and iterative reasoning, thereby mitigating its limitations in visual grounding and memory, and formulating context-aware goals. Then a \textit{low-level, diffusion-based motion generation}  (\cref{sec:motion_gen}) translates these abstract goals into realistic full-body behavior. The generated motion is executed in the environment and fed back into the agent’s perception, closing the \textbf{perception–action loop} for continuous interaction.

\begin{figure*}[t]
  \centering
  \includegraphics[width=1\linewidth]{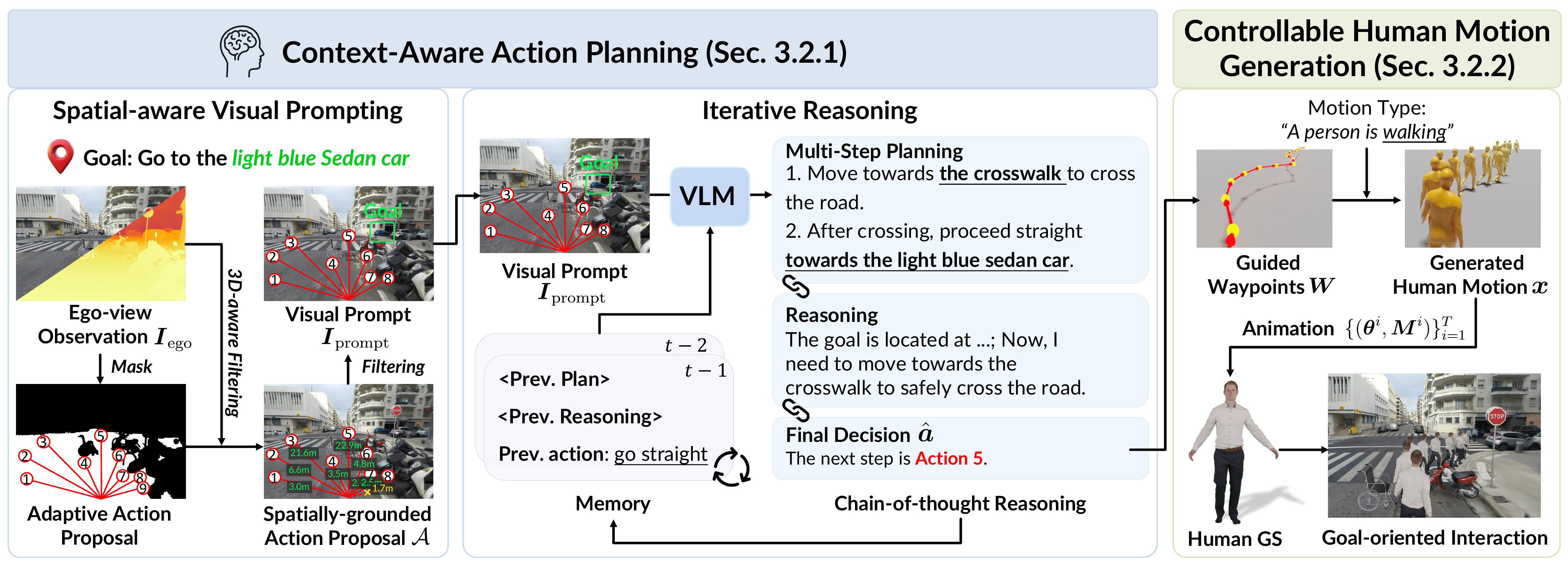}
  \vspace{-4.2ex}
  \caption{Our \textbf{Visually Grounded Humanoid Agent} comprises a two-level framework:
    (1) A \textbf{context-aware action planning module} (high-level planner) that predicts actions from ego-centric observations. It utilizes spatial-aware visual prompting to generate physically viable, spatially grounded proposals and apply goal highlighting for contextual cues, combined with iterative reasoning for multi-step decision making~(\cref{sec:high_level_vlm}).
    (2) A \textbf{controllable motion generation module} (low-level controller) that converts the planner's command into waypoints, which then condition a motion diffusion model to synthesize full-body motion~(\cref{sec:motion_gen}).}
\vspace{-1.5ex}
  \label{fig:human}
\end{figure*}

\subsection{The Base - World Layer} \label{sec:scene_recon_layer}

To enable autonomous humanoid agents to interact within open-worlds, an ideal reconstructed environment for benchmarking embodied interaction should integrate three core components: \textbf{1)}  photorealistic rendering under unconstrained camera trajectories as agents can move arbitrarily;
\textbf{2)} physical meshes for collision detection, and \textbf{3)} object-level semantics for high-level reasoning. 
While actively explored in indoor domains~\cite{miao2025sage3d}, there is a critical absence of such environments for large-scale outdoor scenes. 
Existing large-scale city datasets~\cite{jiang2025horizon, kerbl2024hiergs, zhang2024ucgs, li2023matrixcity} lack ready-to-use semantic annotations or robust segmentation methods. Meanwhile, autonomous driving datasets~\cite{wilson2023argoverse, xiao2021pandaset, sun2020waymo} provide manually annotated, vehicle-centric 3D bounding boxes, failing to support diverse human-scene interactions.

To construct our interactive environment, we build upon a 2DGS-based~\cite{huang20242dgs} representation~\cite{liu2024citygaussian, liu2024citygaussianv2} to reconstruct photorealistic city scenes and extract collision meshes, as detailed in \cref{supp:scene-recon}. 
We then focus on endowing the digital twin with detailed object-level semantics to support goal-directed agent interaction.
Scaling general object-centric or in-door level semantic GS to vast, complex urban environments presents two major challenges: 1) \textit{Severe occlusion}, where objects are frequently hidden by others (e.g., a car blocking a fire hydrant), leading to unreliable semantic feature propagation for occluded regions; and 2) \textit{Sparse supervisory signals}, where open-world landmarks are often visible in only a few training views, resulting in weak feature learning and poor generalization across the scene. 
To address these issues, we introduce an automated, occlusion-aware semantic scene reconstruction pipeline, as shown in~\cref{fig:scene_recon}.

\noindent\textbf{Occlusion-Aware Semantic Scene Reconstruction.}
\label{sec:scene_recon}
Given reconstructed 3D scene Gaussians $\mathcal{G}=\{\mathcal{G}_i\}_{i=1}^N$, we aim to augment each $\mathcal{G}_i$ with a learnable feature vector $\boldsymbol{f}_i \in \mathbb{R}^C$. The core idea is an association between 2D and 3D (\cref{fig:scene_recon}, Stage 1): we first obtain accurate 2D instance segmentations from multi-view images, lift these masks into a consistent 3D feature space, and subsequently project the 3D features back into 2D to refine segmentation to optimize its feature. This process tackles the issues from four aspects: ~\textit{instance discrimination},~\textit{instance-level consistency}, ~\textit{occlusion handling}, and~\textit{open-vocabulary annotation}.

\noindent\textbf{Semantic Scene Feature Learning.} 
We leverage 2D instance masks from SAM~\cite{kirillov2023sam} to guide feature learning. For an arbitrary training view, we render per-Gaussian features $\boldsymbol{f}=\{\boldsymbol{f}_i\}_{i=1}^N$ into a 2D feature map $\boldsymbol{F} \in \mathbb{R}^{H\times W \times C}$. Given $K$ binary instance masks $\boldsymbol{B}=\{\boldsymbol{B}_{k}\}_{k=1}^K$, we enforce feature consistency via contrastive learning: an intra-mask smoothing loss aligns pixel-wise features $\boldsymbol{F}(p)$ with their mean feature $\bar{\boldsymbol{F}}_k$: $\mathcal{L}_s = \sum_{k=1}^K \sum_{p \in \boldsymbol{B}_k} \|\boldsymbol{F}(p) - \bar{\boldsymbol{F}}_k\|^2, \quad \text{where} \quad \bar{\boldsymbol{F}}_k = \frac{\sum_{p \in \boldsymbol{B}_k} \boldsymbol{F}(p)}{|\boldsymbol{B}_k|}$. An inter-mask contrastive loss separates the mean features of different instances to enhance their distinctiveness: $\mathcal{L}_c = \frac{1}{K(K-1)} \sum_{k=1}^K \sum_{j \ne k}^K \frac{1}{\|\bar{\boldsymbol{F}}_k - \bar{\boldsymbol{F}}_j\|^2}$,
where $\boldsymbol{F}(p)$ is the feature vector at the pixel $p$.
This process produces view-consistent and distinct instance features directly from 2D supervision.

\noindent\textbf{Coarse-to-fine Codebook Discretization.}
To ensure all Gaussians belonging to same object share identical instance features, we use a coarse-to-fine codebook discretization strategy.  For coarse level, we cluster the Gaussians using their learned instance features $\boldsymbol{f}$ and 3D coordinates $\boldsymbol{x}$. Within each coarse cluster, we perform fine-level quantization using only instance features $\boldsymbol{f}$. This step identifies distinct object instances within each geometric chunk (\cref{fig:scene_recon}, Stage 2). To train the model to produce these discrete features, we optimize the quantization loss $
\loss_q = \|\boldsymbol{F}-\boldsymbol{F}_q\|_1$, enforcing feature map $\boldsymbol{F} \in \mathbb{R}^{H\times W\times C}$, rendered from the original continuous features learned in Stage 1, matches the feature map $\boldsymbol{F}_q$, rendered from the newly quantized features.

\noindent\textbf{Occlusion-Aware Masks and View Selection.} 
Nevertheless, naively training a coarse cluster by only rendering its Gaussians neglects occlusions from other scene objects, leading to fragmented segmentation. For example, if cluster $\mathcal{C}_b$ occludes cluster $\mathcal{C}_a$ from a viewpoint (\cref{fig:scene_recon}, Stage 3), semantic features can propagate erroneously. To resolve this, we introduce two mechanisms:
{\textbf{1)} \textit{Occlusion-Aware Cluster Masks}: For each target cluster, we render all other clusters as depth-only occluders. This generates an occlusion-aware, 2D binary mask $\boldsymbol{\hat{B}}_c$ that isolates the visible regions of the target cluster for a given camera. The quantization loss $\loss_q$ is applied only within this mask, preventing feature contamination from occluding objects.
{\textbf{2)}} \textit{Strategic View Selection}: To address the sparse supervision issue, we pre-compute an occlusion-aware visibility score for each cluster-view pair. A coarse cluster is then trained only on views where its visible pixel count exceeds a threshold $\delta$. This focuses optimization on high-quality views, accelerating convergence and improving accuracy.
This dual strategy ensures that feature updates are both accurate and efficient, by preventing contamination from occluders, and enables fine-grained 3D segmentation, as demonstrated in our ablation studies~(see \cref{sec:ablation}). 

\definecolor{tablegreen}{RGB}{26,127,55}
\definecolor{tableorange}{RGB}{203,120,28}
\definecolor{tablered}{RGB}{180,44,44}
\newcommand{\lvlultra}[1]{\textcolor{black}{#1}}
\newcommand{\lvlhigh}[1]{\textcolor{black}{#1}}
\newcommand{\lvlmid}[1]{\textcolor{black}{#1}}
\newcommand{\lvllow}[1]{\textcolor{black}{#1}}
\newcommand{\yescol}{\textcolor{black}{\cmark}}
\newcommand{\nocol}{\textcolor{black}{\xmark}}

\begin{table*}[t]
\centering
\caption{\textbf{Settings of the scenes used in experiments.} We summarize the roles of experiment scenes in world-layer and agent-layer evaluation together with key scene properties.}
\vspace{-1.8ex}
\label{tab:dataset_comparison}
\setlength{\tabcolsep}{8pt}
\resizebox{.85\linewidth}{!}{
\begin{tabular}{l@{\hspace{1pt}}|ccc|ccc|c}
\toprule
\textbf{Scene} & \makecell[c]{\textbf{World}\\\textbf{Eval.}} & \makecell[c]{\textbf{Agent}\\\textbf{Eval.}} &
\makecell[c]{\textbf{Universality}\\\textbf{Eval.}} & \makecell[c]{\textbf{Recon.}\\\textbf{Quality}} & \makecell[c]{\textbf{Inter.}\\\textbf{Range}} & \makecell[c]{\textbf{Sem.}\\\textbf{Richness}} & \makecell[c]{\textbf{Coll.}\\\textbf{Mesh}} \\
\midrule
SmallCity~\cite{kerbl2024hiergs} & \cmark & \cmark &\cmark & High & Massive & High & \yescol \\
XGRIDS~\cite{xgrids} &  & \cmark & \cmark & Ultra & Massive & High & \yescol \\
SAGE-3D~\cite{miao2025sage3d} &  & \cmark &\cmark & High & Room-scale & Medium & \yescol \\
ArgoVerse2~\cite{wilson2023argoverse}, PandaSet~\cite{xiao2021pandaset} &  &  & \cmark & Medium & Large & Vehicle-only & \yescol \\
Mip-NeRF360~\cite{barron2021mip}, DL3DV-10K~\cite{ling2024dl3dv} &  &  &\cmark & Medium & Limited & Low/Medium & \yescol \\
MatrixCity~\cite{li2023matrixcity}, Horizon-GS~\cite{jiang2025horizon} &  &  & \cmark & Low & Massive & High & \yescol \\
SuperSplat~\cite{supersplat}, Pointcosm~\cite{pointcosm} &  &  & \cmark & Ultra & Large & Medium & \nocol \\
\bottomrule
\end{tabular}
\vspace{-1.5ex}
}
\end{table*}

\noindent\textbf{Automated Open-Vocabulary Annotation.} 
Following 3D semantic segmentation, we automatically annotate each instance using Qwen2.5-VL~\cite{bai2025qwen2} via a visual prompting mechanism, yielding richer descriptions (\cref{fig:scene_recon}, Stage 4; see Supp. for details).
Now the resulting photorealistic, semantically-rich scene directly supports 3D bounding box extraction for tasks like goal-based navigation in our benchmark.

\noindent\textbf{Avatar Reconstruction and Agent Placement.} \label{sec:scene_recon_avatar}
Given real-world human videos, we reconstruct $M$ animatable human avatars, each modeled as a collection of Gaussian primitives~\cite{hu2024gaussianavatar, moon2024exavatar}. Specifically, the $j$-th human (where $j \in {1, \dots, M}$) at physical timestep $t$ is represented by a set of $N_h$ Gaussians:
$\mathcal{G}^{h_j}(t) = \{\mathcal{G}^{h_j}_k(t)\}_{k=1}^{N_h},$
where $\mathcal{G}^{h_j}_k(t)$ denotes the $k$-th Gaussian primitive of the $j$-th human at time $t$. The collection of all dynamic human Gaussians in the scene at time $t$ can thus be written as:
$\mathcal{G}^h(t) = \bigcup_{j=1}^{M} \mathcal{G}^{h_j}(t).$ These avatars are randomly placed into the reconstructed scene to populate it with diverse human–scene interactions. Their behaviors are then driven by the \textbf{Agent Layer}.

\subsection{The Agent Layer - Visually Grounded Humanoid Agents} \label{sec:method_human}

To enable {\textit{autonomous}} human-scene interaction, we design our agent based on a two-level framework that mimics the cognitive paradigm of ``slow thinking" for planning and ``fast execution" for human motion control, as shown in~\cref{fig:human}. In the following, we detail the design of each component.

\vspace{0.5ex}

\noindent\textbf{Context-Aware Action Planning.} \label{sec:high_level_vlm}
 For the high-level planner, we leverage the semantic reasoning ability of VLMs in a training-free, zero-shot way, but constrain the inputs, reasoning chain, and outputs with spatial visual prompting and iterative refinement. This grounds planning in first-view perception and aligns it with downstream low-level execution.  Thus, we cast high-level planning as a selection problem over discrete human-centric action primitives~\cite{nasiriany2024pivot}, avoiding the inherent limitations of VLMs~\cite{rahmanzadehgervi2024vision} in continuous control regression and spatial reasoning.

\vspace{0.5ex}

\noindent\textbf{Action Selection via Visual Prompting.}
At each decision step, the agent receives an ego-centric RGB-D observation $\boldsymbol{I}_{\text{ego}} \in \mathbb{R}^{H \times W \times 4}$ and a high-level task description $\mathcal{T}$. We define a discrete action space $\mathcal{A}$ composed of $J$ action primitives: $\mathcal{A} = \{\boldsymbol{a}_1, \boldsymbol{a}_2, \dots, \boldsymbol{a}_J\}$. Each primitive $\boldsymbol{a}_j = (\tau_j, \boldsymbol{d}_j)$ is a tuple containing: 1) $\tau_j$: A \textbf{motion type} prompt, \eg, ``walk'' or ``run''; 2) $\boldsymbol{d}_j$: A \textbf{canonical direction vector} in the 2D image plane, indicating a potential next-move  direction. The VLM is then prompted with the task $\mathcal{T}$ and the visual context $\boldsymbol{I}_{\text{prompt}}$ to select the optimal action primitive $\hat{\boldsymbol{a}}$. 
Simply overlaying fixed directional arrows onto the ego-centric view ignores 3D geometry and traversability, leading the VLM to choose physically impossible or unsafe paths (see ablation in~\cref{sec:exp}). This challenge is magnified in large-scale outdoor scenes, where distant goals and cluttered, occluding objects demand robust long-horizon planning (\cref{fig:scene_recon}).

To this end, we introduce a pipeline with two key innovations:  \textbf{1)} Spatial-Aware Visual Prompting, which {\textit{discretizes}} and {\textit{emphasizes}} candidate directions and grounds them with spatial, semantic, and goal context, giving the VLM a spatial-aware visual input $\boldsymbol{I}_{\text{prompt}}$ for selecting a physically plausible one-step action; and \textbf{2)}~Iterative Reasoning, which {\textit{continues}} planning across multiple steps by maintaining memory and reasoning chains, allowing robust long-horizon navigation, obstacle avoidance, and handling target occlusion.

\noindent\textbf{Spatial-Aware Visual Prompting.} 
Rather than laborious text prompt engineering, we enrich the visual input with contextual cues for spatial understanding: 
\textbf{1)} We generate physically viable, \textit{spatially grounded action proposals} by segmenting the traversable ground with SAM~\cite{kirillov2023sam} and sampling 2D candidate paths~${\boldsymbol{d}_k}$ on it. These paths are then back-projected into 3D points~${\boldsymbol{p}_k}$ using depth maps and pruned if shorter than a safety threshold~$\delta$ (\cref{fig:human}).
\textbf{2)} To facilitate complex reasoning in large-scale outdoor scenes, we leverage open-vocabulary localization: the VLM first highlights the target object with a bounding box, reducing complex planning to ``navigating toward the highlighted object", decreasing ambiguity and clarifying agent-goal relations.
Collectively, these techniques inject physical and relational context into the visual input, significantly enhancing the VLM's spatial reasoning capabilities (see Fig~\ref{fig:qual}).

\vspace{0.2ex}

\noindent\textbf{Iterative Reasoning.} While spatially-aware visual prompting enhances {\textit{single-step}} decision-making, it can encourage myopic strategies that are suboptimal for {\textit{long-horizon}} goals (see~\cref{fig:qual}). To endow the agent with foresight, we introduce an iterative reasoning mechanism inspired by Chain-of-Thought (CoT) prompting~\cite{wei2022cot}, adapted to incorporate visual information.
At each timestep $t$, the text prompt is dynamically updated to include: 1) the high-level language goal, 2) a memory buffer with the plan, reasoning, and actions from previous steps $t-1, t-2, \etc$, and 3) a query for the VLM to update its plan and select the next action based on the current observation. As shown in ~\cref{fig:human}, the VLM outputs an updated natural language plan and the action index $\hat{k}$. This iterative loop enables the agent to balance long-term strategy with immediate visual feedback, reducing collisions by integrating foresight into embodied decision-making. 

\vspace{0.2ex}

\noindent\textbf{Controllable Human Motion Generation.}  \label{sec:motion_gen}
Given a high-level action primitive $\hat{\boldsymbol{a}} = (\hat{\tau}, \hat{\boldsymbol{d}})$ produced by the planner, the low-level controller constructs a sequence of $N_w$ waypoints $\boldsymbol{W} = \{\boldsymbol{w}_1, \dots, \boldsymbol{w}_{N_w}\}$ by linearly interpolating between the agent's current position and the 3D target point $\hat{\boldsymbol{p}}$ in the chosen direction. These waypoints and the motion type text $\hat{\tau}$ (\eg, ``walk") then condition a motion diffusion model~\cite{petrovich2024stmc} to synthesize a plausible full-body motion sequence $\boldsymbol{x}=\{\boldsymbol{x}^i\}_{i=1}^T$, where local poses are factored out from global transformation $\boldsymbol{M}=\{\boldsymbol{M}^i\}_{i=1}^T$.
To steer generation at inference time, we incorporate a training-free guidance mechanism~\cite{liu2024programmable, karunratanakul2024dno}: $\tilde{\boldsymbol{x}}_k = \boldsymbol{x}_k - \alpha \nabla_{\boldsymbol{x}_k} \mathcal{L}(\boldsymbol{x}_k; \boldsymbol{W}, \hat{\tau})$, where $\alpha$ is the guidance scale and $\mathcal{L}$ enforces waypoint adherence, text consistency, and temporal smoothness. The final motion is converted to SMPL parameters $\{(\boldsymbol{\theta}^i, \boldsymbol{M}^i)\}_{i=1}^T$ for animation; full details are provided in \cref{supp:motion-gen}.

\begin{table*}[t]
\centering
\caption{\textbf{Quantitative comparison on \textit{SmallCity} dataset using our navigation benchmarks.} Our method significantly outperforms three SOTA VLN baselines, improving task success rates by roughly $30\%$ across all tasks. Top two results are highlighted by \colorbox{bestcolor}{first} and \colorbox{secondbestcolor}{second}.}
\vspace{-5pt}
\label{tab:main}
\begin{tabular}{@{}l|ccc|ccc|ccc@{}}
\toprule
\textbf{Task} & \multicolumn{3}{c|}{\textbf{SimNav}} & \multicolumn{3}{c|}{\textbf{ObstNav}} & \multicolumn{3}{c}{\textbf{SocialNav}} \\ \midrule
Method & SR~$\uparrow$ & SPL~$\uparrow$ & CR~$\downarrow$ & SR~$\uparrow$ & SPL~$\uparrow$ & CR~$\downarrow$ & SR~$\uparrow$ & SPL~$\uparrow$ & CR~$\downarrow$ \\ \midrule
NaVILA~\cite{cheng2024navila} & 22.5\% & 0.199 & 70.8\% & 20.8\% & 0.176 & 75.0\% & 8.3\% & 0.072 & 84.1\% \\
NaVid~\cite{zhang2024navid} & 37.4\% & 0.279 & \btwo 19.2\% & \btwo 32.5\% & 0.233 & \bone 23.1\% & \btwo 17.5\% & \btwo 0.138 & \btwo 51.7\% \\
Uni-NaVid~\cite{zhang2024uninavid} & \btwo 38.8\% & \btwo 0.370 & \bone 12.8\% & 25.3\% & \btwo 0.242 & \btwo 29.4\% & 12.5\% & 0.112 & 66.7\% \\ \midrule
\textbf{Ours} & \bone 68.3\% & \bone 0.640 & \btwo 13.3\% & \bone 55.8\% & \bone 0.516 & 30.8\% & \bone 39.2\% & \bone 0.366 & \bone 48.3\% \\
\bottomrule
\end{tabular}
\vspace{-3pt}
\end{table*}

\begin{table*}[t]
\centering
\caption{\textbf{Generalization on diverse environments and multi-goal evaluation.} We evaluate on two outdoor scenes from XGRIDS~\cite{xgrids}, four indoor scenes from SAGE-3D~\cite{miao2025sage3d}, and a multi-goal benchmark with 2--5 consecutive landmarks per episode.}
\vspace{-1.5ex}
\label{tab:main2}
\setlength{\tabcolsep}{3pt}
\begin{tabular}{@{}l|ccc|ccc|cccc@{}}
\toprule
\textbf{Benchmark} & \multicolumn{3}{c|}{\textbf{XGRIDS}} & \multicolumn{3}{c|}{\textbf{SAGE-3D}} & \multicolumn{4}{c}{\textbf{Multi-Goal}} \\ \midrule
Method & SR~$\uparrow$ & SPL~$\uparrow$ & CR~$\downarrow$ & SR~$\uparrow$ & SPL~$\uparrow$ & CR~$\downarrow$ & SR~$\uparrow$ & PR~$\uparrow$ & PPL~$\uparrow$ & CR~$\downarrow$ \\ \midrule
NaVILA~\cite{cheng2024navila} & 17.8\% & 0.160 & 71.1\% & 31.7\% & 0.293  & 67.2\% & 0.0\% & \btwo 21.6\% & \btwo 0.102 & 78.9$\%$  \\
NaVid~\cite{zhang2024navid} & \btwo 40.0\% & \btwo 0.316 & \btwo 35.6\% & \btwo 33.1\% & \btwo 0.324 & 58.4\% & 0.0\% & 17.5\% & 0.096 & 70.2$\%$ \\
Uni-NaVid~\cite{zhang2024uninavid} & 31.1\% & 0.298 & 48.9\% & 25.1\% & 0.228 & \btwo 35.2\% & \btwo 12.3$\%$ & 8.3$\%$ & 0.066 & \btwo 48.9$\%$ \\ \midrule
\textbf{Ours} & \bone 74.7\% & \bone 0.725 & \bone 14.7\% & \bone 58.2\% & \bone 0.534 & \bone 33.3\% & \bone 38.0$\%$ & \bone 73.6$\%$ & \bone 0.707 & \bone 34.8$\%$ \\
\bottomrule
\end{tabular}
\vspace{-0.5cm}
\end{table*}

\section{Experiments} \label{sec:exp}

\subsection{Dataset and Environment}

\noindent\textbf{Scenes.} An effective benchmark for embodied human interaction should provide high rendering quality, sufficient interaction range for long-horizon tasks, rich scene semantics, and collision meshes for physical interaction. Based on these criteria, we curate a diverse hierarchy of scenes (see~\cref{tab:dataset_comparison}), and details of per-scene preprocessing and annotation are provided in~\cref{supp:scene-select}.

\noindent\textbf{Avatars.} To populate the environment, we reconstruct 14 animatable GS-based avatars from monocular videos in PeopleSnapshot~\cite{alldieck2018peoplesnapshot}, GaussianAvatar~\cite{hu2024gaussianavatar}, NeuMan~\cite{jiang2022neuman}, and X-Humans~\cite{shen2023xhumans} datasets, covering diverse appearances, body shapes, and ethnicities. These avatars are used to instantiated humanoid agents across all navigation tasks.

\subsection{Benchmark Design}

\noindent\textbf{Task Design.}
We focus our benchmark on navigation, where comparison is most meaningful to demonstrate autonomy. 
Using semantic goals extracted from the World Layer or preprocessed annotations, we design: 1) three single-goal visual navigation tasks on \textit{SmallCity} with increasing complexity:
\textbf{Level-1 (SimNav)} tests the agent's fundamental goal-reaching ability with a clear path to the goal.
\textbf{Level-2 (ObstNav)} introduces static obstacles, requiring adaptive planning to balance navigation and collision avoidance.
\textbf{Level-3 (SocialNav)} adds dynamic complexity, where the agent must avoid collisions with moving humanoid agents through reactive and socially-aware planning. 2) Beyond single-goal evaluation, we construct a multi-goal benchmark comprising 2--5 consecutive landmarks per episode on \textit{SmallCity}, assessing {\textbf{long-horizon planning}}. 
3) We further evaluate generalization via basic navigation on XGRIDS~\cite{xgrids} and SAGE-3D~\cite{miao2025sage3d} without augmented obstacles or dynamic agents, with 60--200 testing cases per dataset.

\noindent\textbf{Motion Versatility.}
To focus on high-level planning rather than motion patterns, the agent's motion type is fixed to ``walking" during evaluation if not specified.
Nonetheless, our framework readily extends to diverse locomotion styles, and we additionally evaluate navigation performance across these styles in \cref{sec:exp_agent}.
Beyond locomotion, the framework also supports event-triggered actions, social engagement, and scene-dependent interactions (see \cref{fig:robustness_gallery} and \cref{supp:diverse-motion}).

\noindent\textbf{Baselines and Metrics.}
(i)~\textit{Agent layer:} We compare against three state-of-the-art VLN approaches: NaVILA~\cite{cheng2024navila}, NaVid~\cite{zhang2024navid}, and Uni-NaVid~\cite{zhang2024uninavid}. To ensure a fair comparison of high-level planning, we intercept their mid-level language commands and convert them into waypoints for our motion generation model, with VLM query frequency kept consistent across all methods.
We report the success rate (SR)~\cite{anderson2018evaluation}, success weighted by path length (SPL)~\cite{batra2020objectnav}, and collision rate (CR), averaged over three runs.
For the multi-goal benchmark, we additionally adopt the progress rate (PR) and progress weighted by path length (PPL) from MultiON~\cite{wani2020multion}.
(ii)~\textit{World layer:} We compare against three representative semantic GS methods: Feature-3DGS~\cite{zhou2024featuregs}, Gradient-Weighted 3DGS~\cite{joseph2025gradient}, and OpenGaussian~\cite{wu2024opengaussian}.
Since no ground-truth semantic annotations exist for \textit{SmallCity}, we manually curate 240 evaluation samples consisting of accurate 2D SAM masks paired with annotated text, and report the mean IoU (mIoU) and mean accuracy (mAcc).

\begin{table*}[h]
\centering
\caption{\textbf{Ablation study of the VLM-based planning paradigm.} VP = Spatial-Aware Visual Prompting; IR = Iterative Reasoning. VP enhances visual cues for goal localization, while IR prevents short-sighted decision-making. The top two results are highlighted.}
\vspace{-1.5ex}
\label{tab:ablation}
\begin{tabular}{@{}c@{\hspace{0.5em}}cc|ccc|ccc|ccc@{}}
\toprule
 & \multicolumn{2}{c|}{\textbf{Method}} & \multicolumn{3}{c|}{\textbf{SimNav}} & \multicolumn{3}{c|}{\textbf{ObstNav}} & \multicolumn{3}{c}{\textbf{SocialNav}} \\
\cmidrule{2-12}
 & VP & IR & SR~$\uparrow$ & SPL~$\uparrow$ & CR~$\downarrow$ & SR~$\uparrow$ & SPL~$\uparrow$ & CR~$\downarrow$ & SR~$\uparrow$ & SPL~$\uparrow$ & CR~$\downarrow$ \\
\midrule
(a) & \xmark & \xmark & 53.3$\%$ & \btwo 0.519 & 46.7$\%$ & 38.3$\%$ & 0.365 & 60.0$\%$ & 30.3$\%$ & \btwo 0.287 & 69.1$\%$ \\
(b) & \cmark & \xmark & \btwo 57.5$\%$ & 0.501 & 32.5$\%$ & \btwo 46.7$\%$ & \btwo 0.395 & 47.5$\%$ & 25.8$\%$ & 0.247 & 59.2$\%$ \\
(c) & \xmark & \cmark & 49.2$\%$ & 0.445 & \bone 7.6$\%$ & 44.2$\%$ & 0.387 & \bone 17.7$\%$ & \btwo 31.7$\%$ & 0.274 & \btwo 51.8$\%$ \\
(d) & \cmark & \cmark & \bone 68.3$\%$ & \bone 0.640 & \btwo 13.3$\%$ & \bone 55.8$\%$ & \bone 0.516 & \btwo 30.8$\%$ & \bone 39.2$\%$ & \bone 0.366 & \bone 48.3$\%$ \\
\bottomrule
\end{tabular}
\vspace{-0.5ex}
\end{table*}

\begin{figure*}[t]
  \centering
  \includegraphics[width=0.98\linewidth]{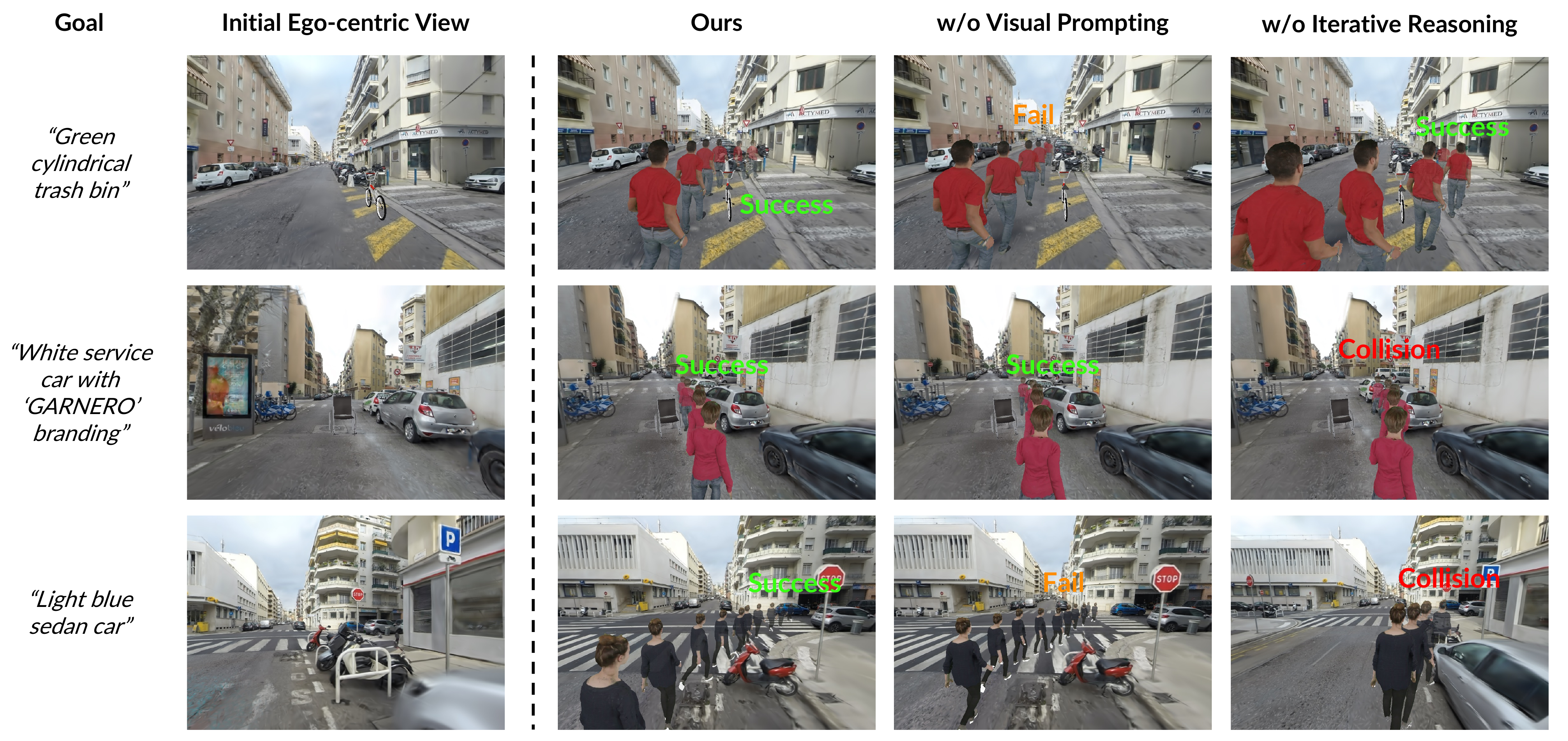}
  \vspace{-1.5ex}
  \caption{\textbf{Qualitative ablation of the VLM-based planning paradigm.}
Without visual prompting, the agent loses track of the goal after detouring around obstacles. Without iterative reasoning, it follows myopic straight-line paths, leading to frequent collisions. Our full model combines both to produce robust, goal-directed trajectories.}
\label{fig:qual}
\vspace{-2ex}
\end{figure*}

\subsection{Agent Layer: Visually Grounded Navigation}
\label{sec:exp_agent}
\noindent\textbf{Main Results for Single Goal Navigation.}  
\cref{tab:main} indicates several key insights: {\textbf{1)}} our method consistently outperforms state-of-the-art VLN approaches, achieving an SR approximately 30$\%$ higher than the strongest baseline. This substantial margin underscores our model's remarkable visual grounding capacity; {\textbf{{2)}}} Our CR is on par with or lower than baselines that fine-tune VLMs on expert navigation trajectories. This demonstrates that our proposed visual prompting and iterative reasoning equip the agent with effective goal-oriented planning and collision avoidance; {\textbf{3)}} Occasional collisions in crowded scenes stem from agents' limited FoV blind spots for nearby obstacles and the difficulty of reacting to fast-moving pedestrians.
Qualitative results in \cref{fig:qual} further showcase our agent's complex reasoning capability: it successfully bypasses obstacles and navigates to distant destinations, demonstrating strong long-horizon planning. 
Moreover, results on XGRIDS and SAGE-3D (\cref{tab:main2}) show consistent advantages, confirming the robustness of our Agent Layer to various environments. \cref{fig:robustness_gallery} extensively illustrates this versatility across additional diverse scenes.

\noindent\textbf{Main Results for Long-horizon Planning.} The multi-goal benchmark (\cref{tab:main2}, right) shows our method outperforms all baselines significantly. Multi-goal episodes require chaining multiple sub-tasks, where failure at any stage can compound over time, making the problem substantially more challenging and placing greater demands on long-term memory. Our iterative reasoning mechanism re-evaluates the scene state at each sub-goal, while visual prompting continuously anchors the spatial location of the next target, thereby preserving planning coherence throughout the episode.

\vspace{0.3ex}

\noindent\textbf{Ablation of Spatial-Aware Visual Prompting.}
As shown in~\cref{tab:ablation}, visual prompting improves SR with notable gains in ObstNav by emphasizing object locations. 
Without it, the agent tends to follow the outdated VLM plans and loses track of the goal near its vicinity or after detouring for obstacle avoidance~(\cref{fig:qual}, row 1 \& 3). Our spatially-aware prompts inject strong semantic cues about the goal's position, urging the agent to update its plan and correct for initial inaccuracies. By better reconciling conflicts between textual and visual information, this mechanism boosts SR by over $10\%$ consistently (from (c) to (d) in~\cref{tab:ablation}). 

\vspace{0.3ex}

\noindent\textbf{Ablation of Iterative Reasoning.}
Integrating the iterative reasoning mechanism into our strategy yields substantial improvements across all three metrics, as evidenced by~\cref{tab:ablation}. The gains in SR and SPL indicate that iterative reasoning aids long-horizon goal achievement and prevents myopic straight-line paths (row two in~\cref{fig:qual}). It also significantly reduces CR, as agents tend to navigate around obstacles more carefully during the planning. Interestingly, applying iterative reasoning without visual prompting decreases SR while achieving the lowest CR (\cref{tab:ablation} (c)). This occurs because the agent becomes overly cautious to avoid collisions yet loses track of the goal without visual cues (\cref{fig:qual}).

\noindent\textbf{Effect of locomotion style.}
We evaluate with different locomotion styles in \cref{tab:locomotion}. While \textit{Running} naturally leads to higher CR due to increased speed, \textit{Slow} motion prioritizes safety, achieving the lowest CR. 
Notably, even our \textit{Running} style outperforms the strongest baseline Uni-NaVid~\cite{zhang2024uninavid}, highlighting the robustness of our planning framework across varying locomotion dynamics.

\begin{table}[t]
\centering
\caption{\textbf{Navigation performance under different locomotion styles.} Evaluated on SimNav and ObstNav of the \textit{SmallCity}~\cite{kerbl2024hiergs} scene with varying motion types.}
\vspace{-1.5ex}
\label{tab:locomotion}
\setlength{\tabcolsep}{4pt}
\resizebox{\linewidth}{!}{
\begin{tabular}{@{}l|ccc|ccc@{}}
\toprule
\multirow{2}{*}{\textbf{Method}} & \multicolumn{3}{c|}{\textbf{SimNav}} & \multicolumn{3}{c}{\textbf{ObstNav}} \\
\cmidrule(lr){2-4} \cmidrule(l){5-7}
& SR~$\uparrow$ & SPL~$\uparrow$ & CR~$\downarrow$ & SR~$\uparrow$ & SPL~$\uparrow$ & CR~$\downarrow$ \\
\midrule
Uni-NaVid (Default)~\cite{zhang2024uninavid} & 38.8$\%$ & 0.370 & 12.8$\%$ & 25.3$\%$ & 0.242 & 29.4$\%$ \\
\midrule
Ours (Default) & \bone 68.3$\%$ & \bone 0.640 & \bone 13.3$\%$ & \btwo 55.8$\%$ & \bone 0.516 & \btwo 30.8$\%$ \\
Ours (Run) & 56.7$\%$ & 0.518 & 35.8$\%$ & 41.6$\%$ & 0.382 & 54.2$\%$ \\
Ours (Slow) & \btwo 67.5$\%$ & \btwo 0.595 & \btwo 14.5$\%$ & \bone 56.4$\%$ & \btwo 0.507 & \bone 29.7$\%$ \\
\bottomrule
\end{tabular}
}
\vspace{-1.5ex}
\end{table}

\begin{figure}[t]
  \centering

  \begin{subfigure}[b]{0.46\linewidth}
    \centering
    \includegraphics[width=\linewidth]{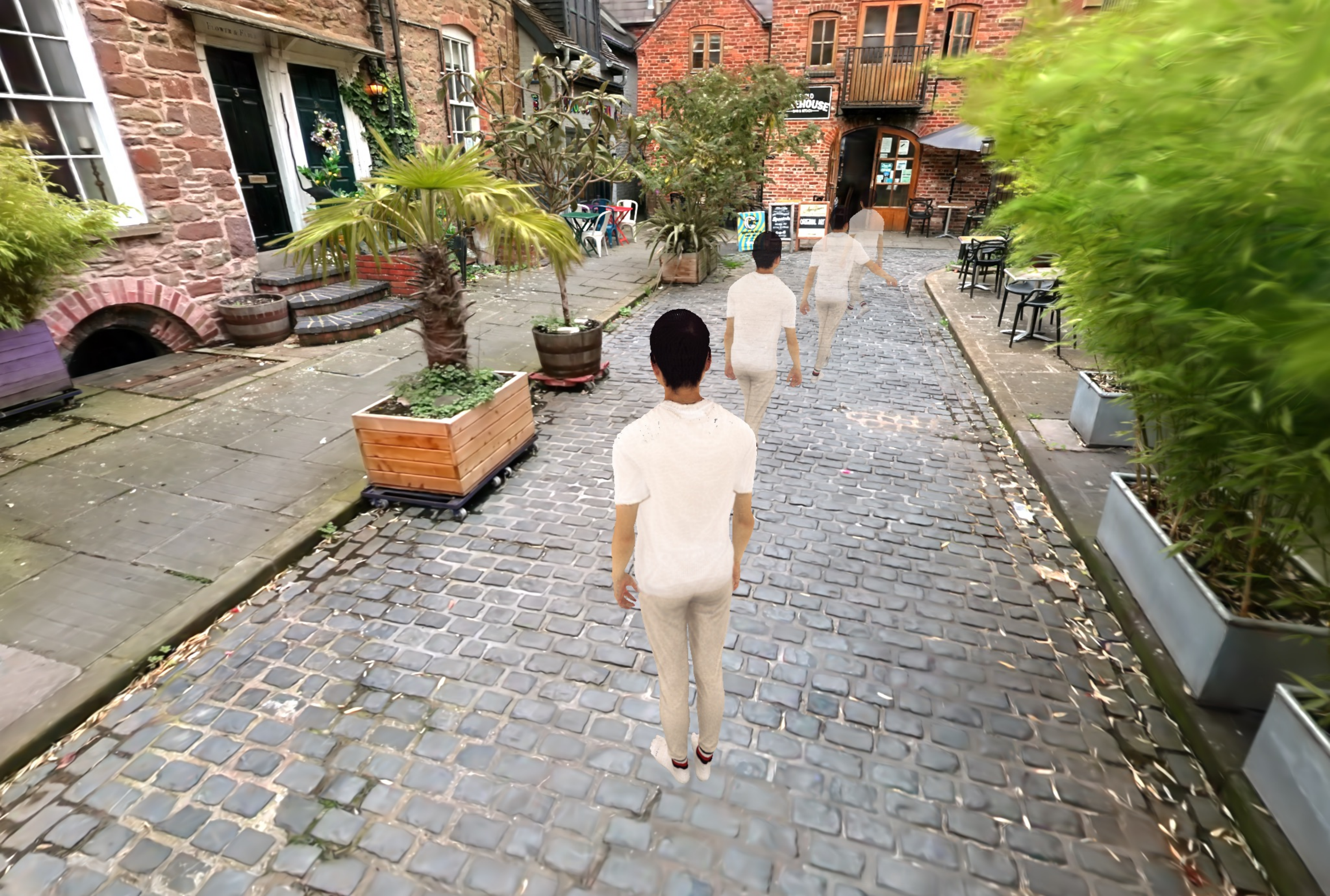}
    \caption{Go to the \textit{Restaurant}}
  \end{subfigure}
  \hspace{0.03\linewidth}
  \begin{subfigure}[b]{0.46\linewidth}
    \centering
    \includegraphics[width=\linewidth]{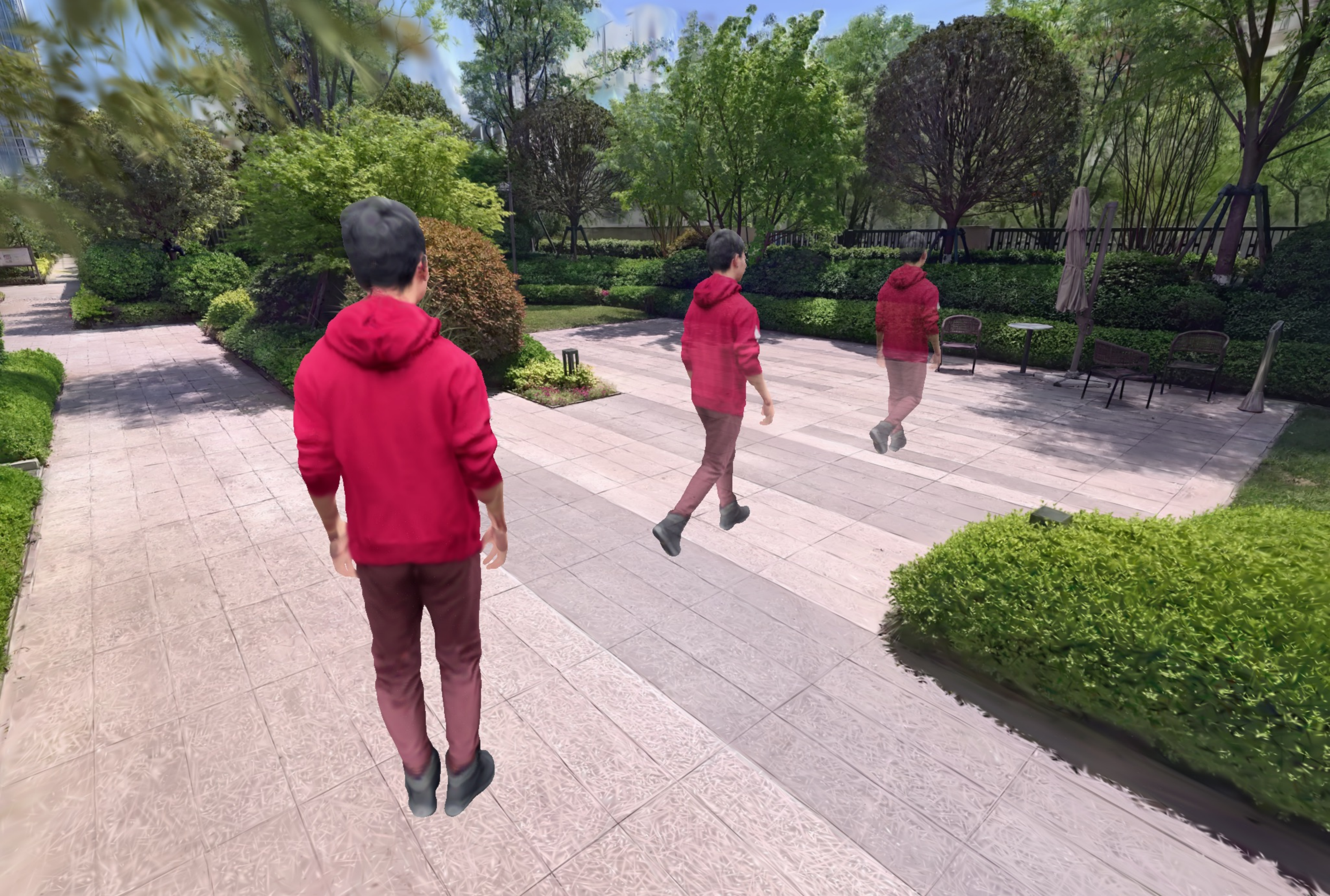}
    \caption{Go to the \textit{Table}}
  \end{subfigure}

  \begin{subfigure}[b]{0.46\linewidth}
    \centering
    \includegraphics[width=\linewidth]{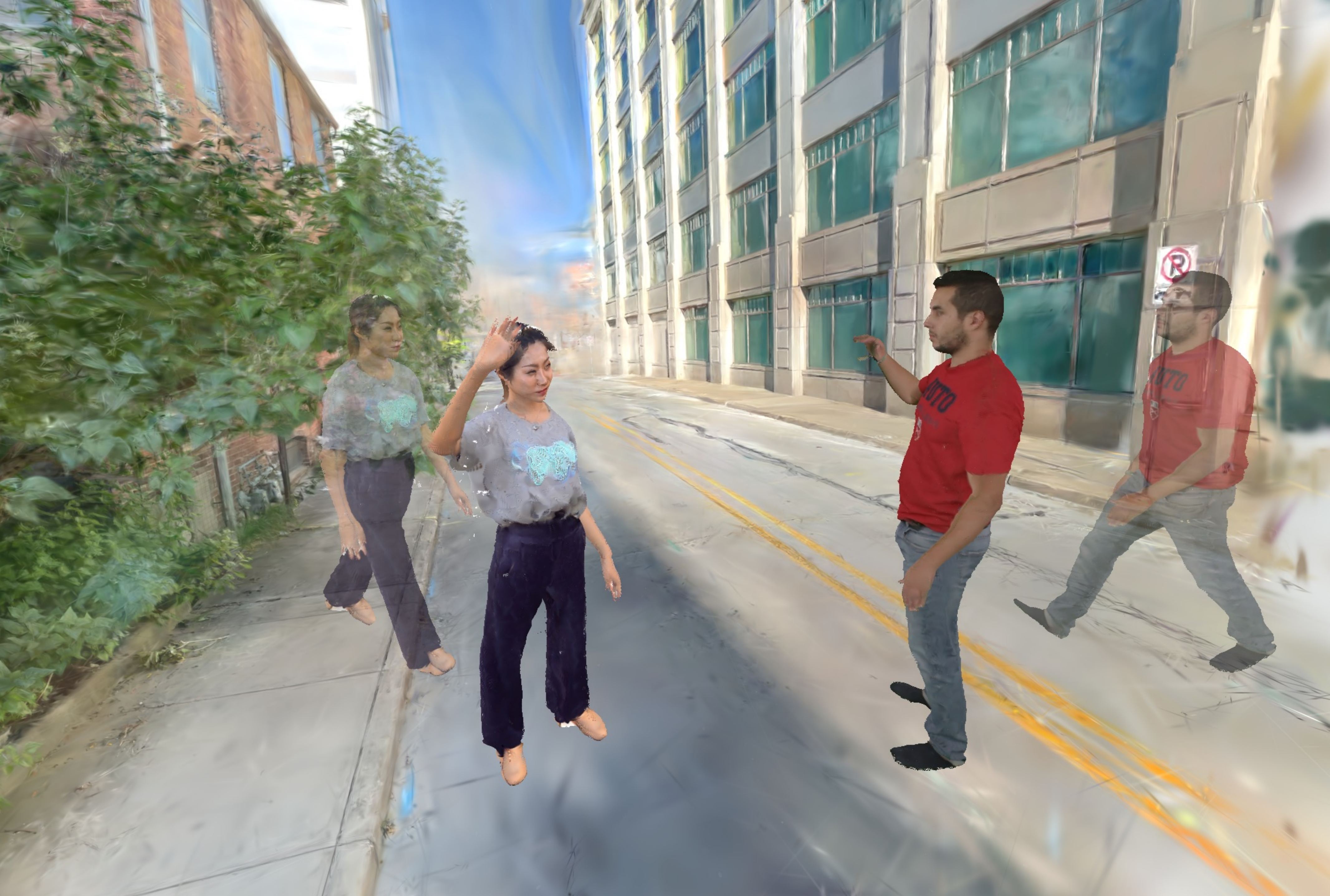}
    \caption{Social Interaction}
  \end{subfigure}
  \hspace{0.03\linewidth}
  \begin{subfigure}[b]{0.46\linewidth}
    \centering
    \includegraphics[width=\linewidth]{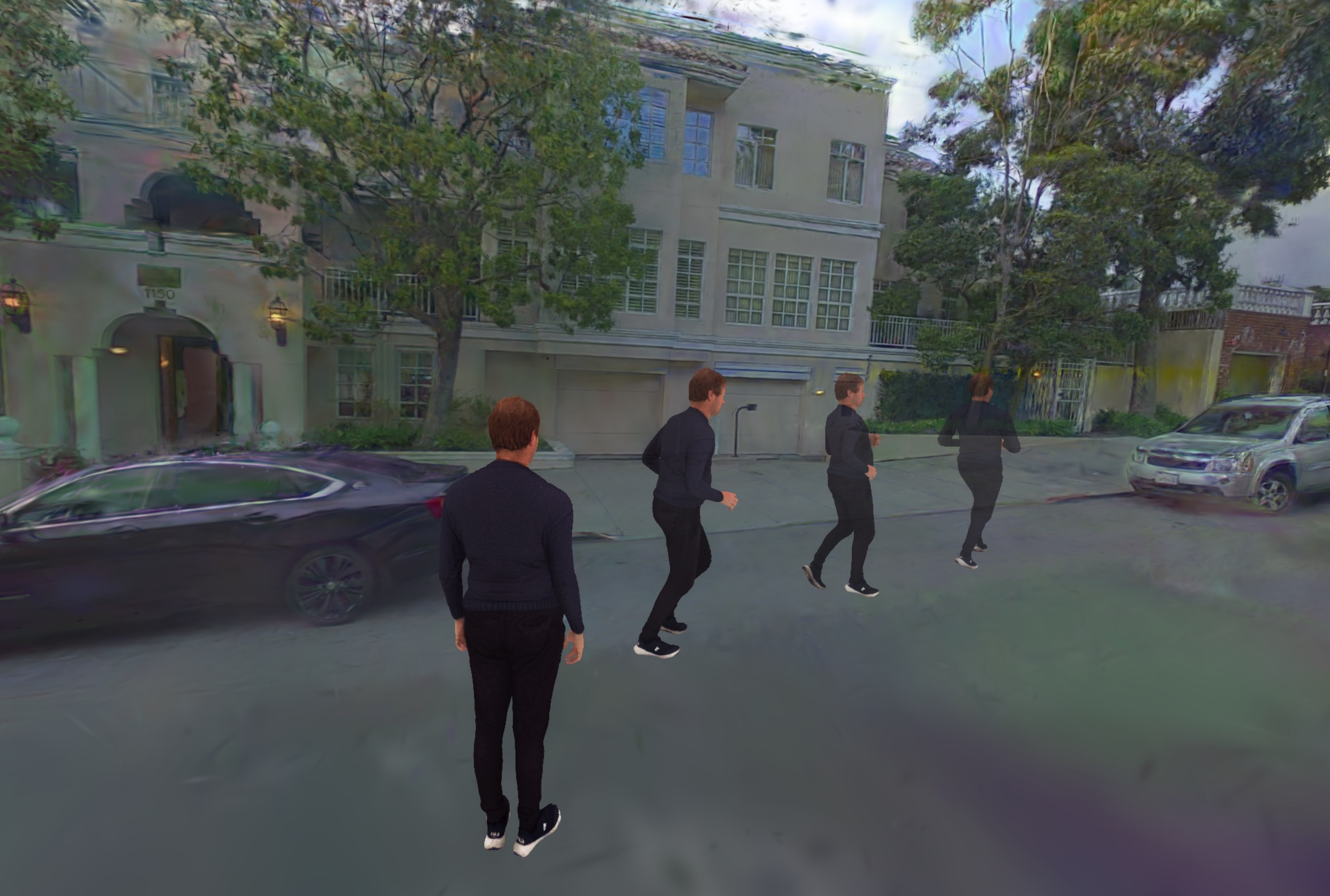}
    \caption{Run to the Car}
  \end{subfigure}

  \begin{subfigure}[b]{0.46\linewidth}
    \centering
    \includegraphics[width=\linewidth]{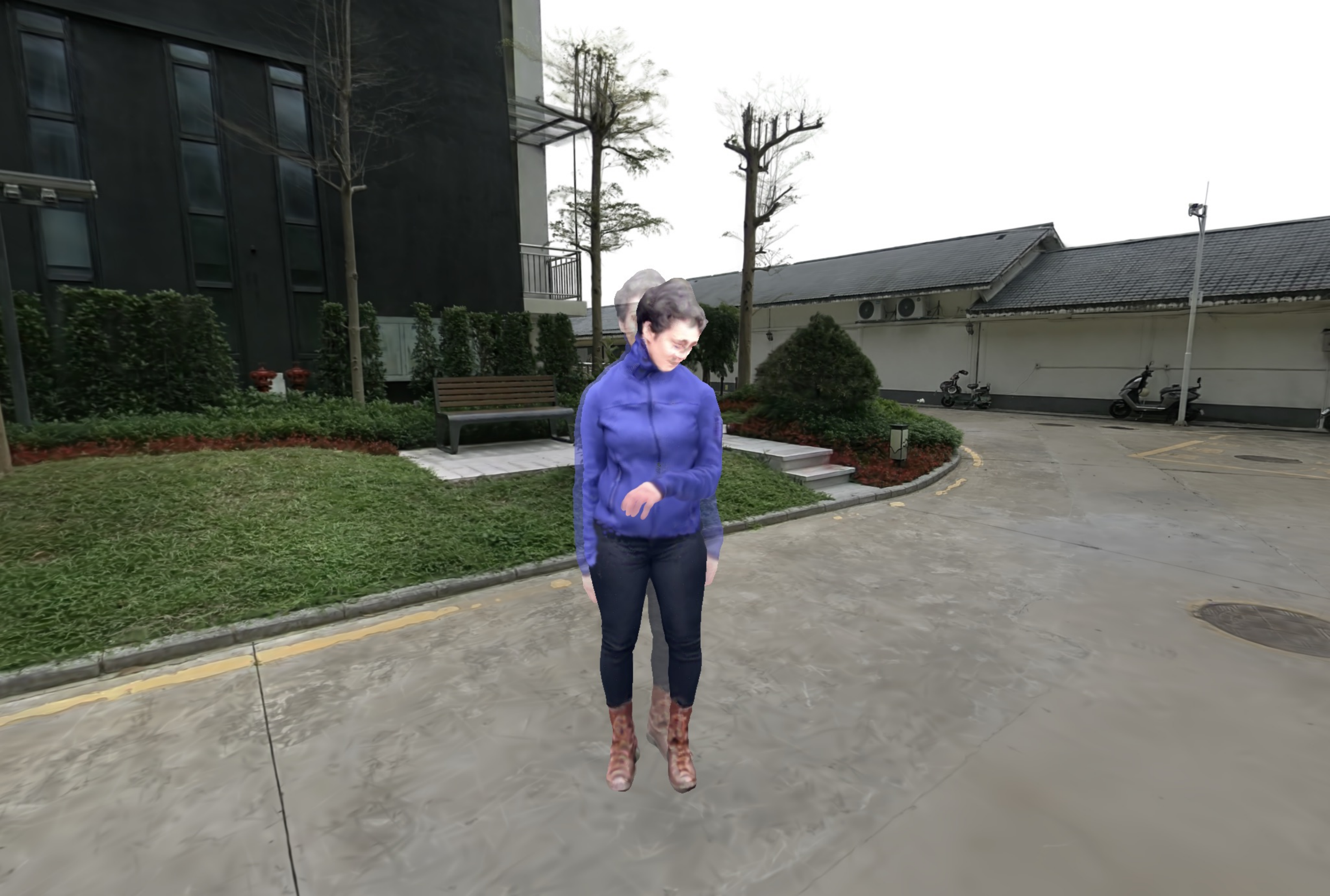}
    \caption{Check the Watch}
  \end{subfigure}
  \hspace{0.03\linewidth}
  \begin{subfigure}[b]{0.46\linewidth}
    \centering
    \includegraphics[width=\linewidth]{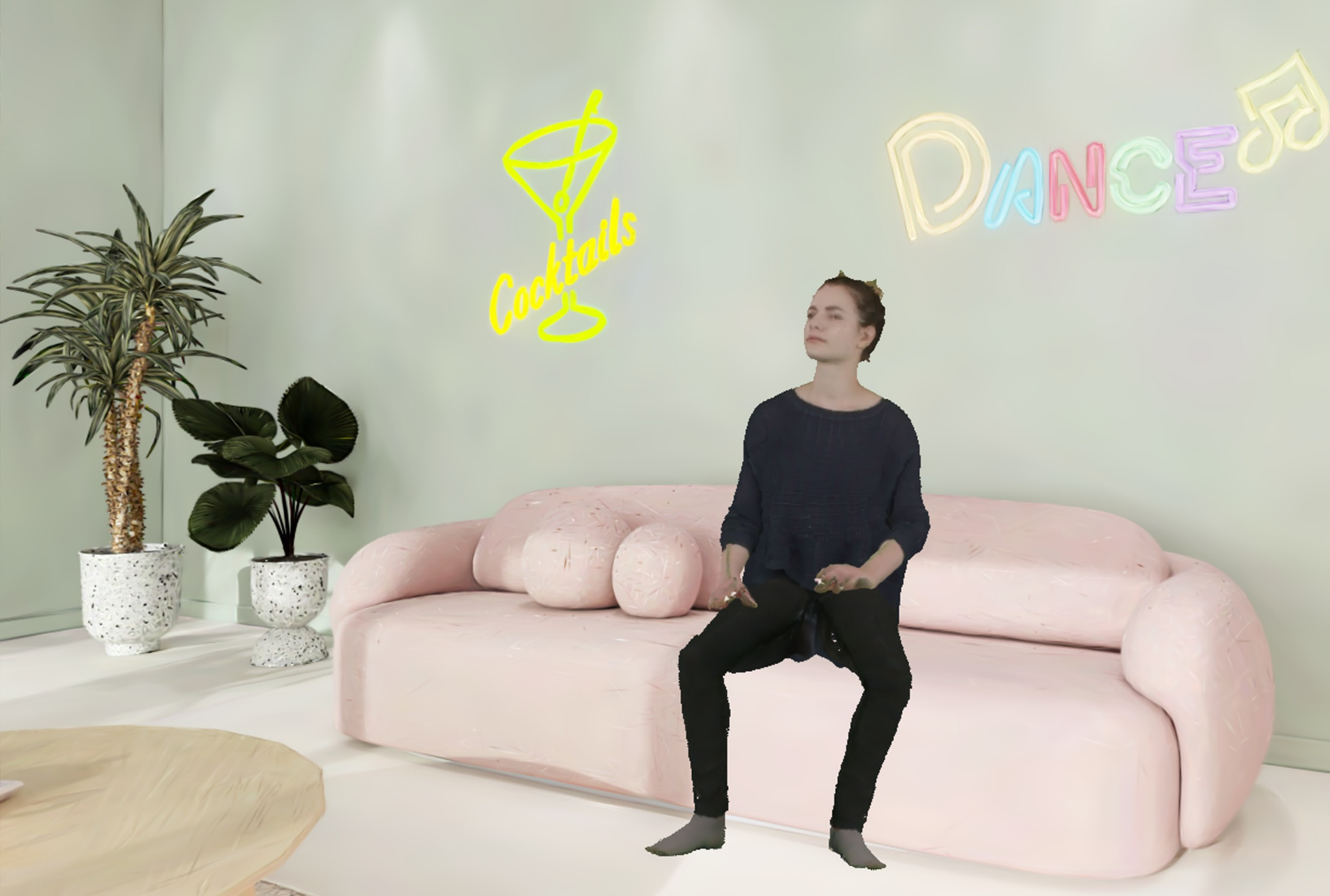}
    \caption{Sit on the Sofa}
  \end{subfigure}

  \vspace{-1.5ex}
  \caption{\textbf{Qualitative results across diverse environments.}}
  \label{fig:robustness_gallery}
  \vspace{-3ex}
\end{figure}

\begin{table}[t]
\centering
\caption{\textbf{Quantitative evaluation of semantic scene reconstruction.} Comparison with Feature-3DGS~\cite{zhou2024featuregs}, GW-3DGS~\cite{joseph2025gradient}, and OpenGaussian~\cite{wu2024opengaussian}, along with ablation of our proposed components in the World Layer. The \colorbox{bestcolor}{best} results are highlighted.}
\vspace{-1ex}
\label{tab:comparison_and_ablation}
\resizebox{\linewidth}{!}{
\setlength{\tabcolsep}{2pt}
\begin{tabular}{@{}l|ccc|ccc@{}}
\toprule
\multirow{2}{*}{Metric} & \multicolumn{3}{c|}{Baselines} & \multicolumn{3}{c}{Ablation (Ours)} \\
\cmidrule(lr){2-4} \cmidrule(l){5-7}
& F-3DGS & GW-3DGS & OpenGS & (a) Base & (b) +Occ & (c) +ViewSel \\
\midrule
mIoU~$\uparrow$ & 0.256 & 0.025 &  0.483 & 0.518 & 0.556 & \bone 0.601 \\
mAcc~$\uparrow$ & 0.647 & 0.109 & 0.675 & 0.636 &  0.699 & \bone 0.733 \\
\bottomrule
\end{tabular}
}

\end{table}
\vspace{1ex}

\subsection{World Layer: Semantic Scene Reconstruction}
\label{sec:ablation}
\begin{figure}[h]
  \centering
    \vspace{-0.3cm}
  \includegraphics[width=0.97\linewidth]{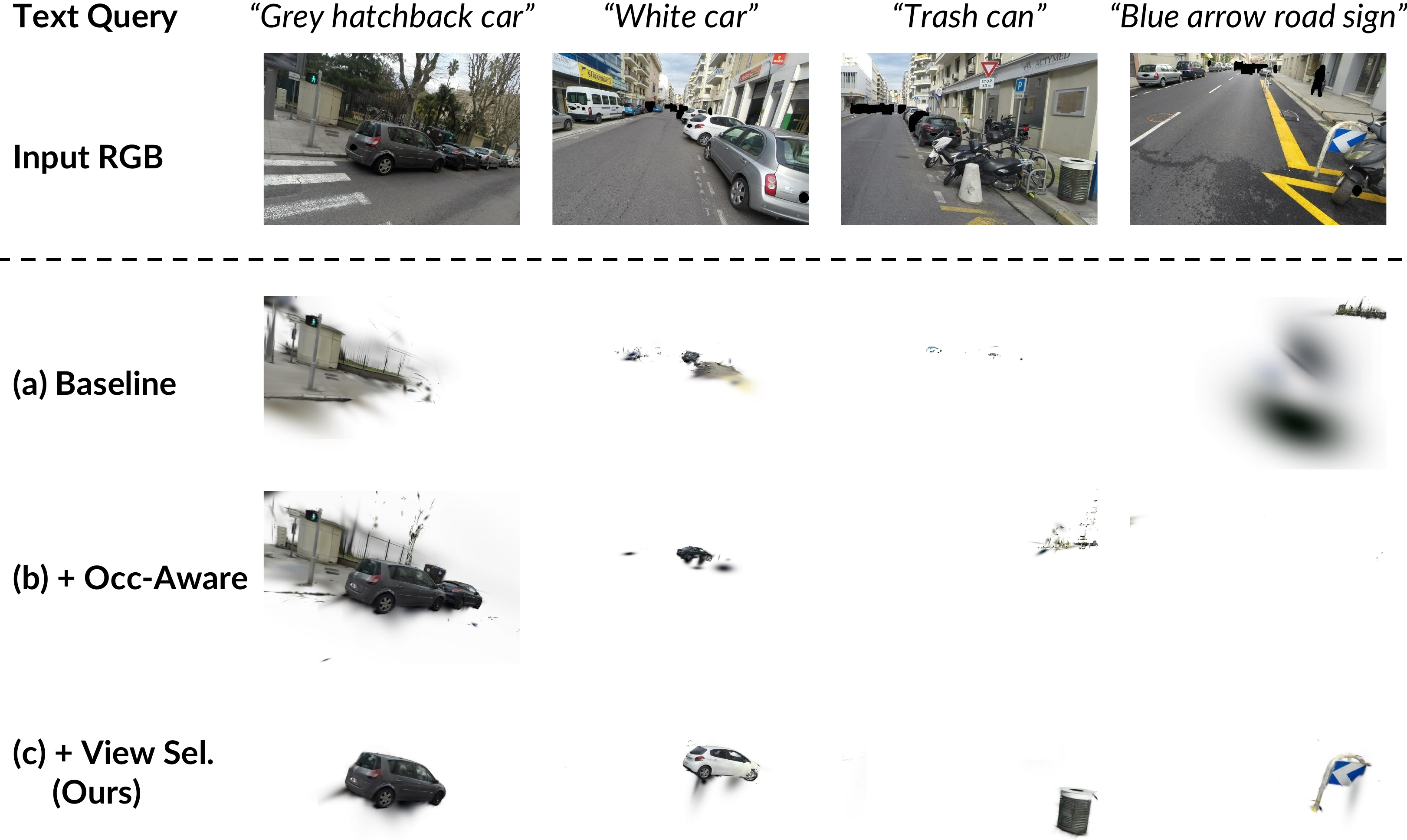}
  \vspace{-0.3cm}
 \caption{\textbf{Qualitative Ablation on occlusion-aware semantic scene reconstruction.} Our framework achieves precise 3D instance segmentation with well-defined boundaries, demonstrating robustness to severe occlusion while successfully recognizing thin or small objects in large-scale outdoor scenes. Zoom in for details. 
 }
\label{fig:scene_ablation}
\vspace{-0.3cm}
\end{figure}

\noindent\textbf{Comparison with baselines.}
As shown in \cref{tab:comparison_and_ablation}, feature-lifting methods designed for indoor or small-scale scenes degrade severely in our large-scale outdoor setting. OpenGS performs more competitively but still falls short due to its lack of occlusion handling. In sharp contrast, our full method surpasses all baselines by a substantial margin.

\vspace{0.5ex}

\noindent\textbf{Ablation of Occlusion-Aware masks and view selection.}
We evaluate the effectiveness of our occlusion-aware semantic reconstruction pipeline. 
As shown in \cref{tab:comparison_and_ablation} and \cref{fig:scene_ablation}, occlusion-aware masks prevent erroneous feature propagation; without them, signals contaminated by foreground occluders result in fragmented and noisy objects~(\cref{fig:scene_ablation}, first column).
However, masks alone are insufficient for fine-grained results. Our view selection strategy complements them by prioritizing frames with high instance visibility. This enhances the model's ability to learn fine details, thus enabling the recovery of thin or sparsely observed objects (\eg, trash cans). Consequently, the combination of both modules yields the best performance, producing semantically cleaner and more complete reconstructions.

% \vspace{0.5ex}
\section{Conclusion} \label{sec:conclusion}
In this work, we introduced Visually-grounded Humanoid Agents, a coupled two-layer paradigm for embodied digital humans, consisting of a world layer that reconstructs semantically enriched 3D environments with animatable Gaussian avatars and an agent layer that governs their behavior through perception–action loops.
Looking ahead, we plan to extend this framework to richer embodied skills, such as conversations and physical interaction with scenes, and applications in robotics, such as human-centric robotic learning.

\clearpage

\section*{Acknowledgments}
We sincerely thank Wentao Zhu, Honglin He, and Shikun Ban for their insightful feedback and valuable discussions throughout the course of this project.

{
    \small
    \bibliographystyle{ieeenat_fullname}
    \bibliography{reference_header,main}
}

\cleardoublepage
\renewcommand\thesection{\Alph{section}}  
\renewcommand\thefigure{\Alph{section}\arabic{figure}} 
\renewcommand\thetable{\Alph{section}\arabic{table}}
\renewcommand\theequation{\Alph{section}\arabic{equation}}
\setcounter{section}{0}
\setcounter{figure}{0}
\setcounter{table}{0}
\setcounter{equation}{0}
\counterwithin*{figure}{section}
\counterwithin*{table}{section}
\counterwithin*{equation}{section}

\maketitlesupplementary

\vspace{2ex}
This supplementary material further enriches and clarifies the contributions presented in the main paper.
Sec.~\ref{sec-ip-details} provides detailed implementation settings for both our framework and the benchmark environments.
Sec.~\ref{supp_sec_bench} elaborates on the design of the benchmark, including task definitions, scenario generation, and baseline details.
Sec.~\ref{sec:exp_supp} presents additional quantitative and qualitative results addressing several key questions.
Sec.~\ref{sec-supp-related} discusses additional preliminary concepts and related works.
Finally, Sec.~\ref{sec-supp-limitation} outlines the limitations and future directions of our approach.

\makeatletter
\titlecontents{section}
    [0em]
    {\addvspace{0.5pc}}
    {\contentsmargin{0em}\thecontentslabel\quad}
    {\contentsmargin{0em}}
    {\titlerule*[0.5pc]{.}\contentspage}
\titlecontents{subsection}
    [1.5em]
    {\addvspace{0.2pc}}
    {\contentsmargin{0em}\thecontentslabel\quad}
    {\contentsmargin{0em}}
    {\titlerule*[0.5pc]{.}\contentspage}
\setcounter{tocdepth}{2}
\startcontents[supp]
\printcontents[supp]{}{0}{\section*{Contents of Supplementary Material}}
\makeatother

\vspace{2ex}

\section{Implementation Details}\label{sec-ip-details}

\subsection{Semantic Scene Reconstruction}\label{supp:scene-recon}
In this subsection, we detail how we overcome the bottlenecks of state-of-the-art approaches to improve reconstruction and annotation quality in large-scale scenes through several practical designs. Furthermore, we provide comprehensive details regarding our training schedules.

\vspace{0.5ex}

\noindent\textbf{Large-scale Scene reconstruction.}
We partition the \textit{SmallCity} scene~\cite{kerbl2024hiergs} into 4 spatially adjacent blocks ($2 \times 2$ layout), each reconstructed in parallel with the standard CityGaussian pipeline~\cite{liu2024citygaussian, liu2024citygaussianv2} and subsequently fused. To improve geometric fidelity, we enhance the base reconstruction with three supervision signals: 1) a geometric consistency loss inspired by 2DGS~\cite{huang20242dgs} that regularizes the 3D Gaussians towards 2D disks; 2) a depth loss guided by monocular depth priors from~\cite{yang2024depthv2}, following Hier-GS~\cite{kerbl2024hiergs}, which substantially improves quality on flat road surfaces; and 3) a normal regularization loss from Vid2Sim~\cite{xie2025vid2sim} that yields a smoother underlying mesh for accurate collision detection, as illustrated in~\cref{fig:normal_reg}.

\begin{figure}[t]
  \centering
  \begin{subfigure}{0.48\linewidth}
    \centering
    \includegraphics[width=\linewidth]{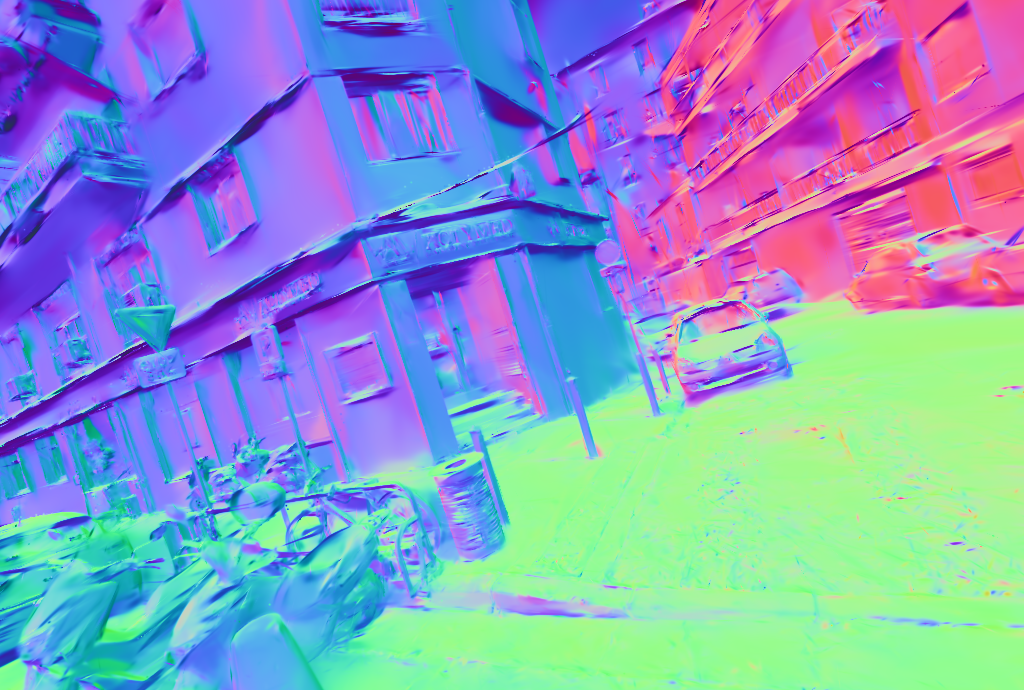}
    \vspace{0.5ex}
    \includegraphics[width=\linewidth]{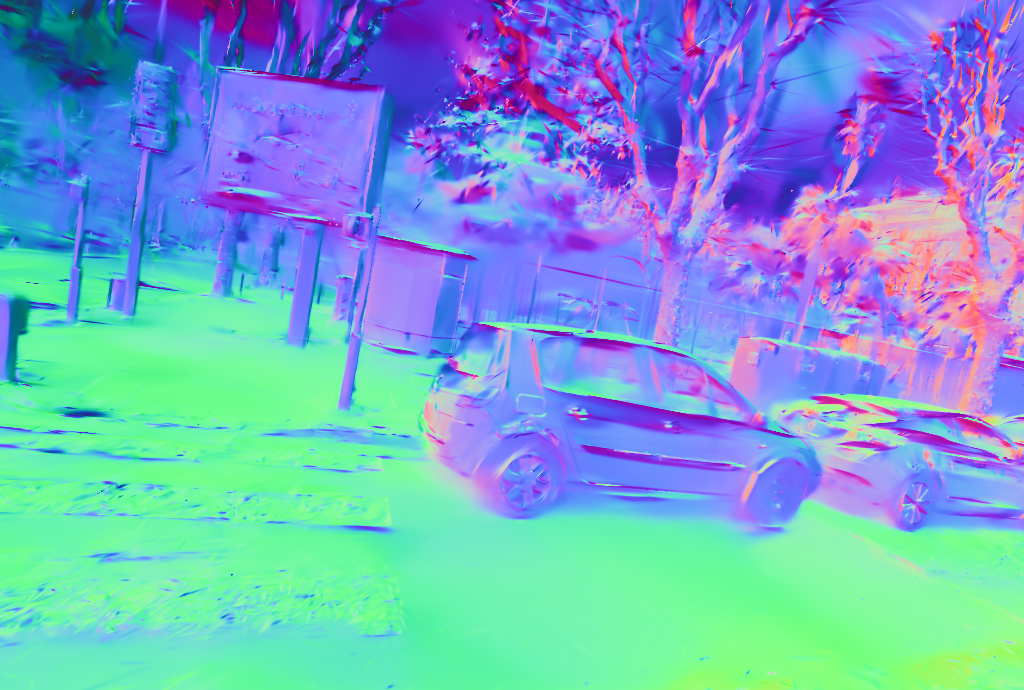}
    \caption{Without normal regularization}
    \label{fig:before_normal_reg}
  \end{subfigure}
  \hfill
  \begin{subfigure}{0.48\linewidth}
    \centering
    \includegraphics[width=\linewidth]{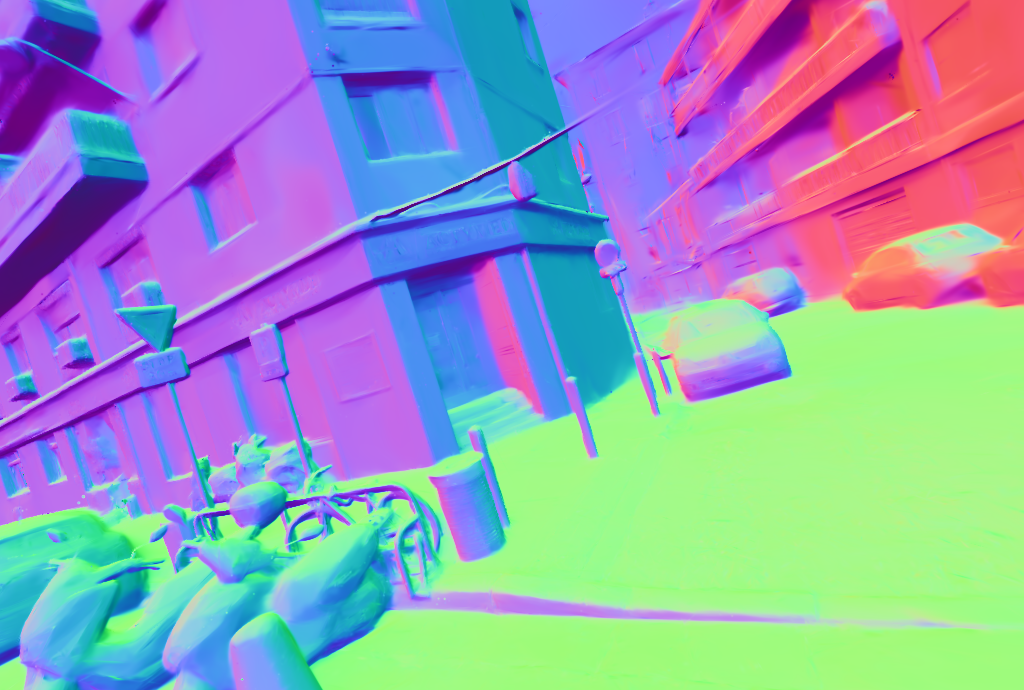}
    \vspace{0.5ex}
    \includegraphics[width=\linewidth]{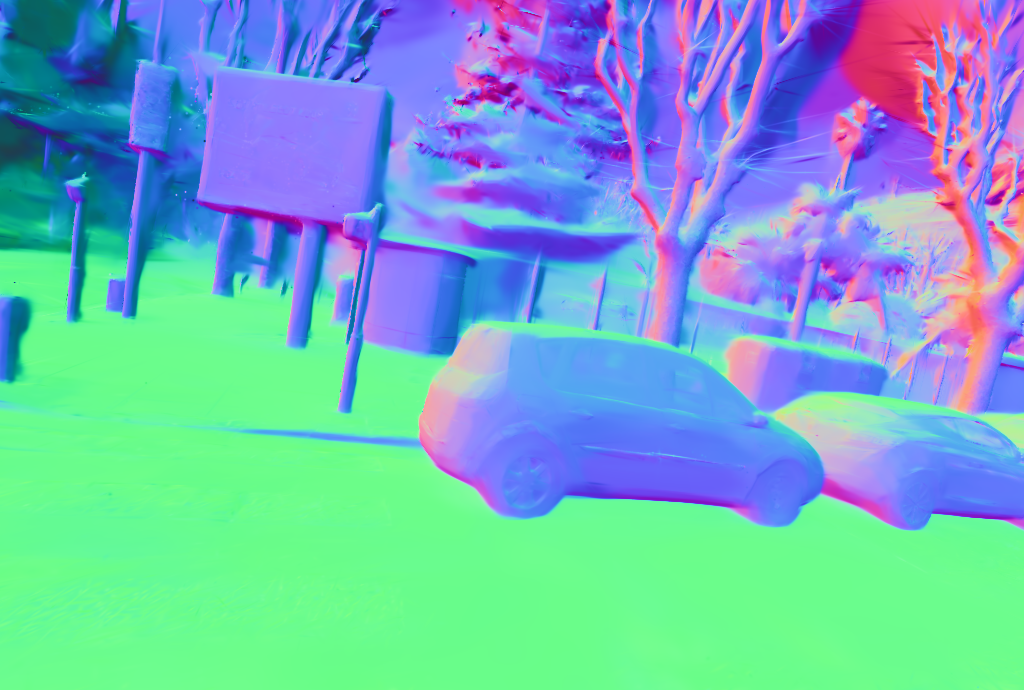}
    \caption{With normal regularization}
    \label{fig:after_normal_reg}
  \end{subfigure}
  \vspace{-1.5ex}
  \caption{\textbf{Effect of normal regularization on collision mesh extraction.} With normal regularization, the surface becomes smoother, enabling more reliable collision detection.}
  \label{fig:normal_reg}
\end{figure}

Training follows a two-stage schedule. The coarse stage runs for 30{,}000 iterations, with densification every 1{,}000 steps starting from step 4{,}000. The depth loss weight is annealed from 1.0 to 0.1, while $\lambda_{\text{normal}} = \lambda_{\text{geo}} = 0.1$. Subsequently, the per-chunk fine stage is initialized from the coarse checkpoint and trained for an additional 60{,}000 iterations, with densification active from step 500 to 30{,}000. The depth loss  with weight annealed from 0.5 to 0.05.

\begin{figure*}[t]
  \centering
  \begin{subfigure}{0.32\linewidth}
    \centering
    \includegraphics[width=\linewidth]{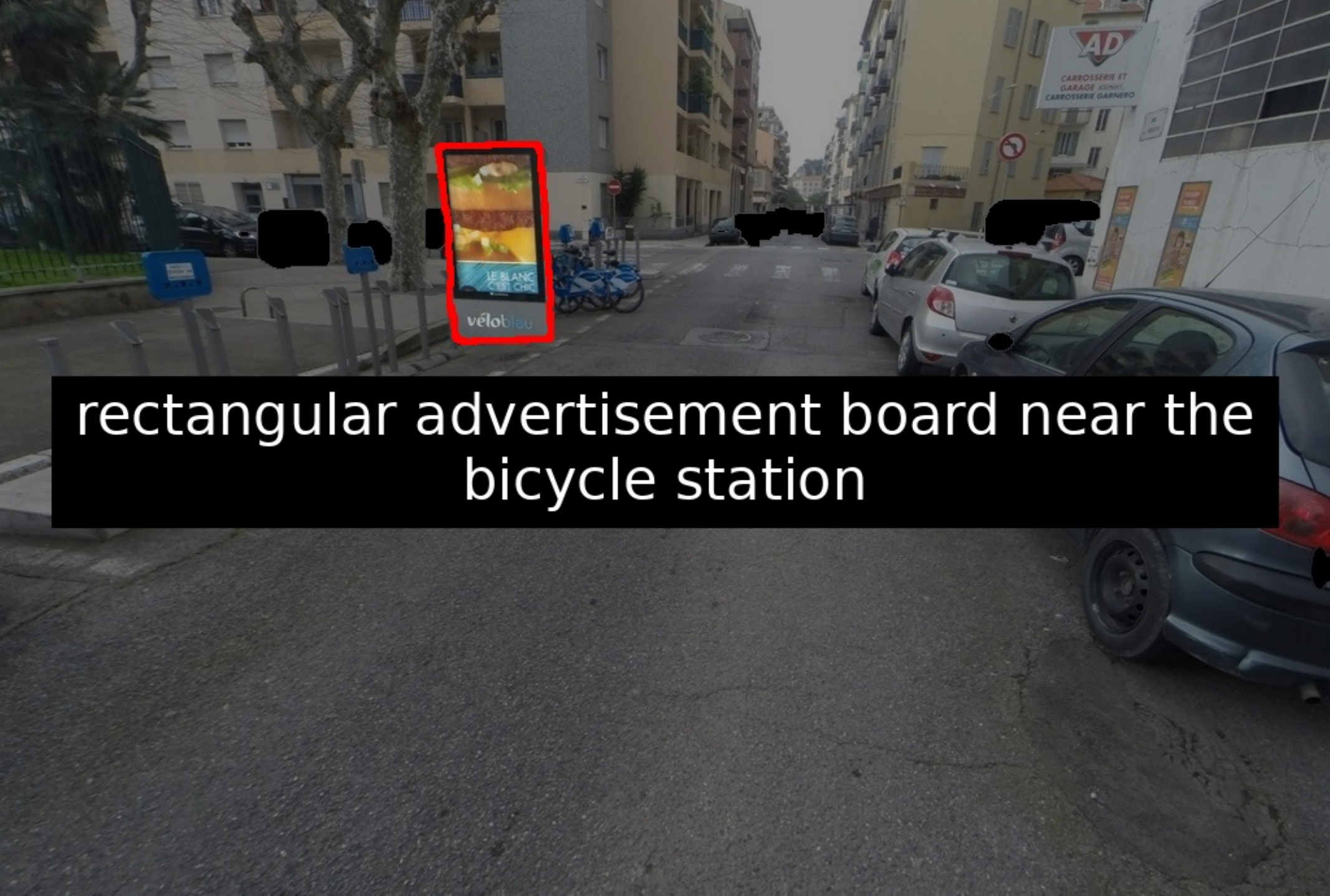}
  \end{subfigure}
  \begin{subfigure}{0.32\linewidth}
    \centering
    \includegraphics[width=\linewidth]{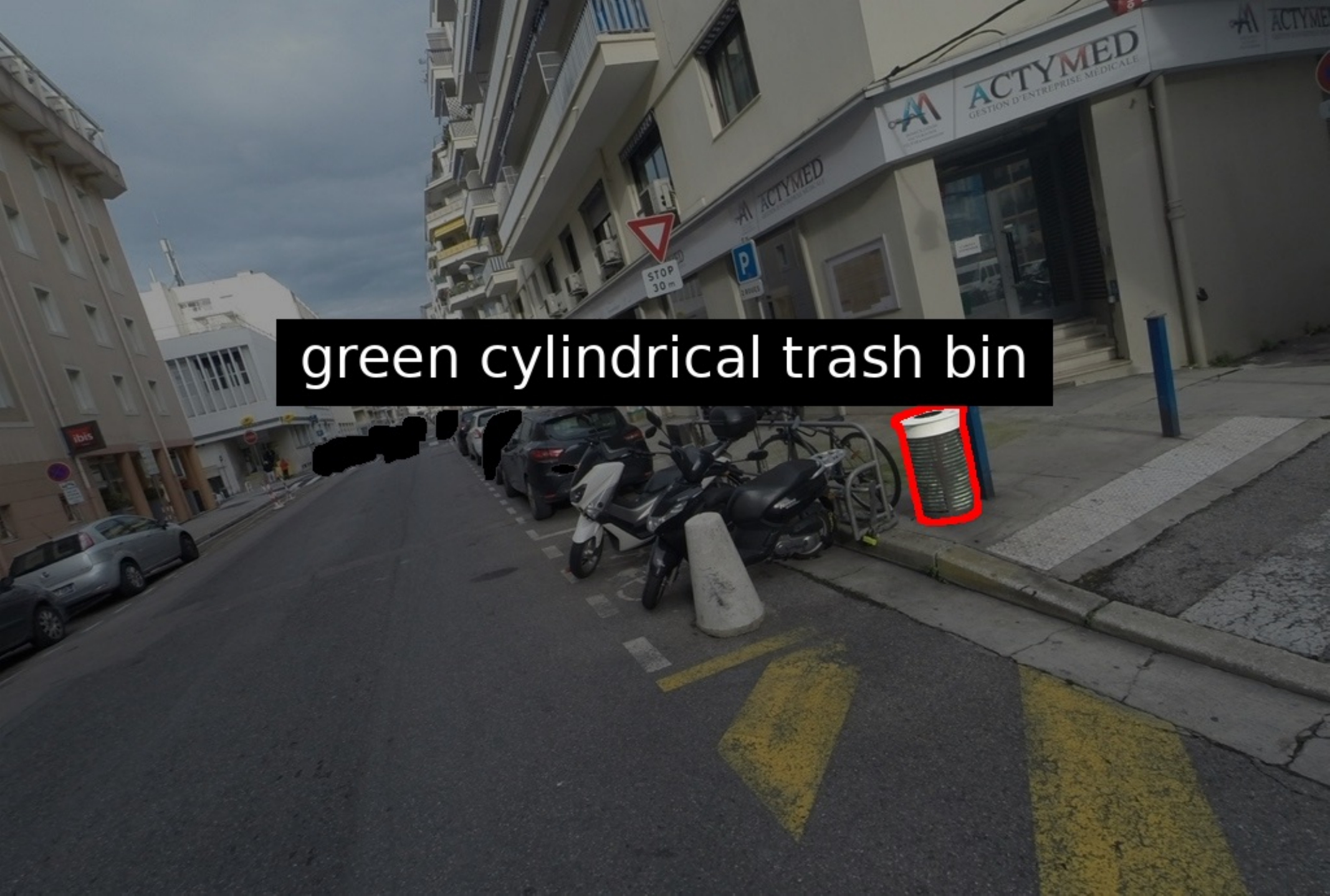}
  \end{subfigure}
  \begin{subfigure}{0.32\linewidth}
    \centering
    \includegraphics[width=\linewidth]{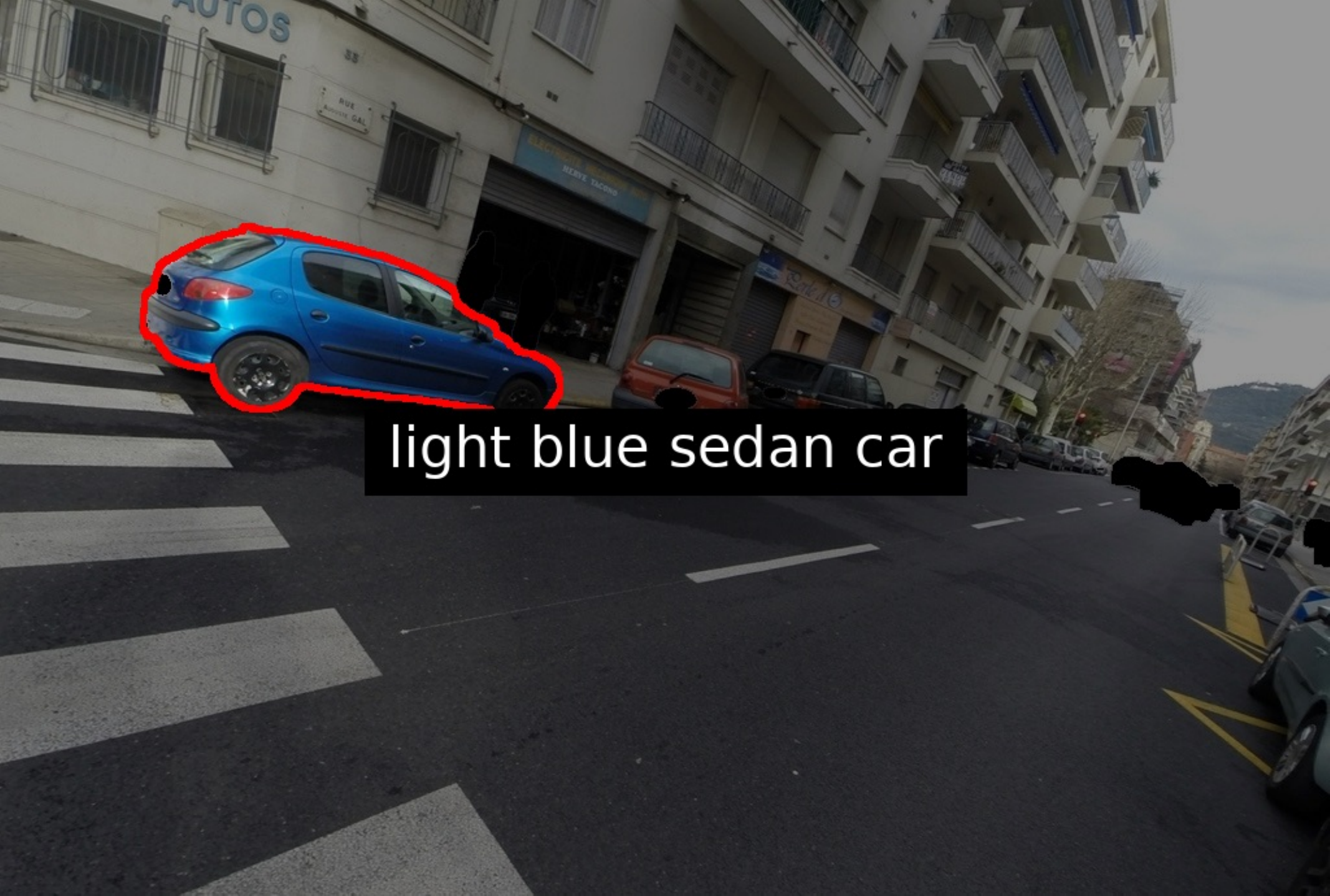}
  \end{subfigure}
  
  \vspace{-1ex}
  \caption{\textbf{Visualizations of Context-Aware Instance Annotation.} 
  For the VLM visual prompt, the target object is highlighted by a red contour, while the background is dimmed to suppress visual distraction while retaining spatial context. 
  The overlaid text shows the detailed descriptions generated by the annotation pipeline.}
  \label{fig:visual_prompt}
  \vspace{-2ex}
\end{figure*}

\vspace{0.5ex}

\noindent\textbf{Mesh Extraction.}
For mesh extraction, we first render paired RGB images and depth maps from the reconstructed 3DGS for each input view and fuse them into a Truncated Signed Distance Function (TSDF) field using Open3D~\cite{zhou2018open3d}. The mesh is extracted from this TSDF field with a voxel size of 0.1\,m and a depth truncation of 50\,m. To eliminate potential simulation instability caused by uneven ground reconstruction, we explicitly remove the ground plane from the extracted mesh.
Initially, we attempted to use Grounded-SAM~\cite{ren2024groundedsam} with the prompt ``ground'' to extract per-image masks. However, this approach struggled to consistently isolate the entire road surface due to prompt ambiguity. 
Instead, inspired by Vid2Sim~\cite{xie2025vid2sim}, we employ a geometric approach: we render the normal map for each input view and mask out pixels where the normal direction deviates from the vertical $+Z$ axis by less than a threshold $\theta$ (\ie, $15^{\circ}$). Similarly, we identify and remove sky regions using SAM-generated masks before mesh extraction. 
Please refer to the video for a visual examination of the mesh quality.

\vspace{0.5ex}

\noindent\textbf{Semantic Mask Generation.}
Following LeRF~\cite{kerr2023lerf} and LangSplat~\cite{qin2024langsplat}, we first utilize the automatic SAM mask generator~\cite{kirillov2023sam} to extract dense segmentation masks for all images to achieve semantic understanding. However, processing all fine-grained masks is computationally prohibitive for large-scale scenes. Since our downstream planning tasks prioritize instance-level understanding over part-level details (e.g., detecting a ``car'' rather than a ``car door''), we only retain the ``large-level" masks from SAM's three-level prediction hierarchy.
Furthermore, we discard masks corresponding to dynamic objects by checking their Intersection over Union (IoU) with the dynamic masks from the preprocessing stage; masks with an IoU $>50\%$ are removed. This curation step ensures a stable and meaningful set of instance masks for the subsequent annotation stage.

\vspace{0.5ex}

\noindent\textbf{Occlusion-Aware Semantic Training.}
For semantic scene reconstruction described in the main paper, the training proceeds in two stages:
\textit{Stage~1 (Semantic Feature Learning)} runs for 60,000 iterations with a learning rate of $10^{-3}$. The loss weights for the inter-mask contrastive and intra-mask smoothing are set to $\lambda_c = 1$ and $\lambda_s = 0.1$, respectively.
\textit{Stage~2 (Coarse-to-Fine Codebook Discretization)} employs a two-level codebook with $k_1=64$ root clusters and $k_2=32$ leaf clusters, yielding up to $64 \times 32 = 2,048$ distinct instance codes. To ensure accurate feature quantization under heavy inter-object occlusion, this stage incorporates occlusion-aware masks and strategic view selection. The discretization occurs in two sequential phases: 
First, root-level clustering relies on concatenated features and 3D coordinates (using a positional weight of 0.5). This phase is trained for 30,000 iterations with a learning rate of $10^{-2}$, updating centroids via k-means every 200 iterations. 
Subsequently, leaf-level clustering operates exclusively on the learned features. It is trained for 60,000 iterations with a learning rate of $5 \times 10^{-3}$, with centroids updated every 50 iterations.

\begin{figure}[t]
  \centering
  
  % --- LEFT COLUMN: One large image ---
  \begin{subfigure}[b]{0.69\linewidth} % Allocate ~65% width for the main image
    \centering
    \includegraphics[width=\linewidth]{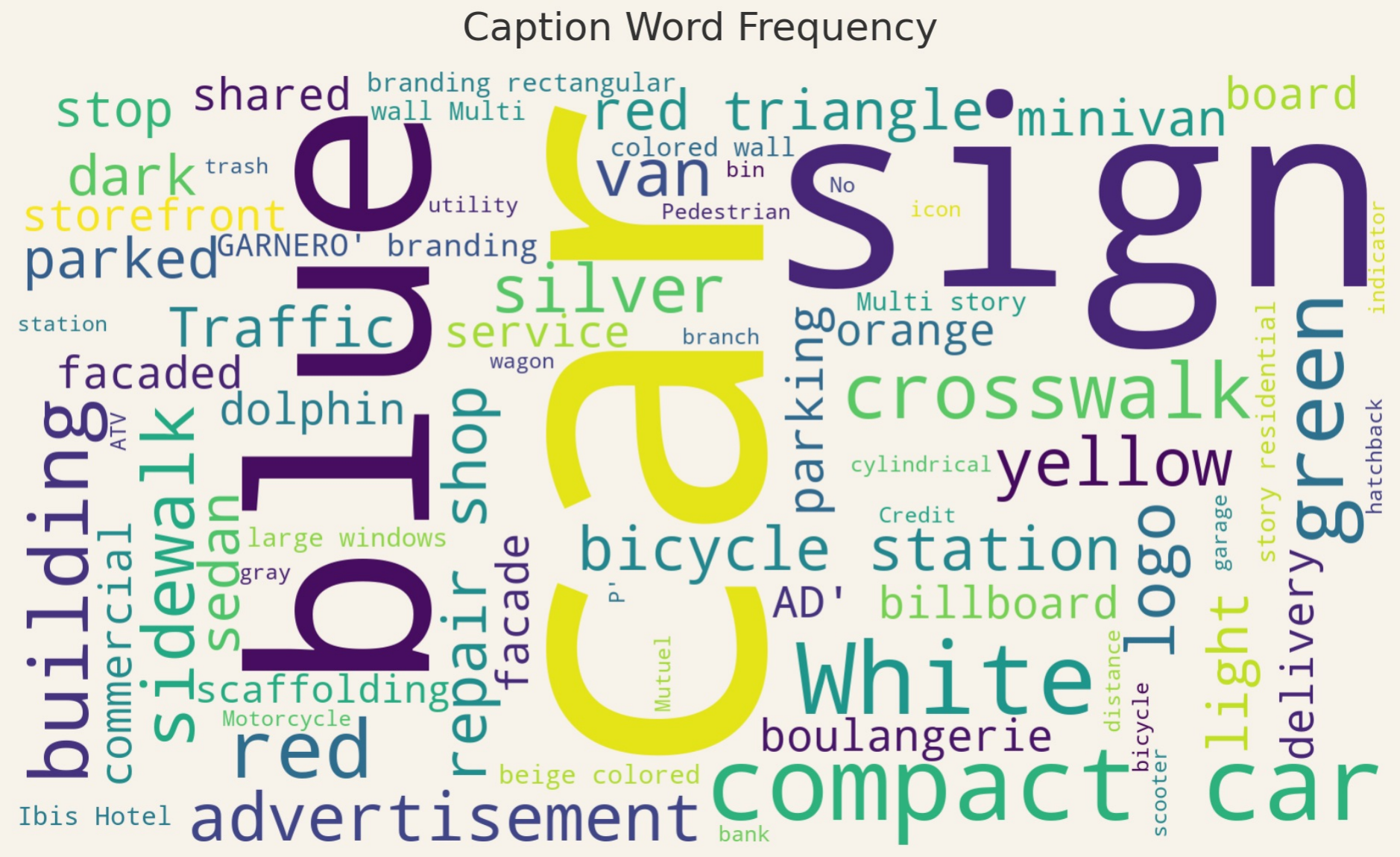}
    \caption*{\textit{SmallCity}~\cite{kerbl2024hiergs}}
  \end{subfigure}% <-- CRITICAL: This percent sign prevents a line break here
  \hfill % This pushes the left and right columns to the edges
  %
  % --- RIGHT COLUMN: A container for the two smaller images ---
  \begin{minipage}[b]{0.28\linewidth} % A minipage acts as a vertical stack container
    % Top small image
    \begin{subfigure}{\linewidth}
      \centering
      \includegraphics[width=\linewidth]{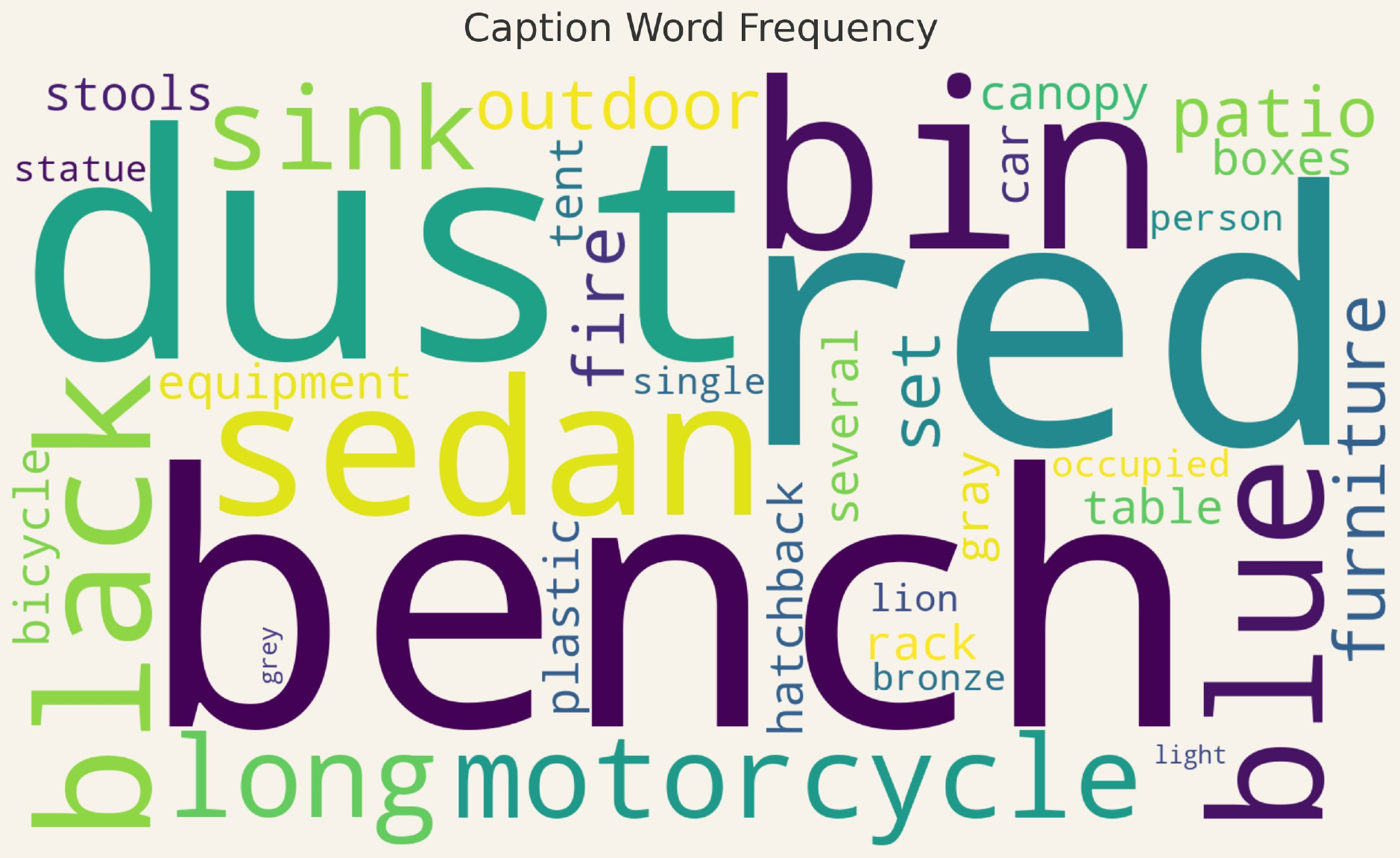}
      \caption*{\textit{XGRIDS}~\cite{xgrids}}
    \end{subfigure}\\[1ex] % \\[length] forces a newline and adds vertical padding
    %
    % Bottom small image
    \begin{subfigure}{\linewidth}
      \centering
      \includegraphics[width=\linewidth]{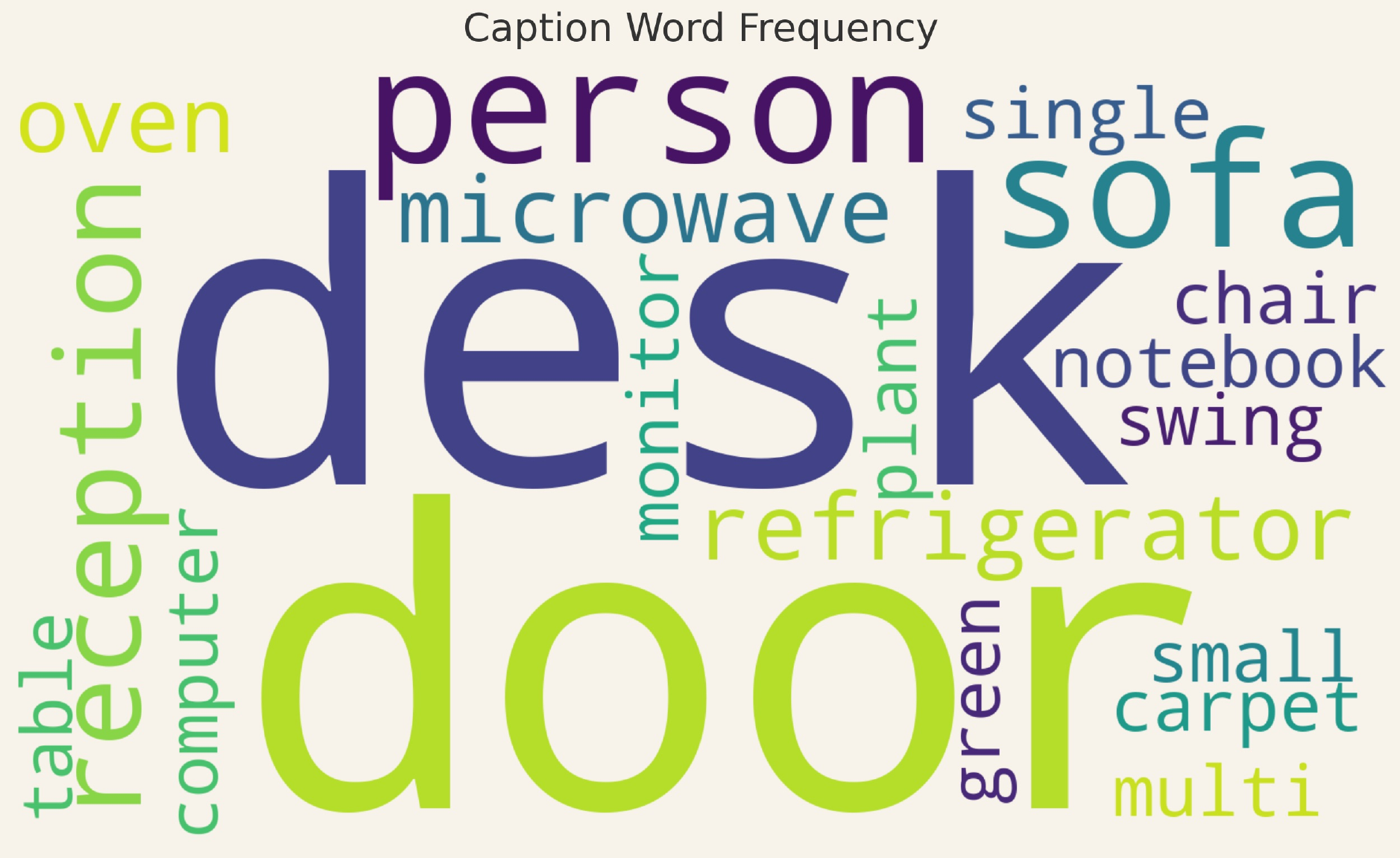}
      \caption*{\textit{SAGE-3D}~\cite{miao2025sage3d}}
    \end{subfigure}
  \end{minipage}

  \vspace{-1ex} % Adjust the gap between images and the main caption
  \caption{\textbf{Word cloud visualization of object semantic annotations across representative benchmark scenes.} For \textit{SmallCity}, annotations are generated by our automated open-vocabulary pipeline; for \textit{XGRIDS}, semantic goals are manually annotated; for \textit{SAGE-3D}, 3D bounding box annotations are provided by the original dataset. Larger words indicate more frequently occurring object categories.}
  \label{fig:wordcloud}
  \vspace{-3ex}
\end{figure}

\noindent\textbf{Automated Open-Vocabulary Annotation.} 
Once the 3D Gaussians are segmented using our two-stage discretization codebook, the final step is to assign meaningful textual annotations to each instance. Unlike prior works such as OpenGaussian~\cite{wu2024opengaussian}, which rely on predefined text prompts for querying (a ``text-to-instance'' task), our objective is the inverse: to automatically generate rich descriptions for discovered instances.
To fully automate this process, we leverage Qwen2.5-VL-72B~\cite{bai2025qwen2} as an annotator. 
A critical challenge lies in focusing the VLM on a specific instance within a cluttered scene without sacrificing global context. 
Approaches like ConceptGraph~\cite{gu2024conceptgraphs} typically crop the image around the instance mask; however, this removal of the background eliminates essential spatial cues. 
In large-scale outdoor scenes, where multiple objects of the same category frequently co-occur (e.g., multiple black sedans), contextual details (e.g., ``the car \textit{near the bus stop}'') are indispensable for resolving ambiguity.
To address this, we employ a context-aware visual prompting technique: we highlight the target object's contour in the image while strictly dimming the background opacity. 
This strategy effectively directs the VLM's attention to the target instance while preserving environmental context, enabling the generation of discriminative descriptions that encompass appearance, material, and spatial relationships.
We provide illustrative examples in \cref{fig:visual_prompt}, which demonstrate how our method generates fine-grained, contextually grounded descriptions that surpass simple category labels.

For each 3D instance, we select a set of $K=5$ optimal camera views to query. The selection is guided by a scoring function $S = \text{IoU} + \alpha \cdot \text{VisibilityRatio}$, which prioritizes views where the object is both strictly segmented (high IoU with SAM masks) and largely visible (low occlusion). The captions generated from these $K$ views are then consolidated by GPT-4~\cite{achiam2023gpt4} into a single, cohesive annotation.
\cref{fig:wordcloud} visualizes the distribution of object semantic annotations across all benchmark scenes as word clouds.

\subsection{Context-Aware Action Planning}\label{supp:action-planning}
We query the VLM for a planning update every $T=20$ simulation steps (approximately 1\,Hz), mimicking natural human decision-making latency. Unless otherwise stated, we employ Qwen2.5-VL-72B~\cite{bai2025qwen2} as the high-level planner.

\vspace{1ex}

\noindent\textbf{Prompt Design.} 
As discussed in the main paper, we first apply an open-vocabulary detector to identify the goal, highlighting it with a green bounding box. We also overlay viable direction arrows on the image to define the action space. Crucially, our system prompt is meticulously designed to induce iterative reasoning. We instruct the VLM to employ Chain-of-Thought (CoT) reasoning: it must first analyze visual observations, then formulate a high-level multi-step plan anchored to landmarks, and finally select the immediate action. The agent iteratively adjusts its plan based on a memory buffer of previous thoughts and current environmental changes, ensuring robustness against temporary goal occlusion or long-horizon deviations.
Figures~\ref{fig:prompt_part1}, \ref{fig:prompt_part2}, and \ref{fig:user_prompt} provide the detailed system prompts and user prompts in each query used in our experiments.

\vspace{1ex}

\begin{figure}[t]
    \centering
    \includegraphics[width=0.7\linewidth]{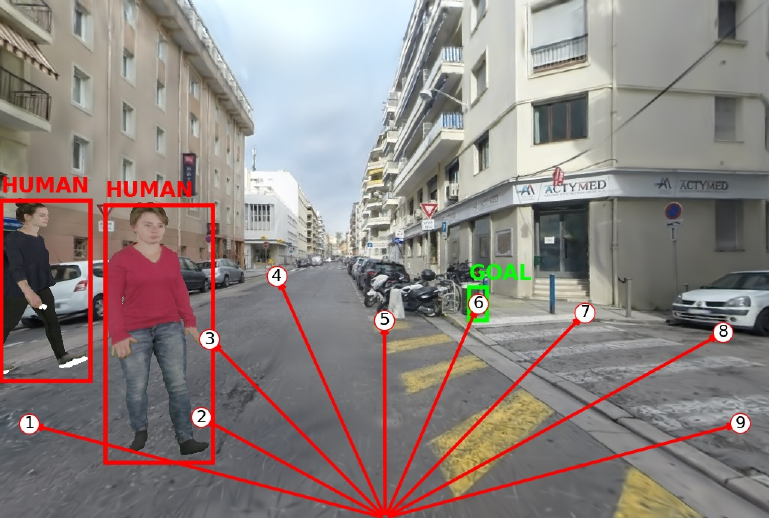}
    \vspace{-0.3ex}
    \caption{\textbf{Visual Prompting for Social Navigation.} 
    In dynamic environments, we extend our visual prompting strategy to mark pedestrians. 
    Humans are detected and highlighted with \textcolor{red}{\textbf{red}} bounding boxes labeled `HUMAN', distinct from the \textcolor{green}{\textbf{green}} `GOAL' box. 
    These explicit visual cues allow a pre-trained VLM to reason in a zero-shot manner about the spatial relationship between the agent's planned path and nearby moving obstacles, thereby supporting safer navigation decisions.}
    \label{fig:social_visual_example}
    \vspace{-3ex}
    
\end{figure}

% define colors
\definecolor{promptbg}{RGB}{245, 247, 250}
\definecolor{promptframe}{RGB}{60, 80, 100}
\definecolor{cmdblue}{RGB}{0, 80, 180}
\definecolor{alertred}{RGB}{180, 30, 30}
\definecolor{jsonkey}{RGB}{0, 100, 80}
\definecolor{strgray}{RGB}{80, 80, 80}

\newcommand{\SectionTitle}[1]{\vspace{0.3em}\noindent{\small\bfseries\color{black} #1}\vspace{0.15em}}
\newcommand{\SubTitle}[1]{\vspace{0.15em}\noindent{\footnotesize\bfseries\color{cmdblue} #1}}
\newcommand{\Highlight}[1]{\textbf{\color{alertred}#1}}
\newcommand{\JsonKey}[1]{\texttt{\color{jsonkey}#1}}

\noindent\textbf{Extension to Social-Aware Navigation.}
For the Social Navigation benchmark (SocialNav), where the agent navigates shared spaces populated by moving pedestrians, we adapt our prompt design to prioritize safety and collision avoidance. 
As detailed in the \textbf{system prompt extensions} in~\cref{fig:prompt_social}, we incorporate a dedicated \textit{Dynamic Obstacle Handling} module. 
Correspondingly, as illustrated by the \textbf{visual prompt example} in~\cref{fig:social_visual_example}, we augment the perception mechanism to detect humans (highlighted with red bounding boxes) and inject their spatial information into the \texttt{dynamic\_obstacles} field.
Critically, we introduce a \texttt{stop\_and\_wait} primitive into the action space. 
The VLM is explicitly instructed to assess collision risks in its Chain-of-Thought reasoning and is forced to select \texttt{stop\_and\_wait} if all viable paths towards the goal are temporarily obstructed by pedestrians, ensuring socially compliant behavior.

\subsection{Controllable Human Motion Generation}\label{supp:motion-gen}
The high-level planning module produces action commands along with sparse guiding signals (\emph{i.e}, motion prompt, trajectory, orientation, and previous poses for temporal alignment). While training-based spatially controllable motion generators~\cite{xie2023omnicontrol, wan2024tlcontrol, shafir2023priormdm} can condition on arbitrary joint positions, their output is not directly compatible with SMPL~\cite{loper2023smpl, pavlakos2019smplx} parameters, requiring expensive test-time optimization for avatar animation~\cite{hu2024gaussianavatar}. Thus, on the consideration of scalability, we therefore build upon MDM-SMPL~\cite{petrovich2024stmc}, which directly predicts SMPL pose parameters $(\boldsymbol{\theta}, \boldsymbol{M})$ rather than the joint-rotation features of HumanML3D~\cite{guo2022humanml3d}.

\vspace{1ex}

\noindent\textbf{Pose Representation.}
The diffusion model operates on a canonicalized representation~\cite{petrovich2024stmc} that encodes root gravity-axis height, local trajectory velocities, angular velocities, local joint rotations (in 6D format), and local joint positions. At each denoising step, this representation is seamlessly decoded back into SMPL poses and root translations for computing the guidance loss.

\vspace{1ex}

\noindent\textbf{Sampling Strategy.}
We run $T{=}100$ DDPM denoising steps with classifier-free guidance (CFG)~\cite{ho2022cfg}. At each step, the denoiser produces both a conditional output $\hat{\boldsymbol{x}}_0^{c}$ (conditioned on the text embedding of the motion type $\hat{\tau}$) and an unconditional output $\hat{\boldsymbol{x}}_0^{u}$. The final prediction is:
\begin{equation}
    \hat{\boldsymbol{x}}_0 = \hat{\boldsymbol{x}}_0^{u} + w \cdot (\hat{\boldsymbol{x}}_0^{c} - \hat{\boldsymbol{x}}_0^{u}),
\end{equation}
where $w=5.0$ denotes the CFG weight. The posterior mean $\boldsymbol{\mu}_k$ is then computed via the standard DDPM posterior $q(\boldsymbol{x}_{k-1} \mid \boldsymbol{x}_k, \hat{\boldsymbol{x}}_0)$.

\vspace{1ex}

\noindent\textbf{Training-Free Spatial Guidance.}
Spatial guidance is applied to $\boldsymbol{\mu}_k$ only when the diffusion timestep $k \leq t_0 T$ (we set $t_0{=}0.25$, corresponding to step 25 out of 100), \ie, during the detail-refining stage of the reverse process. At each guided step, we perform $N_g{=}20$ gradient-descent iterations:
\begin{equation}
    \boldsymbol{\mu}_k \leftarrow \boldsymbol{\mu}_k - \alpha \nabla_{\boldsymbol{\mu}_k} \mathcal{L}(\boldsymbol{\mu}_k;\, \boldsymbol{W},\, \boldsymbol{x}^{0}_{\text{prev}}),
\end{equation}
where $\alpha{=}1.0$ is the guidance scale. The composite loss $\mathcal{L}$ consists of three terms:
\begin{equation}
    \mathcal{L} = \mathcal{L}_{\text{traj}} + \lambda_{\text{orient}} \mathcal{L}_{\text{orient}} + \lambda_{\text{init}} \mathcal{L}_{\text{init}}.
\end{equation}
Each term is detailed below:
\vspace{0.5ex}

\noindent\textit{1) Trajectory guidance} $\mathcal{L}_{\text{traj}}$: enforces the generated root translation to follow the waypoints $\boldsymbol{W}$ via point-to-point matching. We compute the loss on every $K{=}5$-th keyframe (plus the last frame) rather than all frames, balancing motion naturalness with trajectory fidelity: $\mathcal{L}_{\text{traj}} = \sum_{i \in \mathcal{K}} \|\boldsymbol{t}^i - \boldsymbol{w}^i\|_2$, where $\boldsymbol{t}^i$ is the predicted root translation and $\mathcal{K}$ denotes the sampled keyframe indices.

\vspace{0.5ex}

\noindent\textit{2) Orientation guidance} $\mathcal{L}_{\text{orient}}$: aligns the generated heading angle $\phi^i$ with the reference heading $\hat{\phi}^i$ derived from the waypoint trajectory via an L1 loss: $\mathcal{L}_{\text{orient}} = \sum_{i \in \mathcal{K}} |\phi^i - \hat{\phi}^i|$. We set $\lambda_{\text{orient}}{=}1.0$.

\vspace{0.5ex}

\noindent\textit{3) Initial pose alignment} $\mathcal{L}_{\text{init}}$: enforces temporal continuity across consecutive motion clips by aligning the first frame's body pose $\boldsymbol{\theta}^{0}$ with the last pose of the previous clip $\boldsymbol{x}^{0}_{\text{prev}}$: $\mathcal{L}_{\text{init}} = \|\boldsymbol{\theta}^{0} - \boldsymbol{x}^{0}_{\text{prev}}\|_2$. We set $\lambda_{\text{init}}{=}2.0$.

\vspace{1.5ex}
All guidance computations are performed in the canonical coordinate frame. After denoising, the generated motion is transformed to the global frame via the transformation derived from the agent's initial global orientation and position.

\subsection{Extension to More Scenes and Avatars}\label{supp:more-scenes}

Beyond quantitative benchmarks, we provide extensive qualitative results to demonstrate the versatility and generalizability of our framework across a diverse range of environments. These include outdoor urban streets (\eg, roads, cities, towns), commercial-grade captures (\eg, churches), and indoor spaces (\eg, living rooms, restaurants). By populating these scenes with animatable avatars of diverse appearances, body shapes, and ethnicities, our visually-grounded humanoids integrate seamlessly into any 3DGS-based environment for realistic, goal-directed behavior synthesis. Please see our project website at \url{https://alvinyh.github.io/VGHuman/} for full demonstrations.

\subsection{Human-Scene Contact}\label{supp:human-scene}
\noindent\textbf{Compositional Rendering.}
We employ PyTorch3D~\cite{ravi2020pytorch3d} to render obstacle meshes, utilizing z-buffering~\cite{catmull1974subdivision} to achieve depth-aware composition between the GS-based scene/human representations and the static obstacles.

\vspace{1ex}

\noindent\textbf{Coordinate Alignment.}
To ensure physically plausible interaction and minimize interpenetration, the world layer must be gravity aligned, such that the support surface is horizontal and the vertical direction coincides with $+Z$. We address this for two types of scenes:

\textit{1) Multi-view reconstructed scenes.}
For scenes reconstructed from multi-view image captures, we adopt the COLMAP re-orientation procedure~\cite{schoenberger2016sfm,schoenberger2016mvs}, which leverages the Manhattan-world assumption to recover an upright scene frame directly from SfM.

\textit{2) Off-the-shelf 3DGS assets.}
For external assets where the original reconstruction pipeline is unavailable, the provided coordinates are not guaranteed to be gravity aligned. In such cases, we apply a geometry-based heuristic to the accompanying collision mesh: we first apply a coarse axis flip when the scene uses a $Y$-up convention, then fit a dominant ground plane to low-lying mesh vertices with RANSAC, rotate the scene so that the estimated ground normal aligns with $+Z$, and translate the scene so that the fitted ground lies at $z=0$.

\textit{Unified transformation.}
This rigid transformation is then applied to both the 3DGS representation and the collision mesh, establishing a consistent coordinate frame for rendering, collision detection, and human placement. In our current implementation, we approximate the support surface as a flat plane at a fixed height; extending this formulation to handle complex terrain via height maps remains as future work. As illustrated in~\cref{fig:axes_alignment}, this alignment eliminates the arbitrary tilt of the raw scene frame, rendering ground-height estimation well-defined for downstream HSI tasks.

\begin{figure}[t]
  \centering
  \begin{subfigure}{0.48\linewidth}
    \centering
    \includegraphics[width=\linewidth]{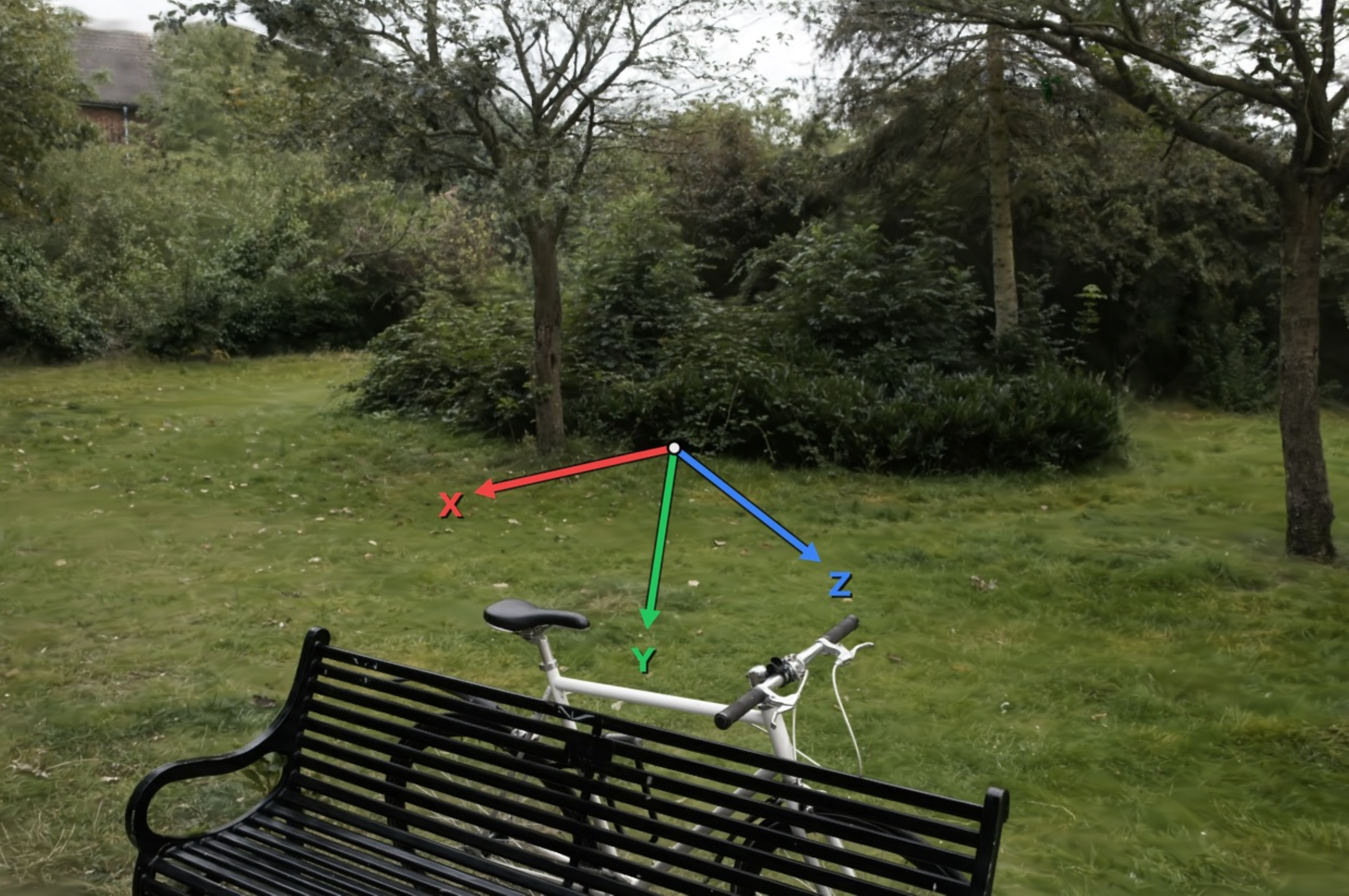}
    \caption{Raw asset frame}
    \label{fig:before_axes}
  \end{subfigure}
  \hfill
  \begin{subfigure}{0.48\linewidth}
    \centering
    \includegraphics[width=\linewidth]{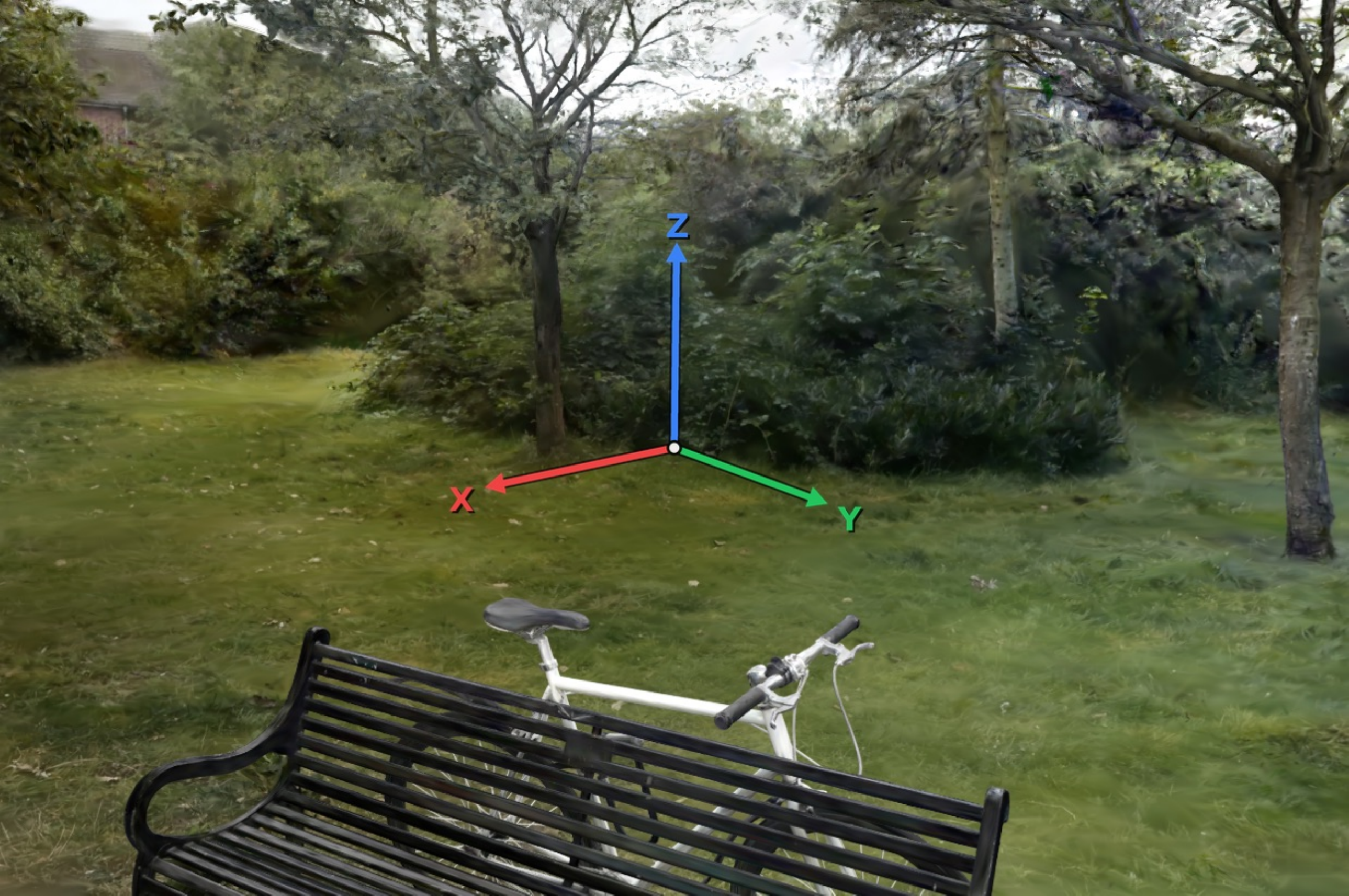}
    \caption{Aligned frame}
    \label{fig:after_axes}
  \end{subfigure}
  \vspace{-1.5ex}
  \caption{\textbf{World-frame alignment for human-scene contact.} Left: raw asset with unaligned orientation. Right: aligned coordinates used for ground-height estimation and human placement.}
  \label{fig:axes_alignment}
\end{figure}

\vspace{1ex}

\noindent\textbf{Collision Detection.}
We utilize a KDTree~\cite{bentley1975multidimensional} for real-time distance computation between humans and the static scene. For human-obstacle interaction, we employ the Kaolin library~\cite{KaolinLibrary} to calculate the unsigned distance from human vertices to obstacle meshes. In both cases, a collision is declared if the distance for more than 10\% of the human vertices falls below a threshold of $0.05m$. For human-human collision detection, we calculate the pairwise Euclidean distance between the root joints of agent pairs; a collision is registered if this distance is less than $0.5m$.

\section{Benchmark and Task Design} \label{supp_sec_bench}

\subsection{Scene Selection and Preprocessing}\label{supp:scene-select}
Our primary benchmark is \textit{SmallCity}, a large-scale street scene from Hier-GS~\cite{kerbl2024hiergs} spanning a $100\text{m} \times 100\text{m}$ area. Its high reconstruction quality, massive interaction range, and complete raw captures support both world-layer semantic grounding and agent-layer quantitative navigation evaluations.
Crucially, our Agent Layer is compatible with any GS-based environment regardless of the reconstruction pipeline.
To evaluate the broad applicability of our approach, we include 2 outdoor scenes from XGRIDS~\cite{xgrids} and 4 indoor scenes from SAGE-3D~\cite{miao2025sage3d}.
While XGRIDS provides the highest visual fidelity via professional capturing hardware, it lacks the raw images and camera poses required for automated world-layer processing; we therefore manually annotate semantic goals for these scenes.
SAGE-3D complements the benchmark with indoor environments and provides ready-to-use 3D bounding box annotations for navigation targets.
Finally, we provide qualitative demonstrations on several additional scenes~\cite{wilson2023argoverse, xiao2021pandaset, barron2021mip, ling2024dl3dv} and assets from SuperSplat~\cite{supersplat} and Pointcosm~\cite{pointcosm}, which are excluded from quantitative benchmarking as they do not satisfy all required criteria.

We visualize the bird's-eye-view (BEV) of our 3 outdoor benchmark scenes in \cref{fig:bev_outdoor}, showing both the 3D Gaussian Splatting rendering (left) and the extracted collision mesh (right). Similarly, \cref{fig:bev_indoor} presents the 4 indoor scenes from SAGE-3D. These visualizations illustrate the diversity in scene scale, layout, and semantic content across our benchmark environments.

\begin{figure*}[htbp]
  \centering
  % Row 1: SmallCity
  \begin{subfigure}{0.49\linewidth}
    \centering
    \includegraphics[width=\linewidth]{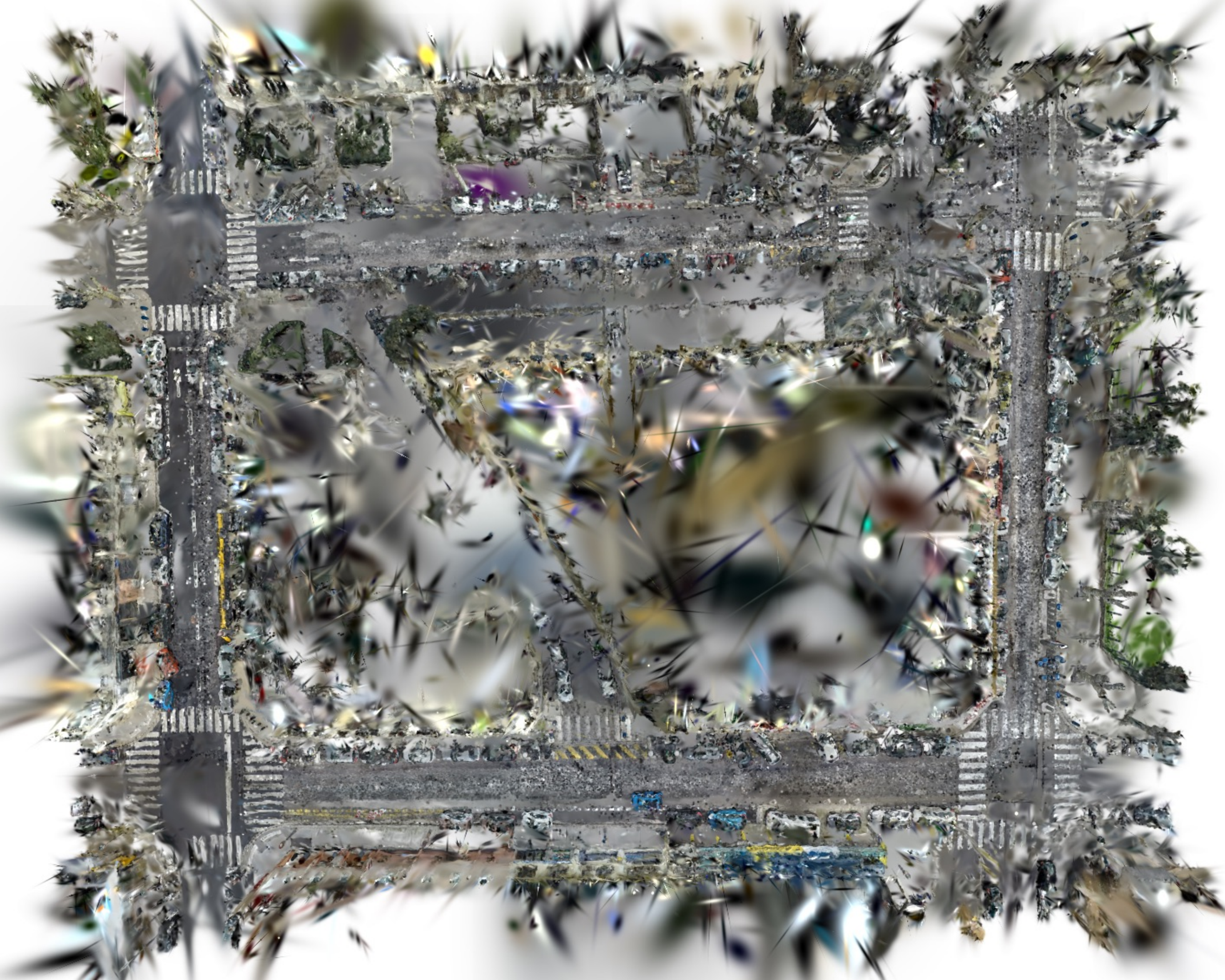}
  \end{subfigure}
  \hfill
  \begin{subfigure}{0.49\linewidth}
    \centering
    \includegraphics[width=\linewidth]{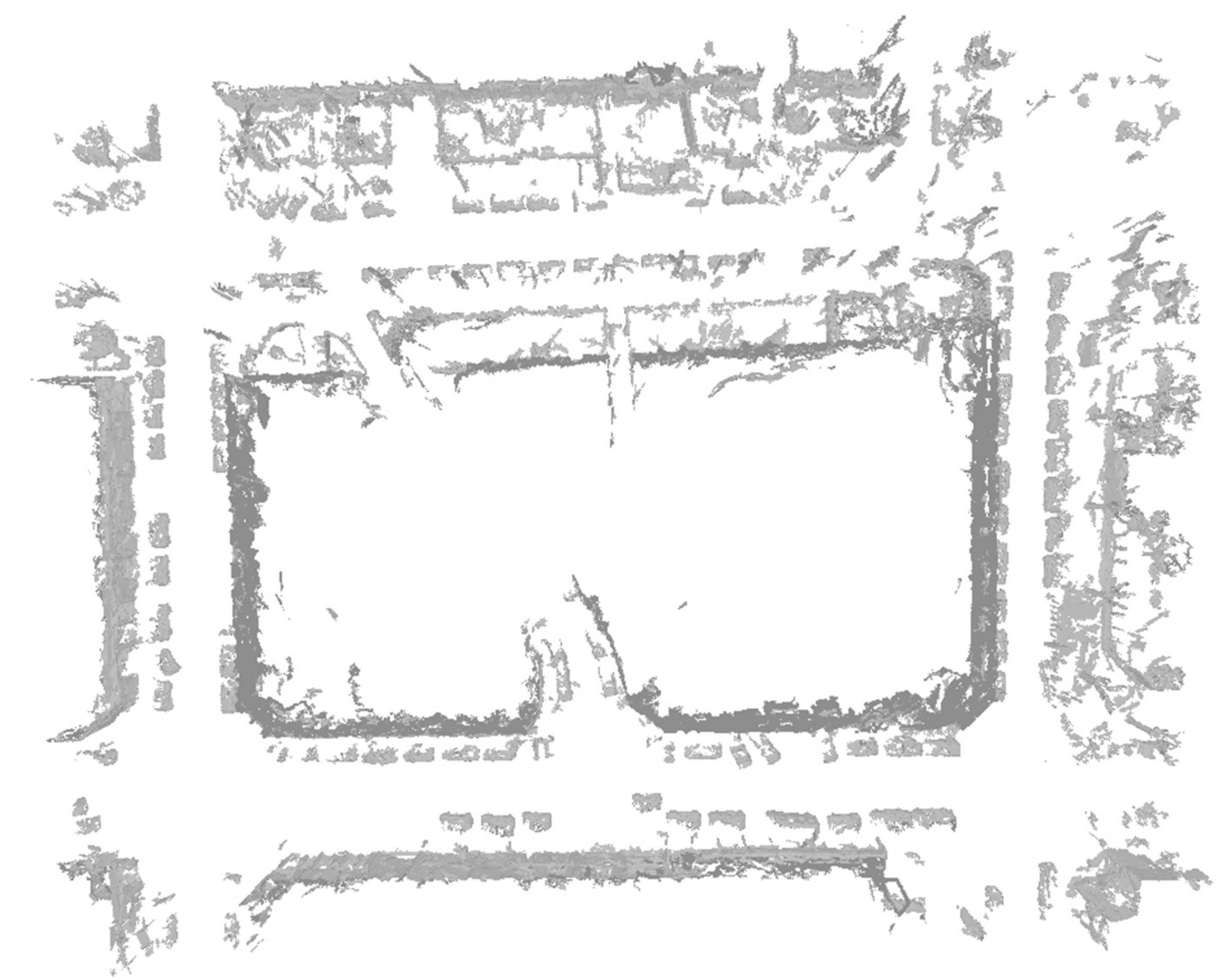}
  \end{subfigure}
  \vspace{-1ex}
  \begin{center}\small\textit{SmallCity}~\cite{kerbl2024hiergs}\end{center}

  \vspace{10pt}
  % Row 2: Church
  \begin{subfigure}{0.49\linewidth}
    \centering
    \includegraphics[width=\linewidth]{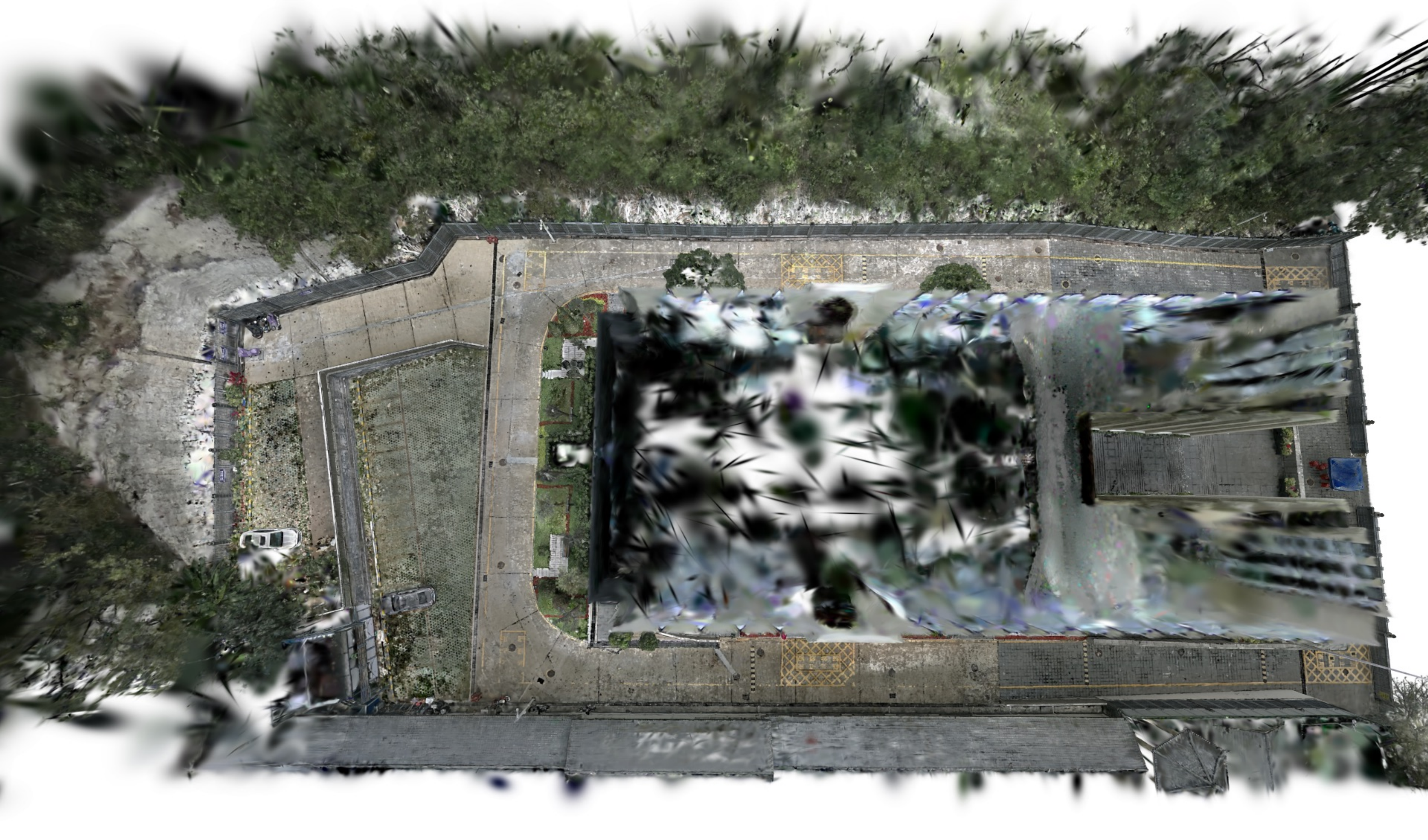}
  \end{subfigure}
  \hfill
  \begin{subfigure}{0.49\linewidth}
    \centering
    \includegraphics[width=\linewidth]{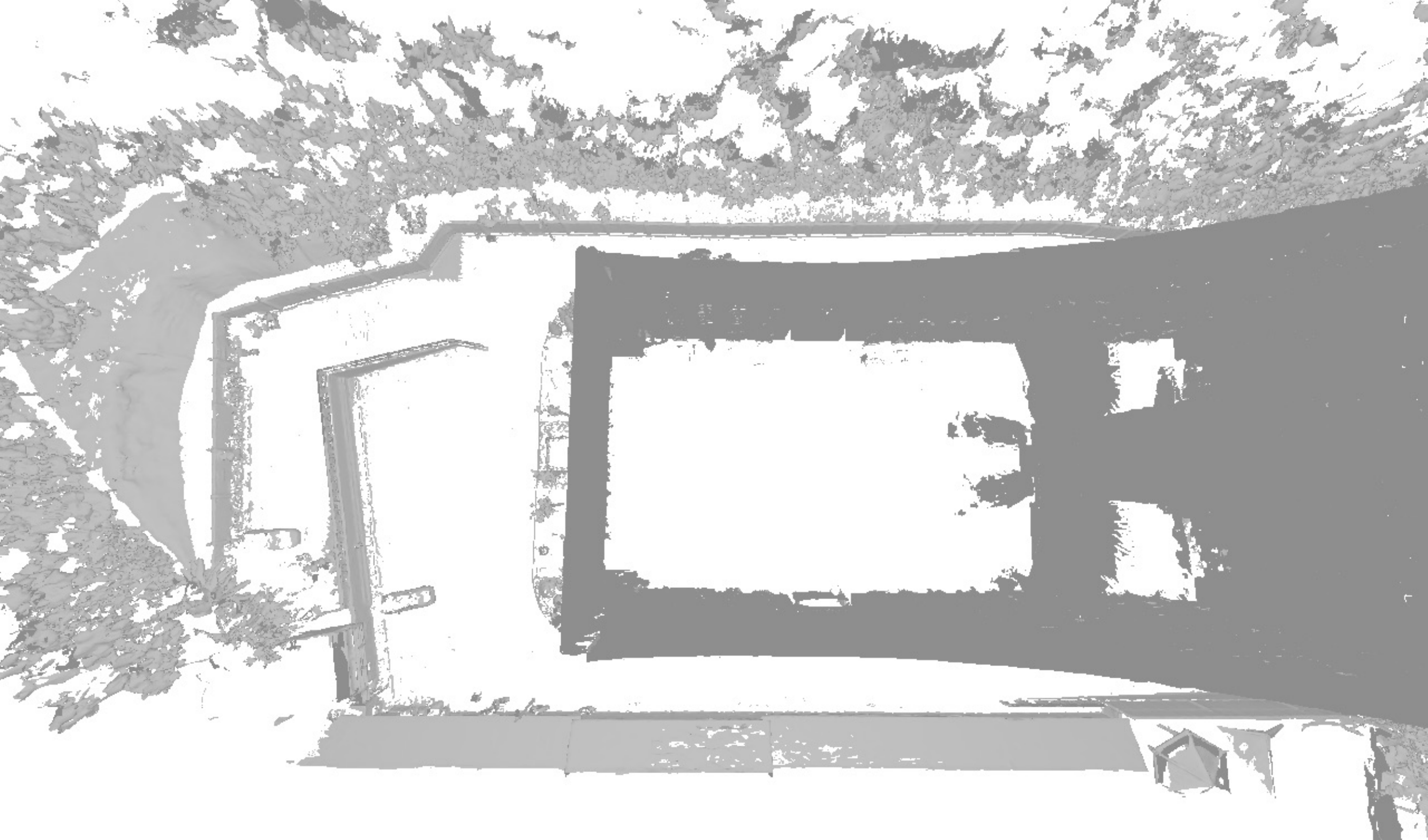}
  \end{subfigure}
  \vspace{-2ex}
  \begin{center}\small\textit{Church} (XGRIDS~\cite{xgrids})\end{center}

  \vspace{10pt}
  % Row 3: Town
  \begin{subfigure}{0.49\linewidth}
    \centering
    \includegraphics[width=\linewidth]{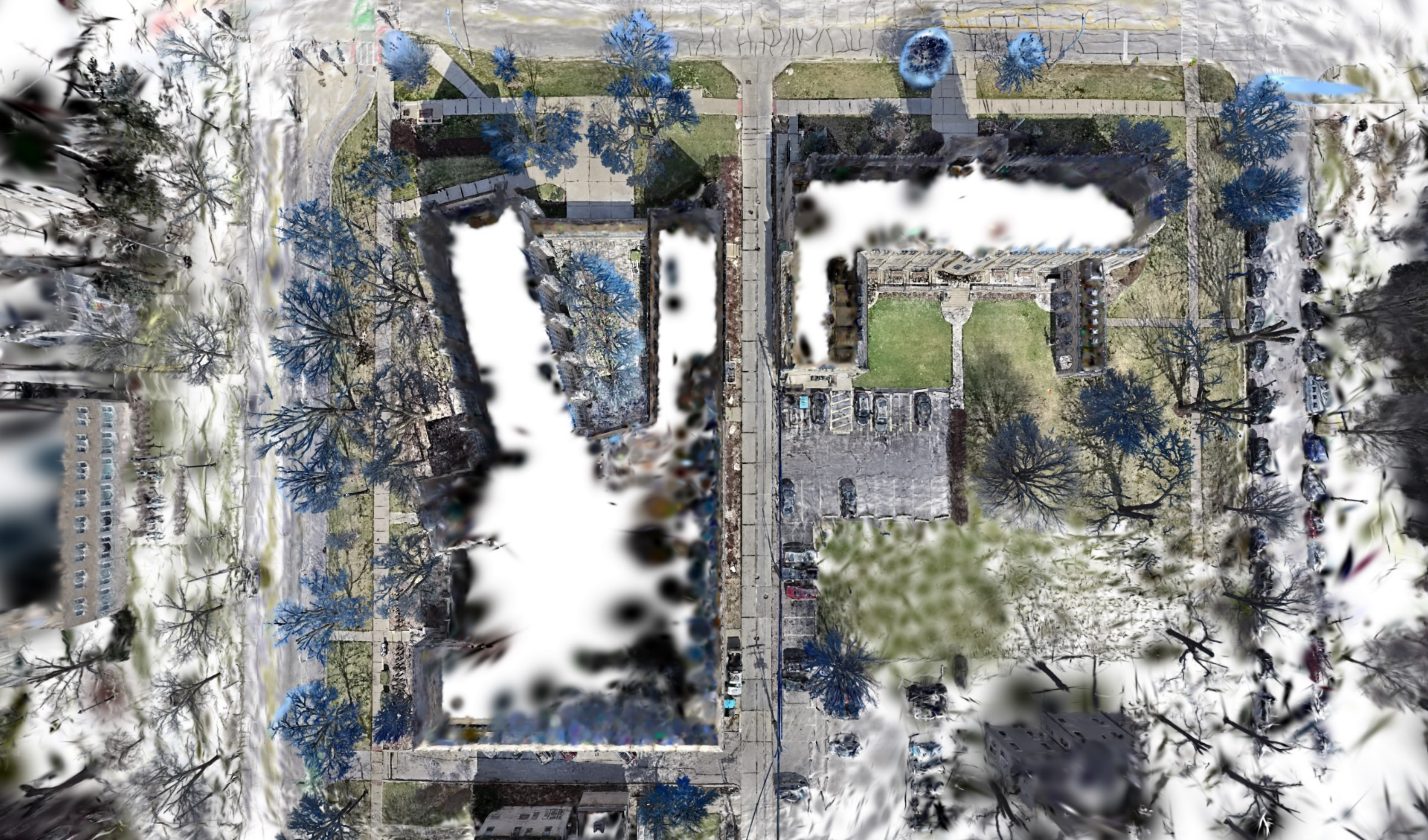}
    % \caption*{3DGS Rendering}
  \end{subfigure}
  \hfill
  \begin{subfigure}{0.49\linewidth}
    \centering
    \includegraphics[width=\linewidth]{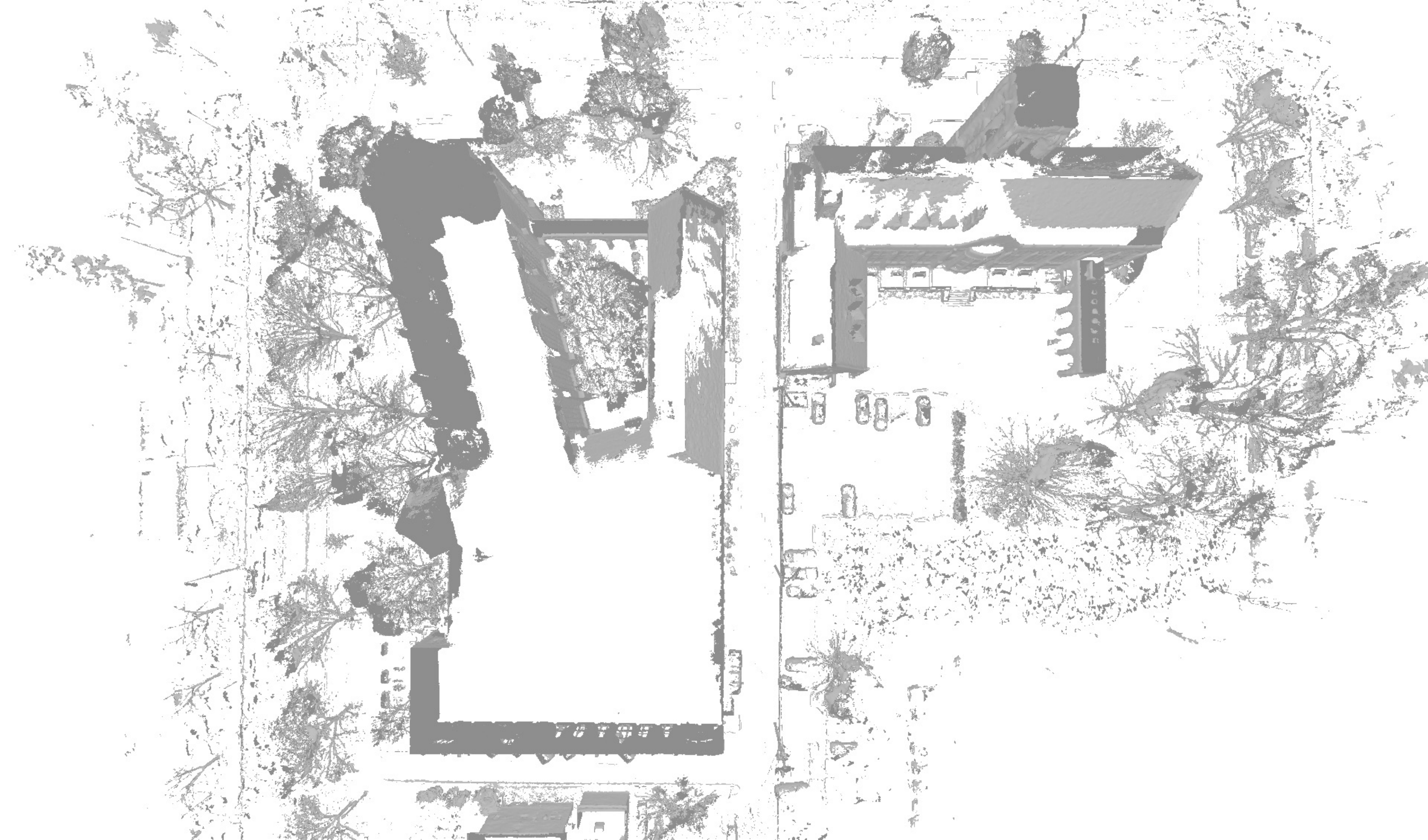}
    % \caption*{Collision Mesh}
  \end{subfigure}
  \vspace{-1ex}
  \begin{center}\small\textit{Town} (XGRIDS~\cite{xgrids})\end{center}

  \vspace{1ex}
  \caption{\textbf{Bird's-eye-view of outdoor benchmark scenes.} Each row shows a scene with its 3DGS rendering (left) and extracted collision mesh (right). From top to bottom: \textit{SmallCity}~\cite{kerbl2024hiergs}, \textit{Church} and \textit{Town} from XGRIDS~\cite{xgrids}. 
  }
  \label{fig:bev_outdoor}
\end{figure*}

\begin{figure*}[htbp]
  \centering
  % Row 1: Restaurant
  \begin{subfigure}{0.45\linewidth}
    \centering
    \includegraphics[width=\linewidth]{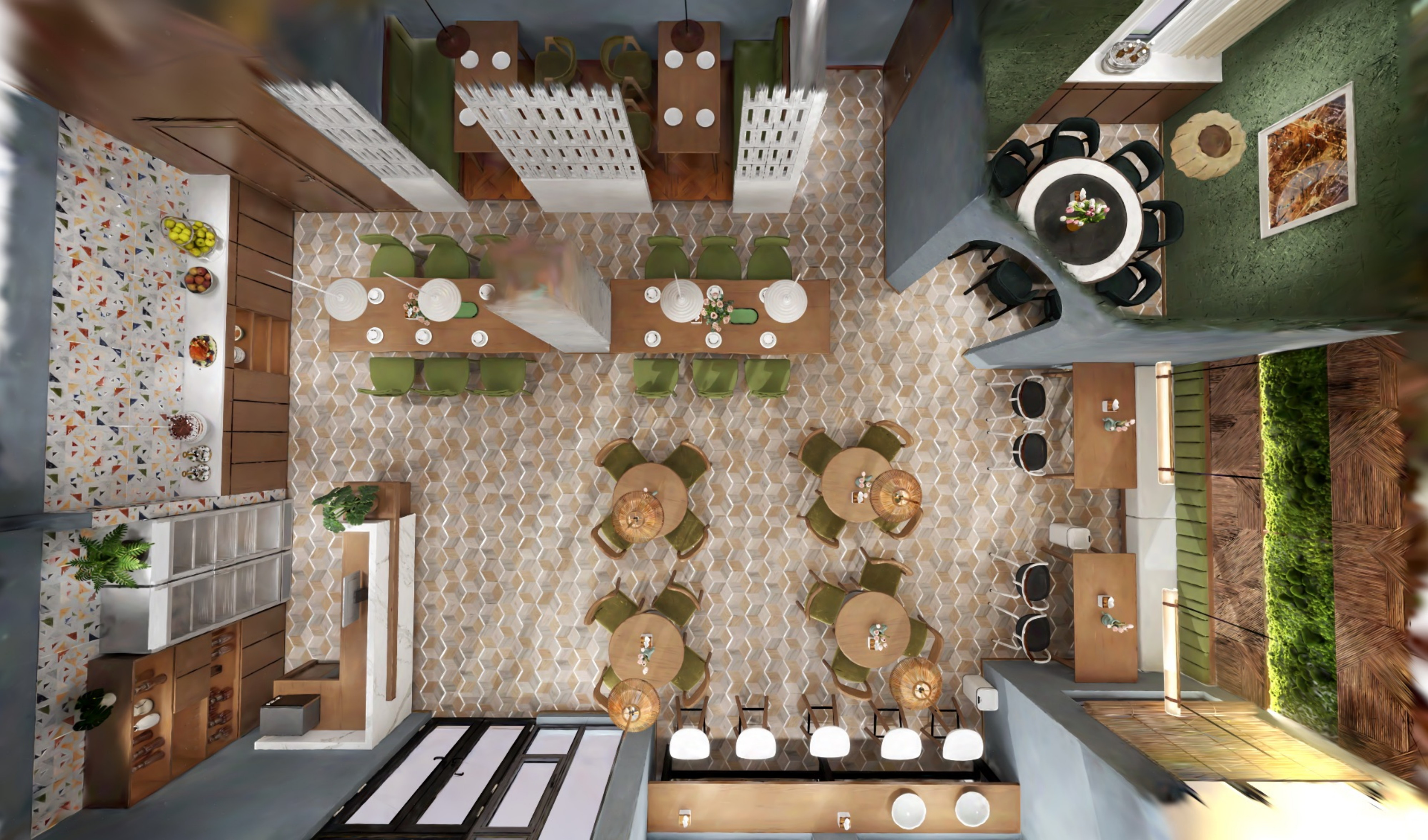}
  \end{subfigure}
  \hspace{0.03\linewidth}
  \begin{subfigure}{0.45\linewidth}
    \centering
    \includegraphics[width=\linewidth]{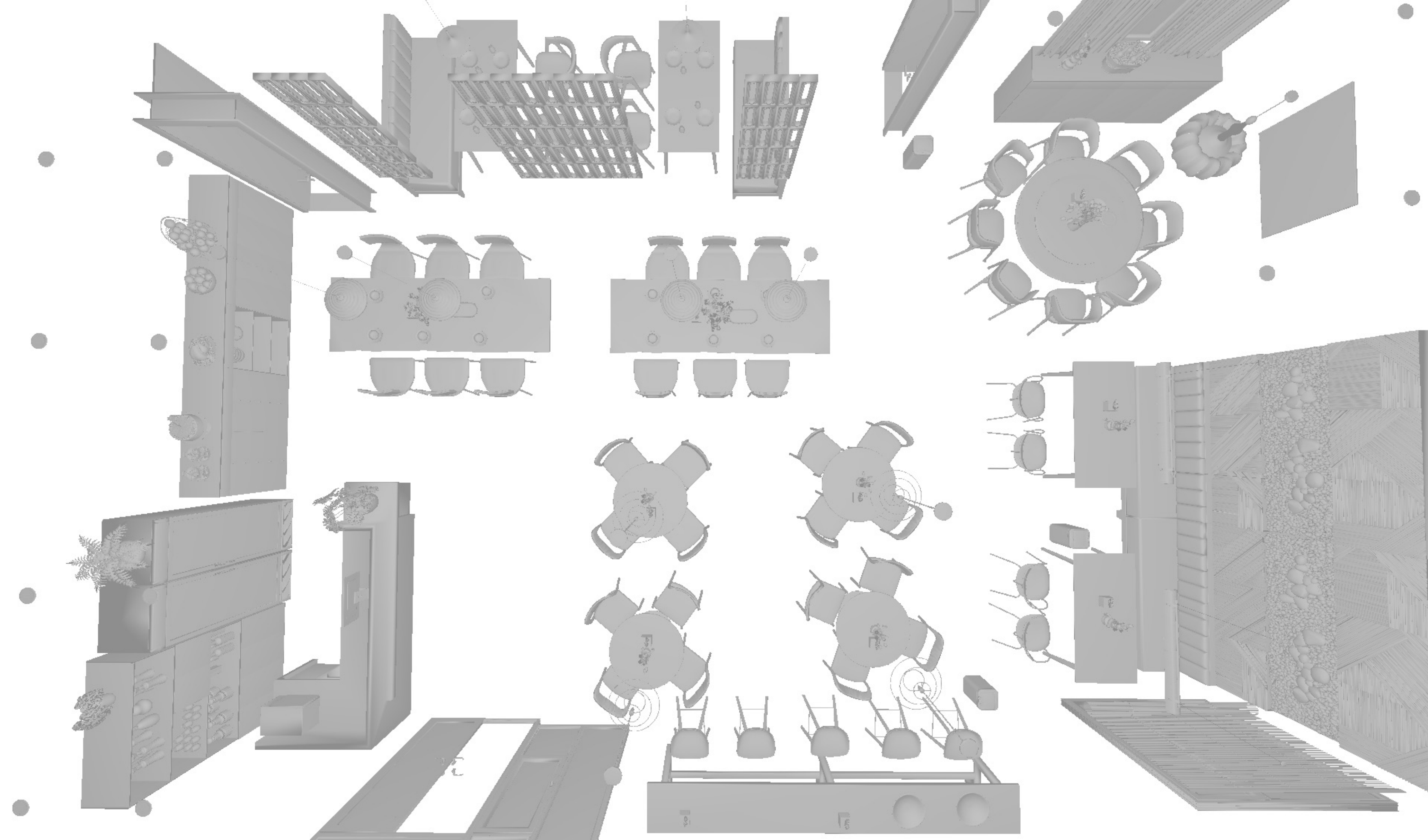}
  \end{subfigure}
  \vspace{-2ex}
  \begin{center}\small\textit{Scene 1: Restaurant}\end{center}

  \vspace{4pt}
  % Row 2: Fast-food Restaurant
  \begin{subfigure}{0.45\linewidth}
    \centering
    \includegraphics[width=\linewidth]{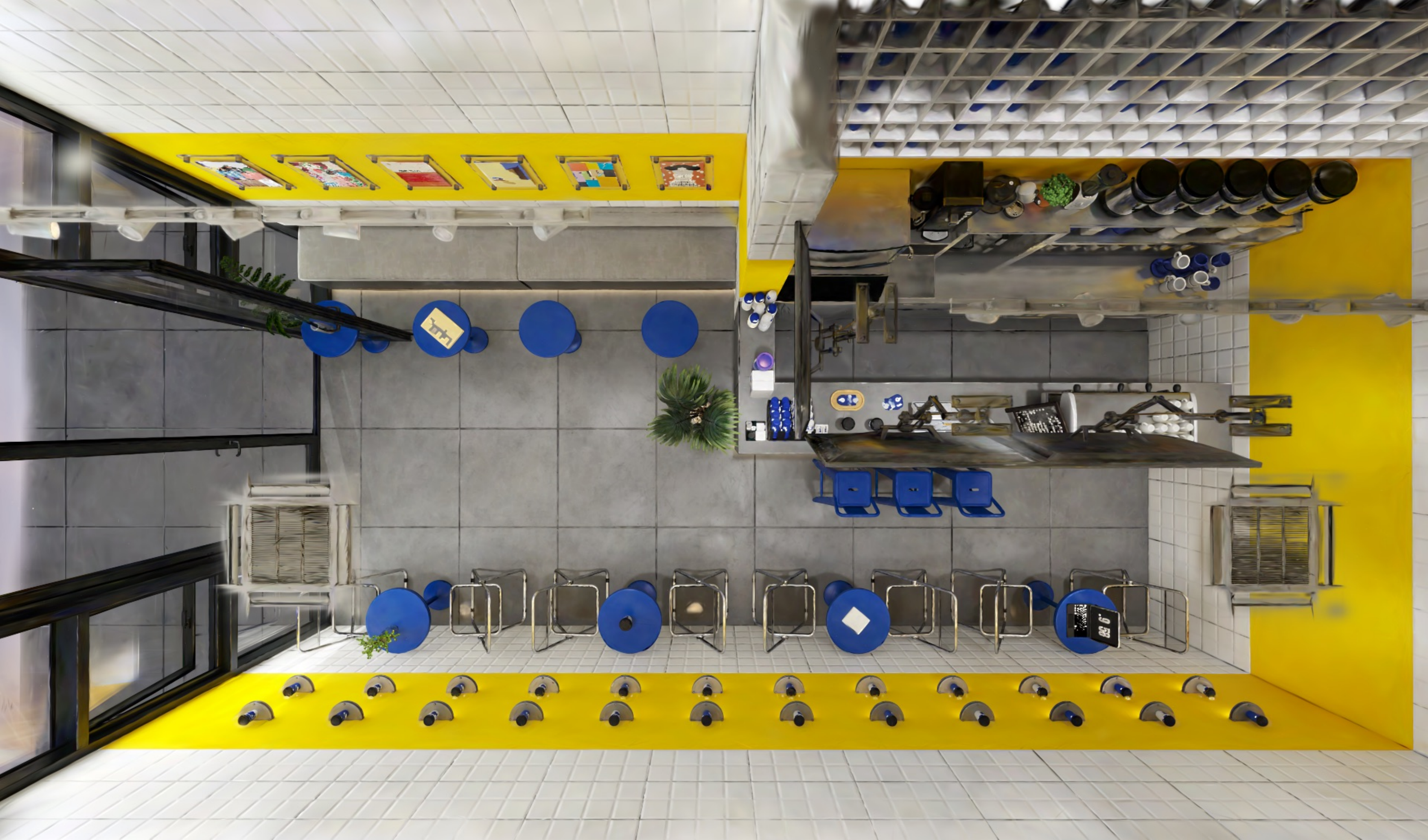}
  \end{subfigure}
  \hspace{0.03\linewidth}
  \begin{subfigure}{0.45\linewidth}
    \centering
    \includegraphics[width=\linewidth]{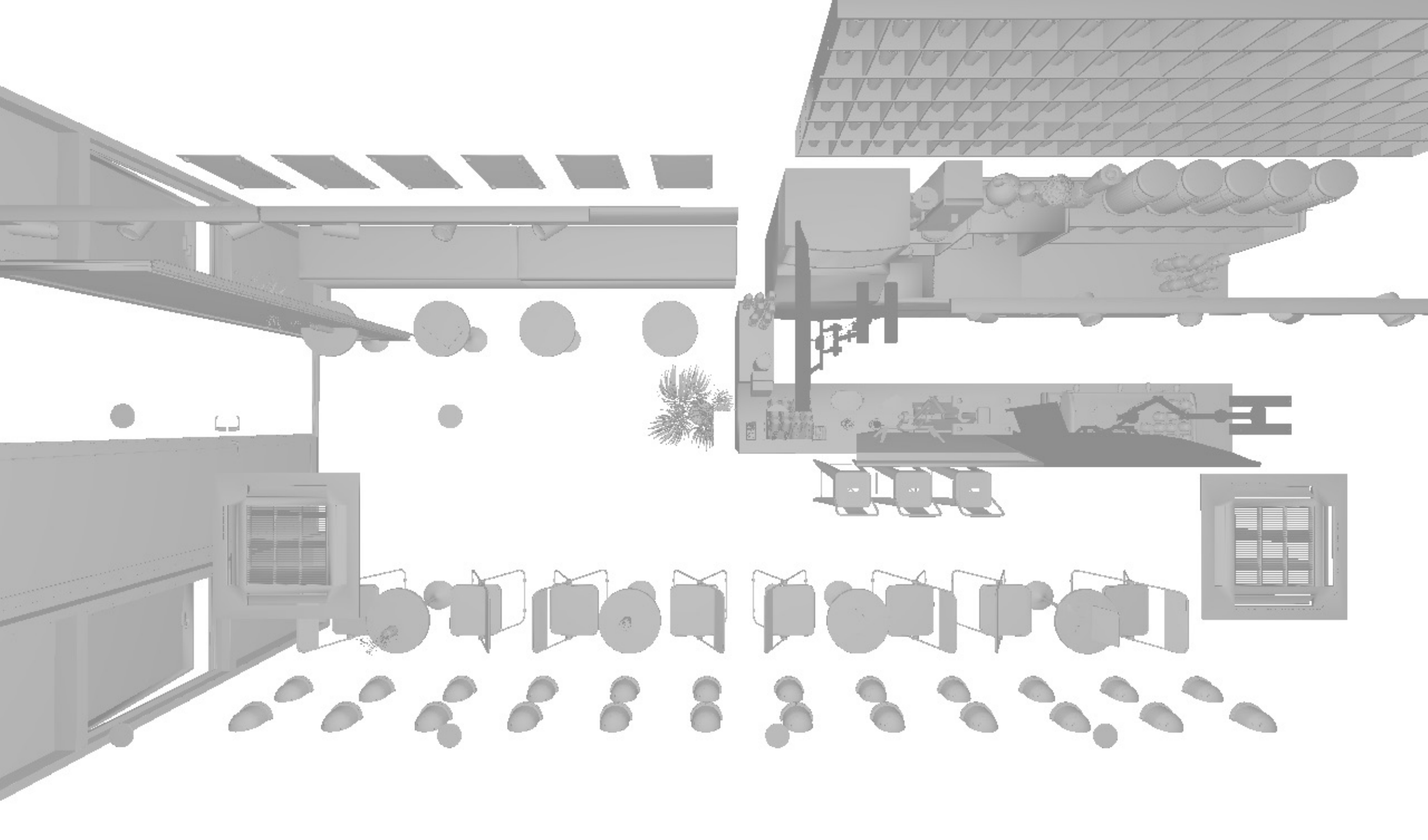}
  \end{subfigure}
  \vspace{-2ex}
  \begin{center}\small\textit{Scene 2: Tea Bar}\end{center}

  \vspace{4pt}
  % Row 3: Children's Room
  \begin{subfigure}{0.45\linewidth}
    \centering
    \includegraphics[width=\linewidth]{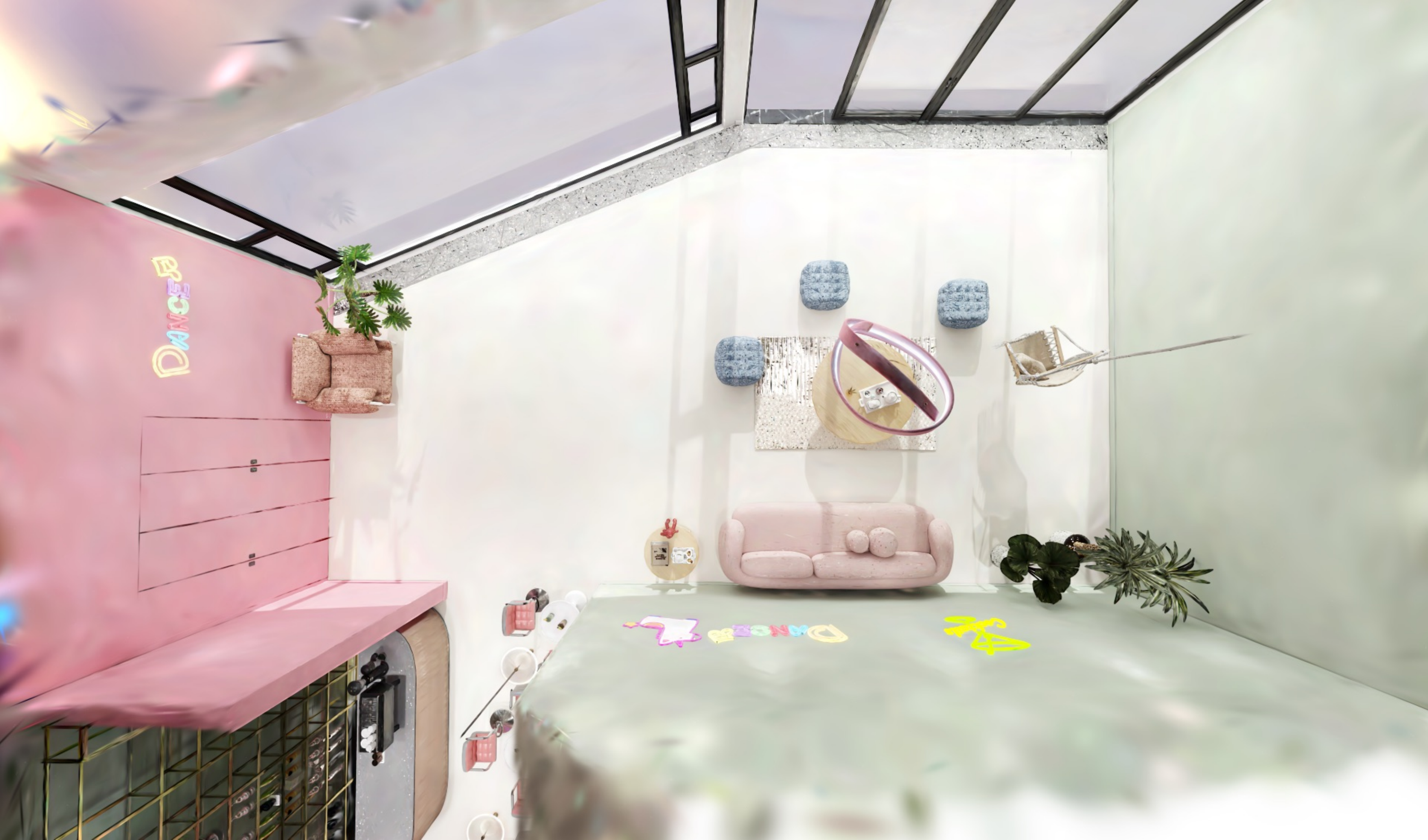}
  \end{subfigure}
 \hspace{0.03\linewidth}
  \begin{subfigure}{0.45\linewidth}
    \centering
    \includegraphics[width=\linewidth]{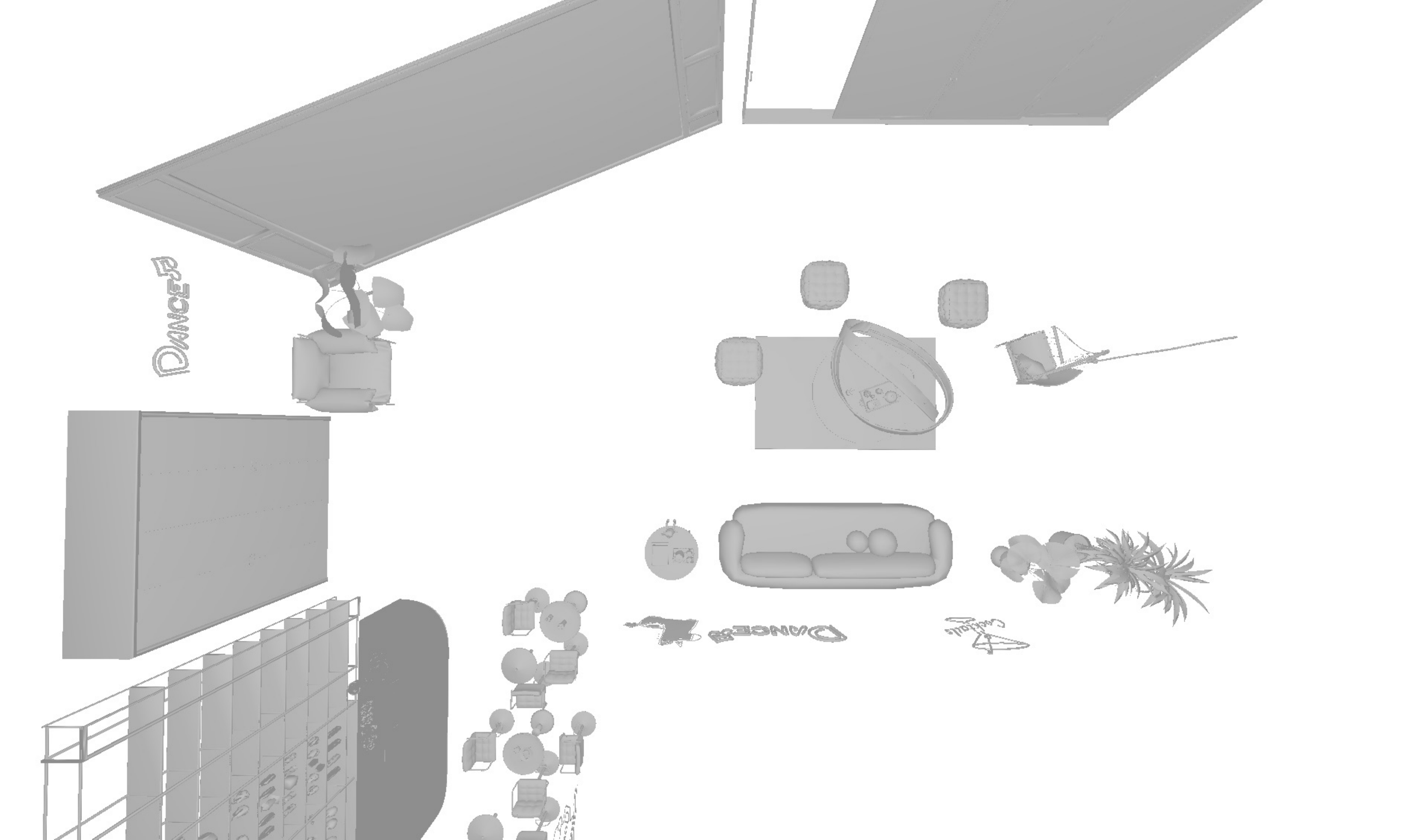}
  \end{subfigure}
  \vspace{-2ex}
  \begin{center}\small\textit{Scene 3: Living Room}\end{center}

  \vspace{4pt}
  % Row 4: Retail Store
  \begin{subfigure}{0.45\linewidth}
    \centering
    \includegraphics[width=\linewidth]{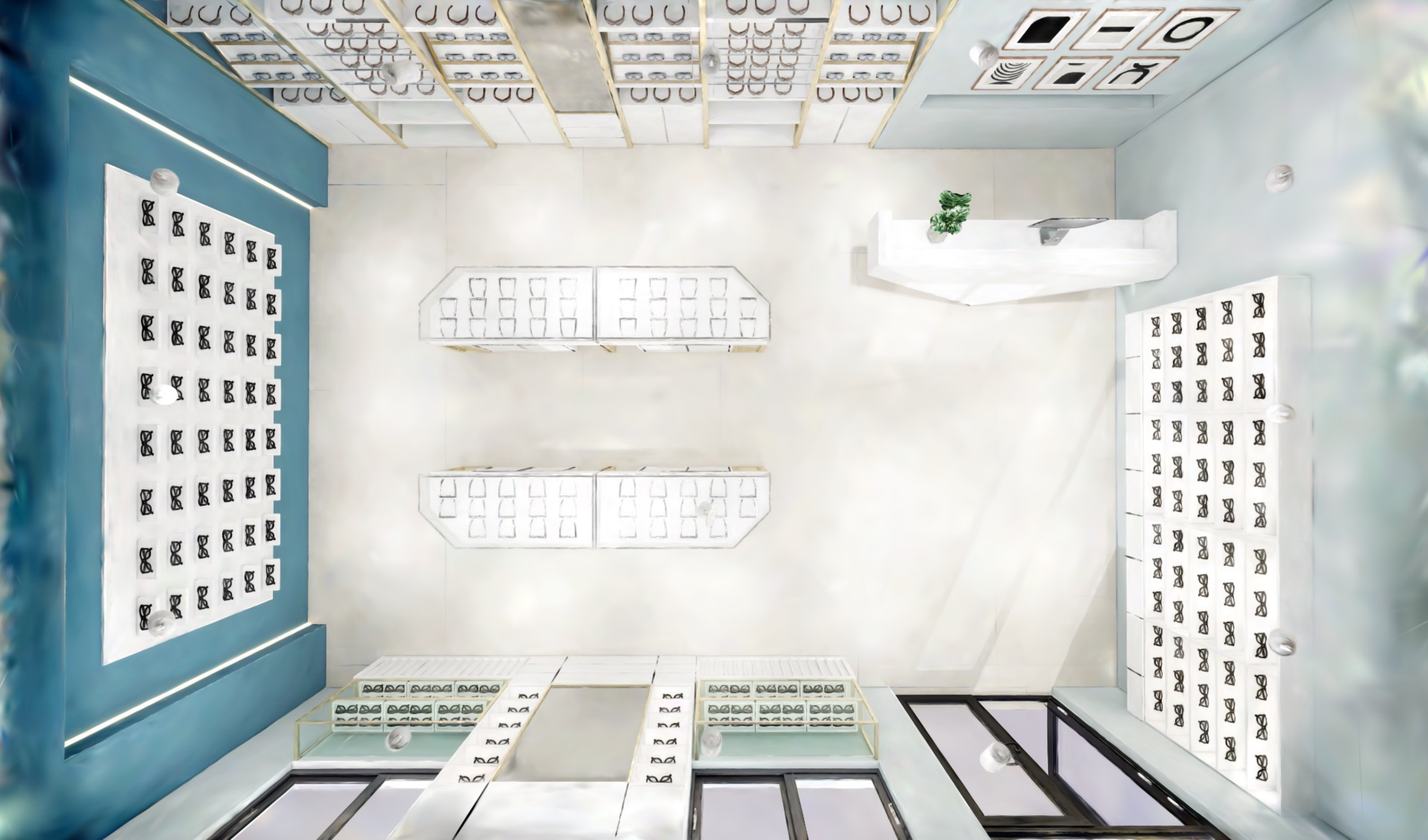}
    % \caption*{3DGS Rendering}
  \end{subfigure}
  \hspace{0.03\linewidth}
  \begin{subfigure}{0.45\linewidth}
    \centering
    \includegraphics[width=\linewidth]{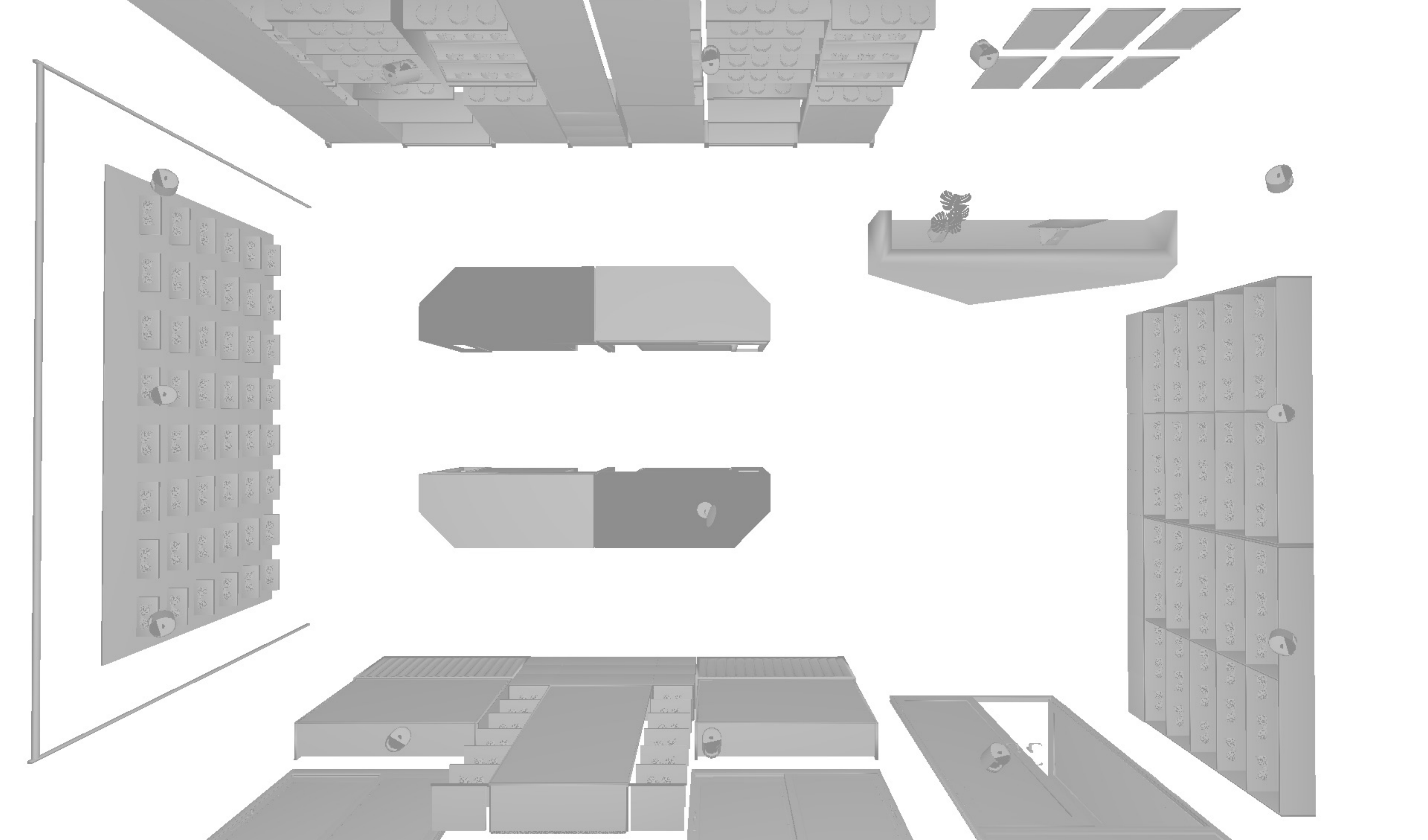}
    % \caption*{Collision Mesh}
  \end{subfigure}
  \vspace{-2ex}
  \begin{center}\small\textit{Scene 4: Optical Store}\end{center}

  \vspace{-2ex}
  \caption{\textbf{Bird's-eye-view of indoor benchmark scenes from SAGE-3D~\cite{miao2025sage3d}.} Four room-scale indoor environments used for agent-layer generalization evaluation, covering diverse functional spaces. Left: 3DGS rendering. Right: collision mesh.}
  \label{fig:bev_indoor}
\end{figure*}

\subsection{Task Setup and Evaluation Metrics}\label{supp:task-setup}
\begin{figure*}[htbp]
  \centering
  % Row 1: SimNav
  \begin{subfigure}{\linewidth}
    \centering
    \includegraphics[width=\linewidth]{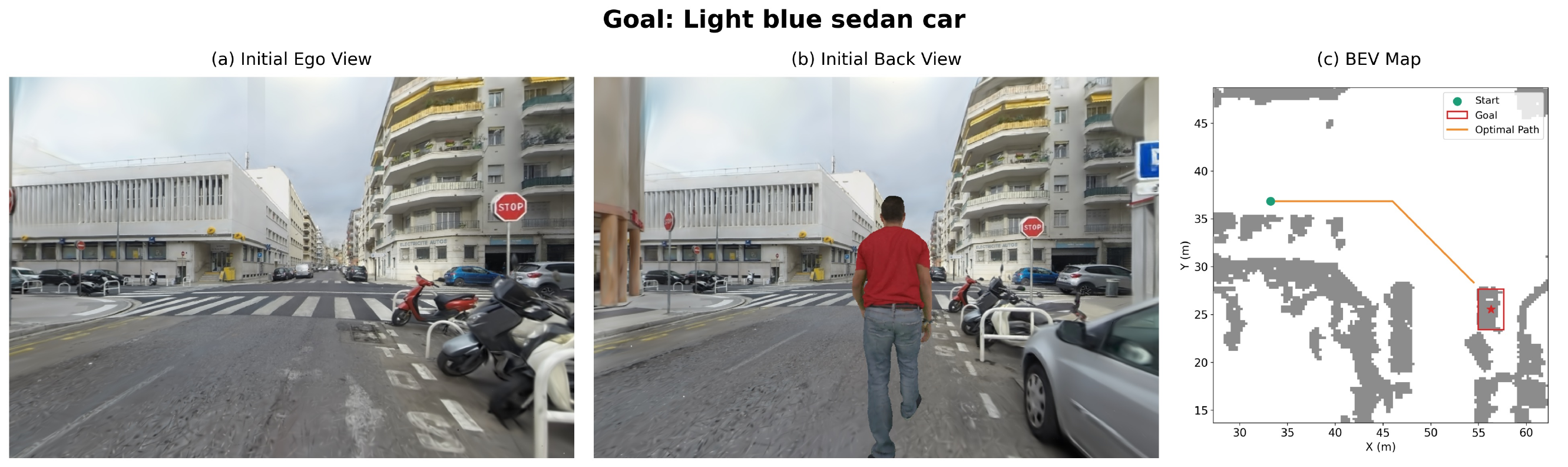}
  \end{subfigure}
  \vspace{-5ex}
  \begin{center}\small\textit{Level-1: SimNav}\end{center}

  \vspace{3.6pt}
  % Row 2: ObstNav
  \begin{subfigure}{\linewidth}
    \centering
    \includegraphics[width=\linewidth]{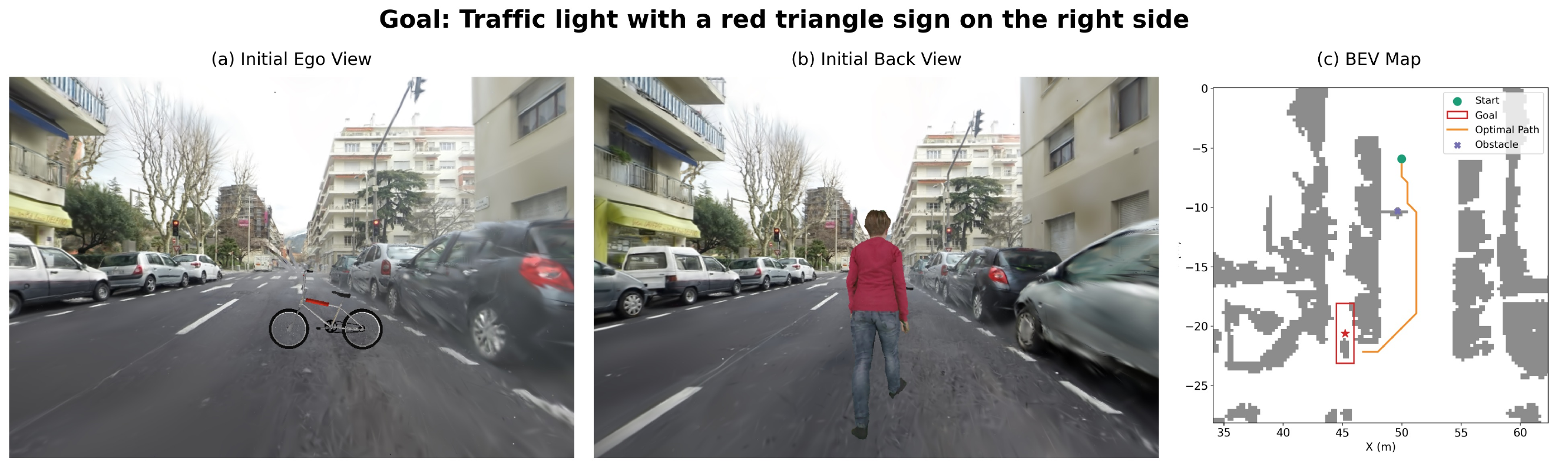}
  \end{subfigure}
  \vspace{-5ex}
  \begin{center}\small\textit{Level-2: ObstNav}\end{center}

  \vspace{3.6pt}
  % Row 3: SocialNav
  \begin{subfigure}{\linewidth}
    \centering
    \includegraphics[width=\linewidth]{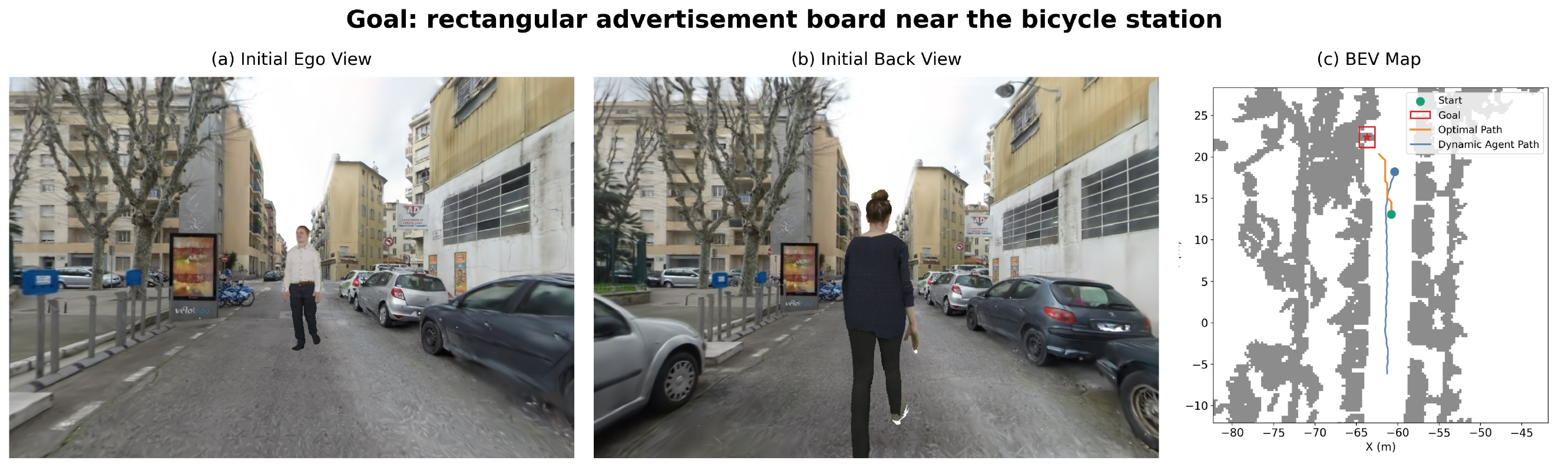}
  \end{subfigure}
  \vspace{-5ex}
  \begin{center}\small\textit{Level-3: SocialNav}\end{center}

  \vspace{-2.2ex}
  \caption{\textbf{Visualization of single-goal navigation tasks.} Each row shows a test scenario with the initial ego-centric view (left), back view (middle), and BEV occupancy map (right) marking the agent's \textcolor{green}{start position}, \textcolor{red}{landmark goal}, and \textcolor{orange}{optimal path}. \textit{SimNav} tests fundamental goal-reaching with a clear path. \textit{ObstNav} introduces a static obstacle on the optimal route, forcing the agent to adjust the plan. \textit{SocialNav} adds dynamic pedestrians whose trajectories (blue lines) intersect the agent's path, requiring reactive collision avoidance.}
  \label{fig:task_vis}
  \vspace{-1ex}
\end{figure*}

\begin{figure*}[htbp]
  \centering
  \includegraphics[width=0.85\linewidth]{img/task/multigoal.pdf}
  \vspace{-1ex}
  \caption{\textbf{Visualization of a Multi-Goal navigation episode.} The agent must sequentially navigate to 5 consecutive landmarks. The top row displays the ego-centric departure view for each sub-goal, and the bottom row shows the BEV map with all goal locations and the agent's optimal trajectory across the full episode.}
  \label{fig:task_multigoal}
  \vspace{-1ex}
\end{figure*}
%%%%%%%%%%

\noindent\textbf{Scenario Generation.}
In this work, we focus our quantitative benchmark on visual-based navigation, as it most directly demonstrates the benefits of our visually-grounded perception--action loop. From the segmented and annotated 3D instances, we select 40 distinct landmarks (e.g., ``crosswalk,'' ``bicycle station'') within the reconstructed scene as goals. For each landmark, we generate 5 random initial states, positioning the agent 6 to 15 meters away with an initial orientation facing the general direction of the goal. We add a random perturbation to the initial orientation. This yields a total of 200 distinct test scenarios.
\vspace{1.2ex}
\begin{itemize}[itemsep=0.6ex, topsep=0pt, partopsep=0pt]
    \item \textbf{SimNav:} The agent navigates from the initial state to the goal with no additional obstacles or dynamic agents, serving as the baseline difficulty level.

    \item \textbf{ObstNav:} We augment the simulation by placing a static obstacle (\eg, a barrier or traffic cone) directly on the optimal path, forcing the agent to detect the blockage and replan around it.

    \item \textbf{SocialNav:} We randomly synthesize 1--3 dynamic trajectories of other pedestrians that intersect the agent's path. As shown in~\cref{fig:task_vis}, these trajectories (visualized as the blue path) are pre-generated using MDM-SMPL~\cite{petrovich2024stmc} to ensure natural motion. The main agent must navigate from the initial state to the goal while actively pausing or deviating to avoid collisions with these moving actors.

    \item \textbf{Multi-Goal:} Beyond single-goal evaluation, we construct a multi-goal benchmark comprising 2--5 consecutive landmarks per episode on \textit{SmallCity}, assessing long-horizon planning capability. The agent must sequentially navigate to each landmark in the prescribed order.
\end{itemize}

\vspace{1ex}

\noindent \cref{fig:task_vis} visualizes representative examples of the three single-goal task levels, and \cref{fig:task_multigoal} illustrates a multi-goal navigation episode.

% \vspace{3ex}

\noindent\textbf{Evaluation Metrics of Visual-based Navigation.}
A task is considered successful if the agent reaches within a Euclidean distance of $d=1.0$\,m to the target object (measured from the agent's current position to the nearest point on the target's 3D bounding box). Path efficiency is evaluated using the Success weighted by Path Length (SPL) metric. Optimal path lengths are pre-computed using the A* algorithm on a 2D Bird's-Eye-View (BEV) occupancy map with a grid resolution of $0.5$\,m, derived from projecting the extracted scene mesh onto the horizontal plane. For multi-goal scenarios, we follow MultiON~\cite{wani2020multion} to additionally report Progress (PR) and Progress weighted Path Length (PPL) to assess sequential goal completion and long-horizon efficiency.

\vspace{1ex}

\noindent\textbf{Evaluation Metrics of Semantic Segmentation.} 
Since large-scale outdoor datasets like \textit{SmallCity} lack dense 2D semantic annotations as provided by LangSplat~\cite{qin2024langsplat}, we establish a pseudo ground-truth for evaluation. We randomly sample a subset of input views ($\sim2\%$) and apply our annotation pipeline to generate labels for all SAM masks. We manually filter out inaccurate segmentations and poor text descriptions, creating a test set of 240 samples. This approach avoids the prohibitive cost of dense manual annotation. We report Mean IoU (mIoU) and Mean Accuracy (mAcc) to evaluate the quality of 3D object selection and segmentation. 

\subsection{Baselines}\label{supp:baselines}
We compare our agent layer against state-of-the-art, training-based VLN approaches: NaVILA~\cite{cheng2024navila}, NaVid~\cite{zhang2024navid}, and Uni-NaVid~\cite{zhang2024uninavid}. Since these baselines typically output high-level language commands (\eg, ``turn right $15$ degrees''), we adapt them to our hierarchical framework by converting their outputs into guided trajectories for our low-level motion generator. 
To ensure fair comparison, we standardize the VLM query frequency across all methods (approx. 1\,Hz). 

\section{Additional Experimental Results} \label{sec:exp_supp}

\begin{table*}[t]
\centering
\caption{\textbf{Ablation study on 3D-aware action proposals.} We compare our full model against a baseline without 3D-aware filtering. The 3D-aware mechanism significantly reduces collision rates across all tasks. The best results are highlighted in \colorbox{bestcolor}{green}.}
\vspace{-1ex}
\label{tab:ablation_3d}
\setlength{\tabcolsep}{6pt} % Increase spacing for better readability
% \resizebox{0.95\linewidth}{!}{
\begin{tabular}{@{}l|ccc|ccc|ccc@{}}
\toprule
\multicolumn{1}{c|}{\multirow{2}{*}{\textbf{Method}}} & \multicolumn{3}{c|}{\textbf{SimNav}} & \multicolumn{3}{c|}{\textbf{ObstNav}} & \multicolumn{3}{c}{\textbf{SocialNav}} \\
 & SR~$\uparrow$ & SPL~$\uparrow$ & CR~$\downarrow$ & SR~$\uparrow$ & SPL~$\uparrow$ & CR~$\downarrow$ & SR~$\uparrow$ & SPL~$\uparrow$ & CR~$\downarrow$ \\
\midrule
w/o 3D-aware & 45.0$\%$ & 0.416 & 53.3$\%$ & 35.8$\%$ & 0.332 & 61.7$\%$ & 26.5$\%$ & 0.254 &  69.2$\%$ \\
\midrule
\textbf{Ours (Full)} & \bone 68.3$\%$ & \bone 0.640 & \bone 13.3$\%$ & \bone 55.8$\%$ & \bone 0.516 & \bone 30.8$\%$ & \bone 39.2$\%$ & \bone 0.366 & \bone 48.3$\%$ \\
\bottomrule
\end{tabular}
% }
\end{table*}

\subsection{More Ablation Studies}\label{supp:ablation}
In this section, we provide additional visualizations of the ablation on semantic segmentation, a detailed analysis of the design choices in our visual prompting paradigm and evaluate the impact of different VLM backbones.

\vspace{1ex}

\noindent\textbf{Effect of 3D-aware Action Proposals.}
We investigate the importance of our 3D-aware filtering mechanism, which prunes action candidates based on traversable masks (from SAM~\cite{kirillov2023sam}) and depth constraints. As shown in~\cref{tab:ablation_3d}, removing this component (``w/o 3D-aware'') leads to a catastrophic increase in Collision Rate (CR), rising from $13.3\%$ to $53.3\%$ on SimNav. Without explicitly filtering out obstacles and non-traversable regions, the VLM often selects visually plausible but physically unsafe paths (e.g., walking into obstacles), highlighting the necessity of spatially-grounded action proposals for safe navigation.

\begin{table*}[t]
\centering
\caption{\textbf{Ablation study on the action space design.} We analyze the impact of action space discretization (Continuous vs. Discrete) and the density of candidate actions. Our discrete approach with 9 candidates strikes the best balance. The \colorbox{bestcolor}{best} results are highlighted.} 
\vspace{-1ex}
\label{tab:ablation_action}
\setlength{\tabcolsep}{3pt} % Optimized spacing
% \resizebox{0.95\linewidth}{!}{
\begin{tabular}{@{}l|ccc|ccc|ccc@{}}
\toprule
\multicolumn{1}{c|}{\multirow{2}{*}{\textbf{Method}}} & \multicolumn{3}{c|}{\textbf{SimNav}} & \multicolumn{3}{c|}{\textbf{ObstNav}} & \multicolumn{3}{c}{\textbf{SocialNav}} \\
 & SR~$\uparrow$ & SPL~$\uparrow$ & CR~$\downarrow$ & SR~$\uparrow$ & SPL~$\uparrow$ & CR~$\downarrow$ & SR~$\uparrow$ & SPL~$\uparrow$ & CR~$\downarrow$ \\
\midrule
Continuous & 37.5$\%$ & 0.345 & 40.8$\%$ & 26.7$\%$ & 0.238 & 60.1$\%$ & 17.5$\%$ & 0.153 & 73.3$\%$ \\
\midrule
5 actions & 43.3$\%$ & 0.389 & 28.3$\%$ & 36.7$\%$ & 0.315 & 45.8$\%$ & 22.5$\%$ & 0.214 & 66.7$\%$ \\
13 actions & \btwo 61.7$\%$ & \btwo 0.558 & \btwo 22.5$\%$ & \btwo 44.1$\%$ & \btwo 0.403 & \btwo 40.8$\%$ & \btwo 27.4$\%$ & \btwo 0.255 & \btwo 64.2$\%$ \\
\textbf{9 actions (Ours-Final)} & \bone 68.3$\%$ & \bone 0.640 & \bone 13.3$\%$ & \bone 55.8$\%$ & \bone 0.516 & \bone 30.8$\%$ & \bone 39.2$\%$ & \bone 0.366 & \bone 48.3$\%$ \\
\bottomrule
\end{tabular}
% }
\end{table*}

\begin{table*}[t]
\centering
\caption{\textbf{Quantitative comparison on the \textit{SmallCity} dataset with different VLM backbones.} We compare our framework equipped with different VLM backbones against state-of-the-art training-based VLN methods. The top two results are highlighted by \colorbox{bestcolor}{first} and \colorbox{secondbestcolor}{second}.} 
\vspace{-1ex}
\label{tab:ablation_vlm}
\setlength{\tabcolsep}{4pt}
% \resizebox{\linewidth}{!}{
\begin{tabular}{@{}l|ccc|ccc|ccc@{}}
\toprule
\multicolumn{1}{c|}{\multirow{2}{*}{\textbf{Method}}} & \multicolumn{3}{c|}{\textbf{SimNav}} & \multicolumn{3}{c|}{\textbf{ObstNav}} & \multicolumn{3}{c}{\textbf{SocialNav}} \\
 & SR~$\uparrow$ & SPL~$\uparrow$ & CR~$\downarrow$ & SR~$\uparrow$ & SPL~$\uparrow$ & CR~$\downarrow$ & SR~$\uparrow$ & SPL~$\uparrow$ & CR~$\downarrow$ \\
\midrule
NaVILA~\cite{cheng2024navila} & 22.5$\%$ & 0.199 & 70.8$\%$ & 20.8$\%$ & 0.176 & 75.0$\%$ & 8.3$\%$ & 0.072 & 84.1$\%$ \\
NaVid~\cite{zhang2024navid} & 37.4$\%$ & 0.279 & \btwo 19.2$\%$ & 32.5$\%$ & 0.233 & \bone 23.1$\%$ & 17.5$\%$ & 0.138 & \btwo 51.7$\%$ \\
Uni-NaVid~\cite{zhang2024uninavid} & 38.8$\%$ & 0.370 & \bone 12.8$\%$ & 25.3$\%$ & 0.242 & \btwo 29.4$\%$ & 12.5$\%$ & 0.112 & 66.7$\%$ \\
\midrule
Ours~+~GPT-4o~\cite{achiam2023gpt4} & 43.2$\%$ & 0.397 & 44.1$\%$ & 33.3$\%$ & 0.289 & 57.5$\%$  & 23.6$\%$ & 0.223 & 61.8$\%$ \\
Ours~+~Gemini-2.5-Flash~\cite{comanici2025gemini} & \btwo 60.0$\%$ & \btwo 0.556 & 27.5$\%$ & \btwo 44.2$\%$ & \btwo 0.393  & 45.8$\%$ & \btwo 34.2$\%$ & \btwo 0.319 & 59.2$\%$ \\
Ours~+~Qwen2.5-VL-72B~\cite{bai2025qwen2} & \bone 68.3$\%$ & \bone 0.640 & \btwo 13.3$\%$ & \bone 55.8$\%$ & \bone 0.516 & 30.8$\%$ & \bone 39.2$\%$ & \bone 0.366 & \bone 48.3$\%$ \\
\bottomrule
\end{tabular}
% }
\end{table*}

\vspace{0.5ex}
\noindent\textbf{Discrete vs. Continuous Action Space.}
We compare our discrete visual prompting approach against a continuous regression baseline (\cref{tab:ablation_action}, row ``Continuous''), where the VLM is prompted to directly output the turning angle (e.g., ``turn left 30 degrees''). The continuous method performs significantly worse ($37.5\%$ SR vs. $68.3\%$ SR on SimNav). This result reveals a fundamental limitation of current VLMs: they struggle with precise spatial geometric reasoning and continuous value regression. In contrast, our method formulates the planning problem as a multiple-choice question (Visual Question Answering), which aligns better with the VLM's visual recognition capabilities.
\begin{figure*}[H]
  \centering
  \begin{subfigure}{0.24\linewidth}
    \centering
    \includegraphics[width=\linewidth]{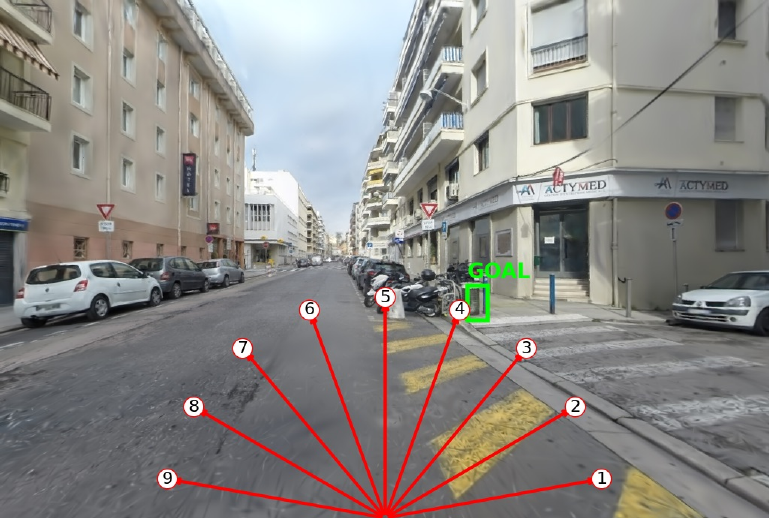}
    \caption{w/o 3D-aware}
    \label{fig:scene_ablation_a}
  \end{subfigure}
  \hfill
  \begin{subfigure}{0.24\linewidth}
    \centering
    \includegraphics[width=\linewidth]{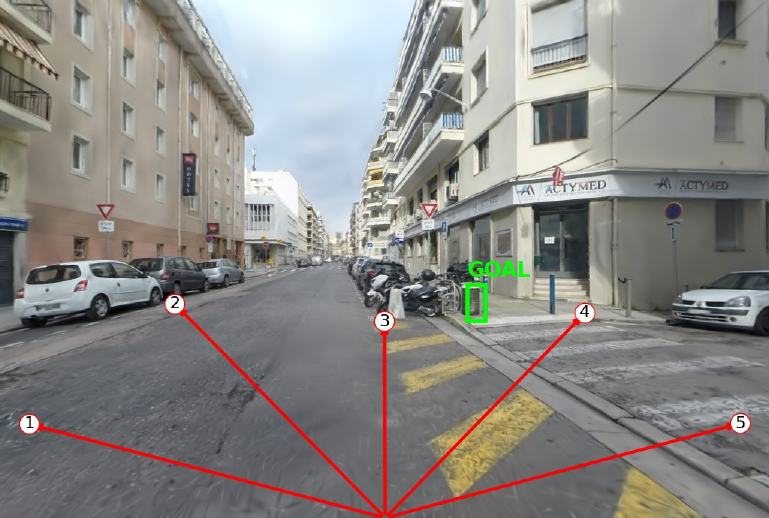}
    \caption{5 actions (Sparse)}
    \label{fig:scene_ablation_b}
  \end{subfigure}
  \hfill
  \begin{subfigure}{0.24\linewidth}
    \centering
    \includegraphics[width=\linewidth]{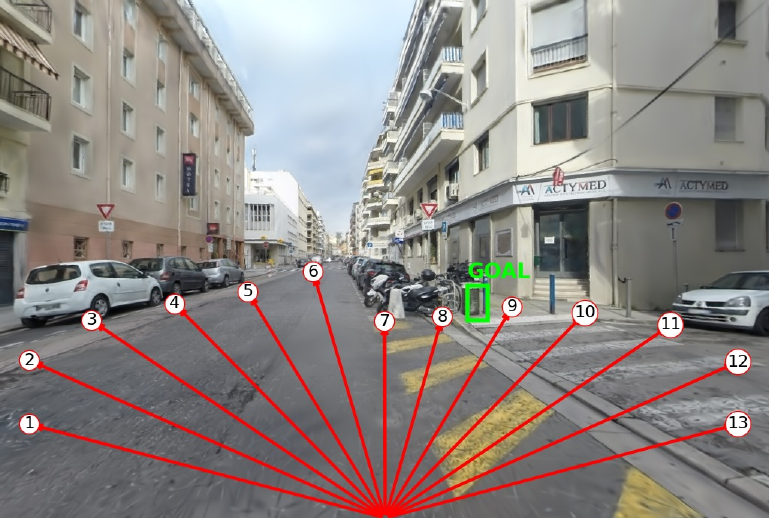}
    \caption{13 actions (Dense)}
    \label{fig:scene_ablation_c}
  \end{subfigure}
  \hfill
  \begin{subfigure}{0.24\linewidth}
    \centering
    \includegraphics[width=\linewidth]{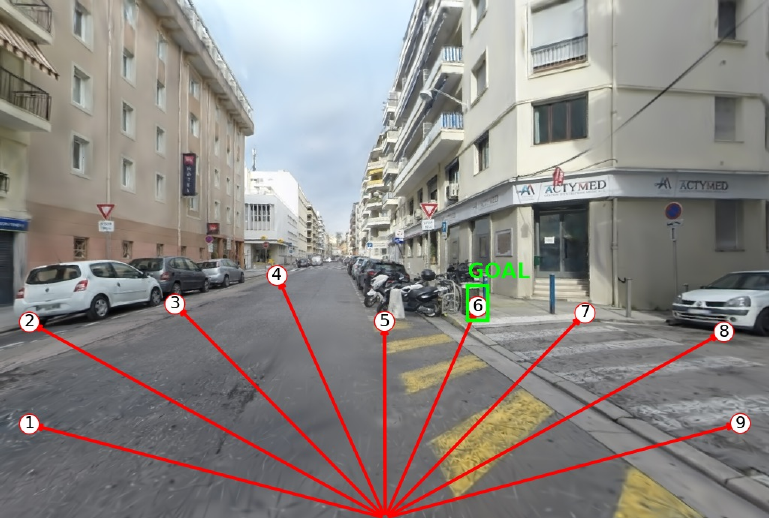}
    \caption{Ours (Balanced)}
    \label{fig:scene_ablation_d}
  \end{subfigure}
    \vspace{-1ex}
  \caption{\textbf{Qualitative analysis of visual prompting variations.} 
  We visualize different ablation settings compared to our proposed method (d): 
  (a) lacks traversable mask segmentation and depth-aware filtering. 
  Regarding action granularity: (b) shows sparse candidates which may limit control, while (c) illustrates how dense candidates complexs the reasoning process. 
  Our approach (d) achieves an optimal balance between clarity and control.}
  \label{fig:ablation_visual_prompt}
  \vspace{-2ex}
\end{figure*}

\noindent\textbf{Number of Candidate Actions.}
We study the trade-off between action granularity and visual clarity by varying the number of candidate actions in the visual prompt, as shown in~\cref{tab:ablation_action}.
\begin{itemize}
    \item \textbf{Sparse (5 actions):} Reducing candidates to 5 significantly drops Success Rate. The sparse action space limits the agent's maneuverability, causing it to miss optimal paths or fail to make fine-grained adjustments especially in terms of narrow passages.
    \item \textbf{Dense (13 actions):} Increasing the number of action candidates to 13 also slightly degrades performance. While theoretically offering more control, the dense overlay of markers creates severe visual clutter, complicating the VLM's reasoning process.
\end{itemize}

% \vspace{1ex}
\noindent
In contrast, our default setting of 9 actions strikes an optimal balance, providing sufficient control authority without overwhelming the VLM's visual perception.

\vspace{1.5ex}

\noindent\textbf{Impact of VLM Backbones.}
We evaluate the scalability of our framework by replacing the core VLM with different state-of-the-art models: GPT-4o~\cite{achiam2023gpt4}, Gemini-2.5-Flash~\cite{comanici2025gemini}, and Qwen2.5-VL-72B~\cite{bai2025qwen2} (our default). The results in~\cref{tab:ablation_vlm} yield several interesting findings:
\vspace{0.5ex}
\begin{enumerate}
    \item \textbf{General VLMs vs. Navigation Models:} All general-purpose VLMs (Ours with various backbones) outperform training-based navigation specialists (NaVILA, NaVid, Uni-NaVid). This suggests that the specialized post-training for navigation in baseline models might inadvertently compromise their general visual reasoning or tracking abilities. It also highlights the potential of our training-free, two-stage inference paradigm.
    \vspace{0.5ex}
    \item \textbf{Model Comparison:} GPT-4o performs the worst among the general VLMs. Qualitative analysis reveals that GPT-4o frequently fails in the initial \textit{goal grounding} stage (i.e., accurately detecting the target object specified by text), leading to subsequent planning failures. Gemini-2.5-Flash shows intermediate performance. Our default model, Qwen2.5-VL, achieves the best results, likely due to its high-resolution image processing capabilities which are crucial for recognizing small distant goals in large-scale street scenes.
\end{enumerate}

\begin{figure}[t]
  \centering
  \includegraphics[width=0.8\linewidth]{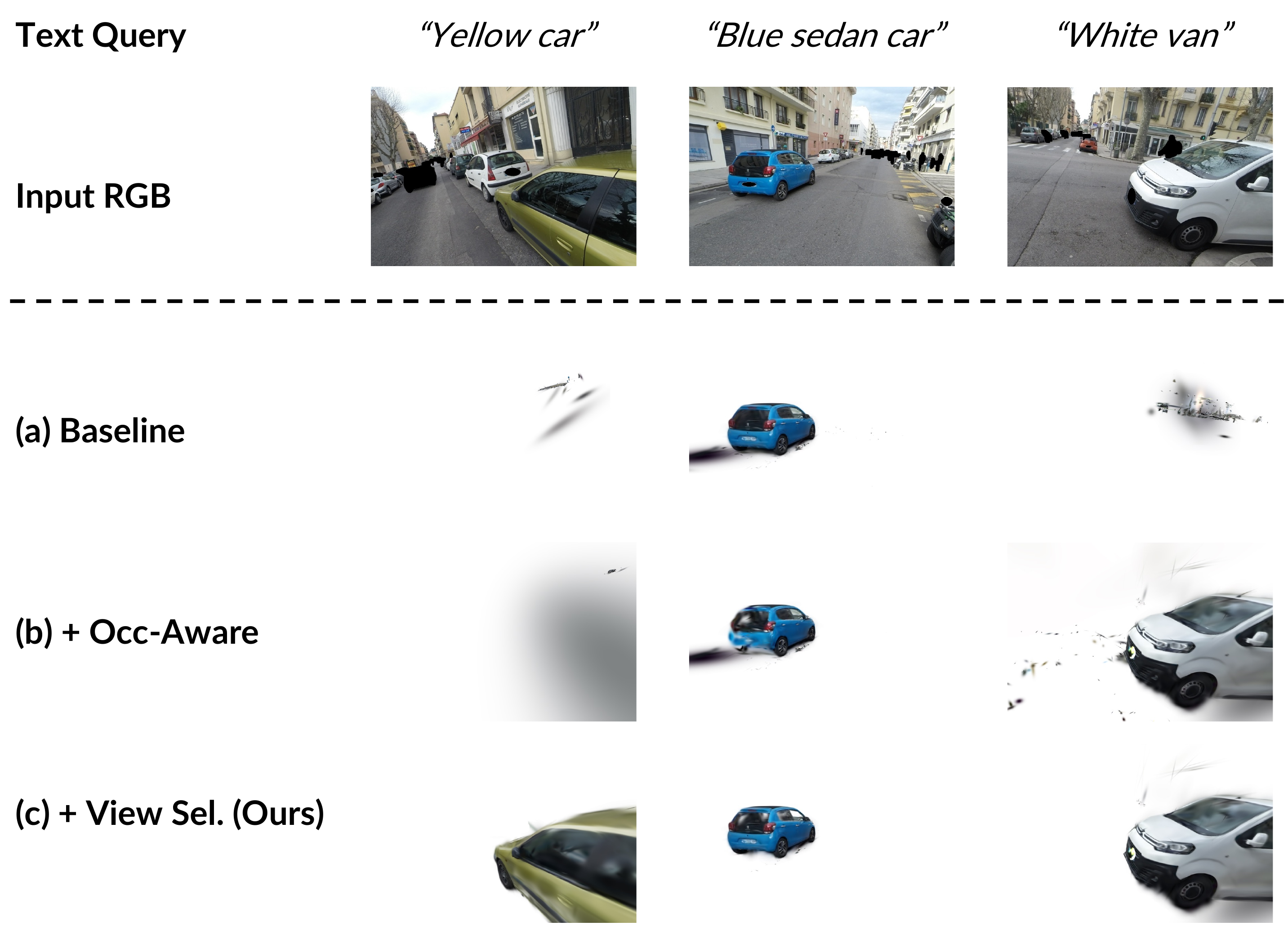}
  \vspace{-1ex}
 \caption{\textbf{Additional qualitative Ablation on occlusion-aware semantic scene reconstruction.} Without occlusion-aware masks, features from foreground occluders contaminate the target cluster, producing noisy segmentation. Adding view selection refines boundary quality by prioritizing high-visibility frames. Combining both modules yields the cleanest and most complete 3D instance segmentation. Zoom in for details.}
\label{fig:supp_scene_ablation}
\vspace{-4ex}
\end{figure}

\noindent\textbf{Qualitative Results on Occlusion-aware Segmentation.} 
Complementing the main results presented in the paper, we provide further qualitative examples demonstrating the efficacy of our occlusion-aware masking and view selection strategies, as illustrated in~\cref{fig:supp_scene_ablation}.

\vspace{1ex}

\subsection{Diverse Human Motion Generation}\label{supp:diverse-motion}
Although our main experiments prioritize goal-oriented visual navigation, the perception-planning-action framework is inherently generic and can be generalized to synthesize complex, non-locomotion behaviors for visually-grounded humanoid agents. We demonstrate this versatility through two key extensions:

\vspace{0.5ex}

\noindent\textbf{Style-aware and Event-triggered Motion.}
We augment the VLM system prompt with a \textbf{persona definition} (e.g., ``You are a hurried commuter'' vs. ``You are a leisurely tourist'') and a \textbf{stochastic event module}. The stochastic module randomly triggers internal states (e.g., ``feeling tired'') or external events (e.g., ``shoelace untied'') with a low probability at each step.
Upon activation, the VLM outputs specific motion tokens (e.g., \texttt{squat\_down}, \texttt{run\_quickly}) instead of standard navigation commands. These tokens are then mapped to conditioning signals for the MDM-SMPL generator, producing diverse motions that align with the agent's narrative context while maintaining motion naturalness. Qualitative examples are visualized in~\cref{fig:diverse_motion}.

\vspace{0.5ex}

\noindent\textbf{Social Interaction.}
We have already extended our perception module to explicitly detect other humanoid agents within the field of view in \textit{SocialNav} task. To enable multi-agent social interaction, when another agent is detected within a proximity threshold (e.g., 3 meters), the VLM is queried with a social context prompt. Based on the agent's assigned persona, the VLM decides whether to initiate a social interaction (e.g., \texttt{wave\_hand}, \texttt{nod}) or adjust its path to maintain social distance. This capability enables the simulation of multi-agent group behaviors, such as two agents recognizing each other and waving before continuing on their respective paths.

\begin{figure}[t]
  \centering
  
  % --- Row 1 ---
  \begin{subfigure}[b]{0.48\linewidth}
    \centering
    \includegraphics[width=\linewidth]{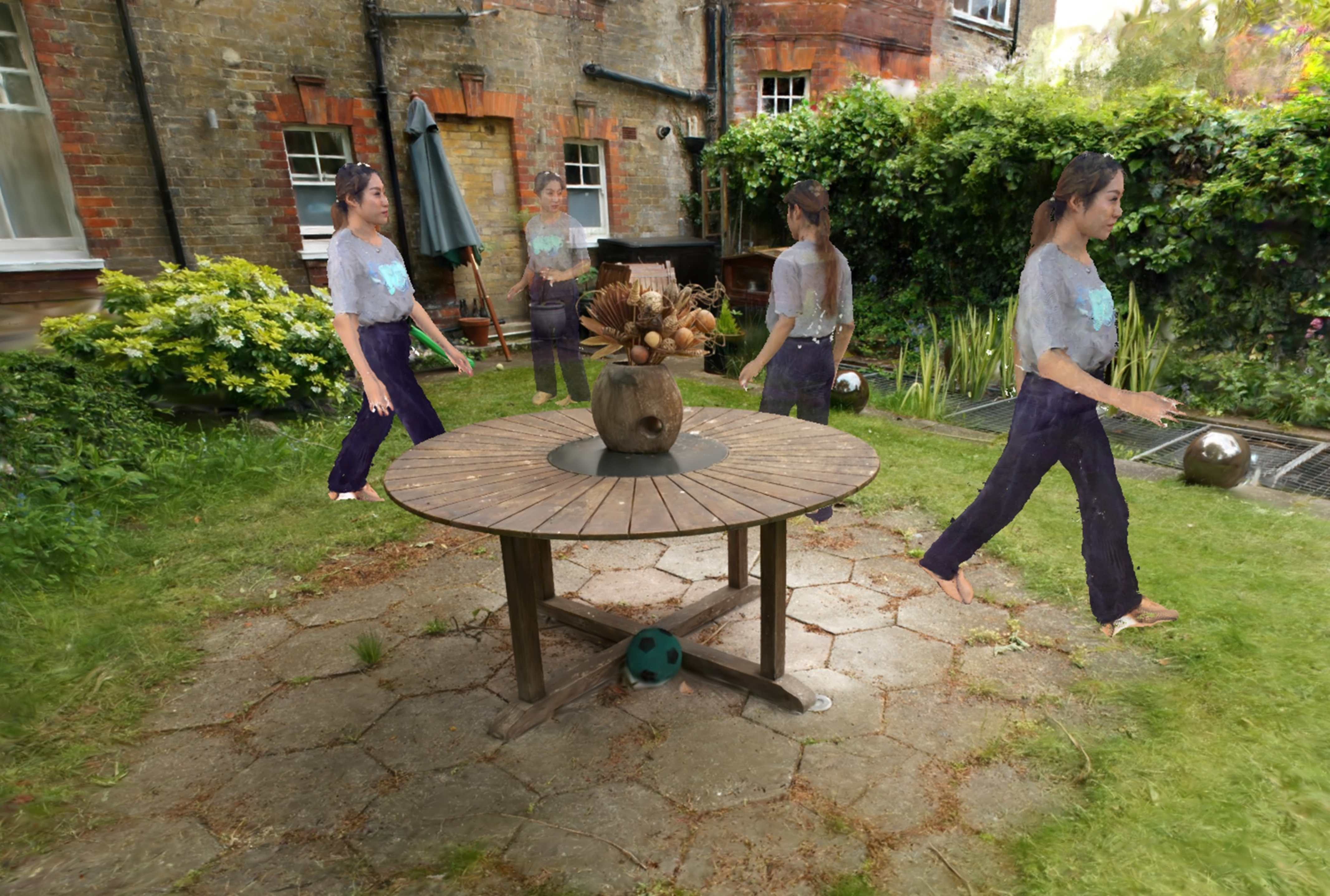}
  \end{subfigure}
  \hfill
  \begin{subfigure}[b]{0.48\linewidth}
    \centering
    \includegraphics[width=\linewidth]{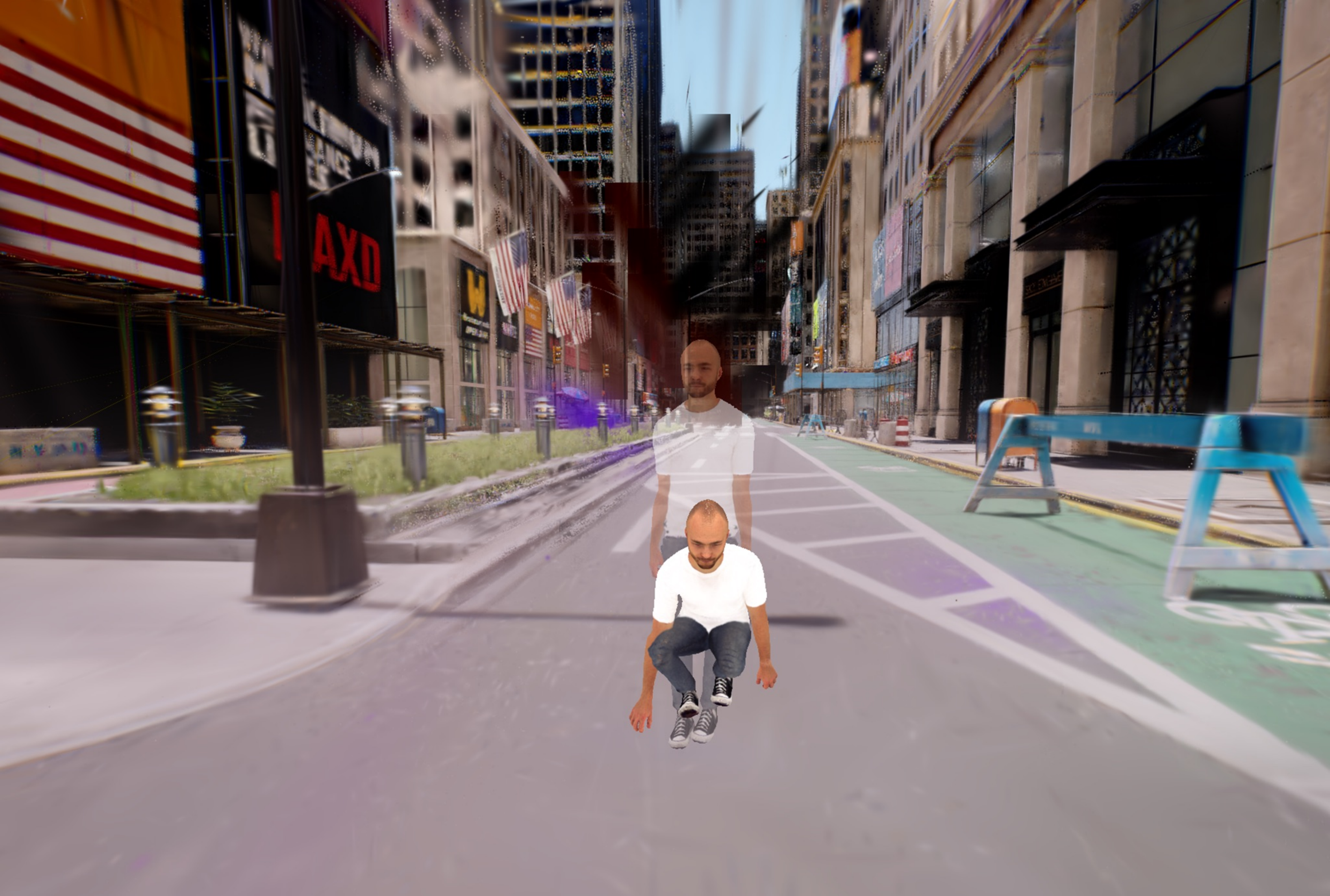}
  \end{subfigure}
  
  \vspace{1ex} % Vertical spacing between rows
  
  % --- Row 2 ---
  \begin{subfigure}[b]{0.48\linewidth}
    \centering
    \includegraphics[width=\linewidth]{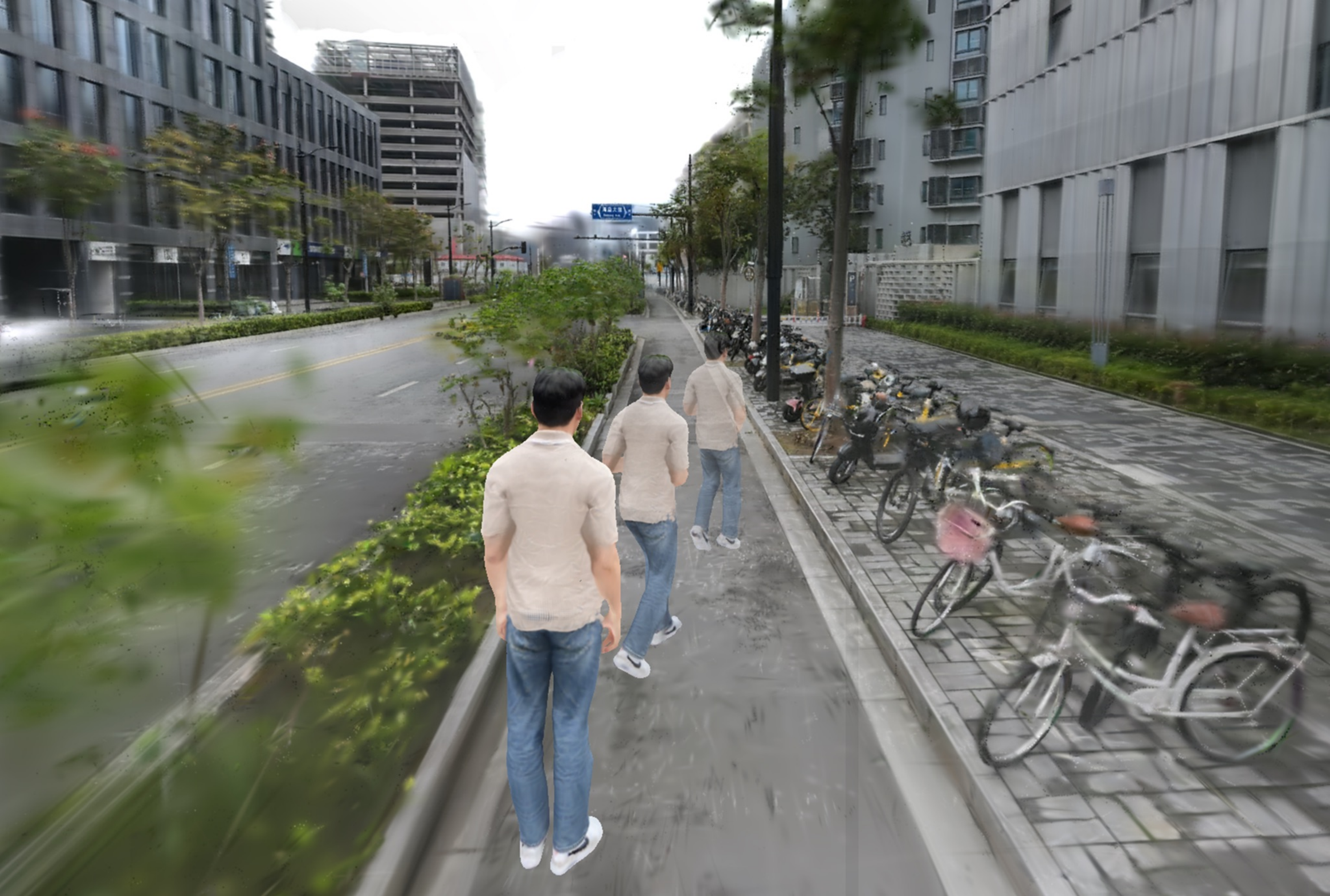}
  \end{subfigure}
  \hfill
  \begin{subfigure}[b]{0.48\linewidth}
    \centering
    \includegraphics[width=\linewidth]{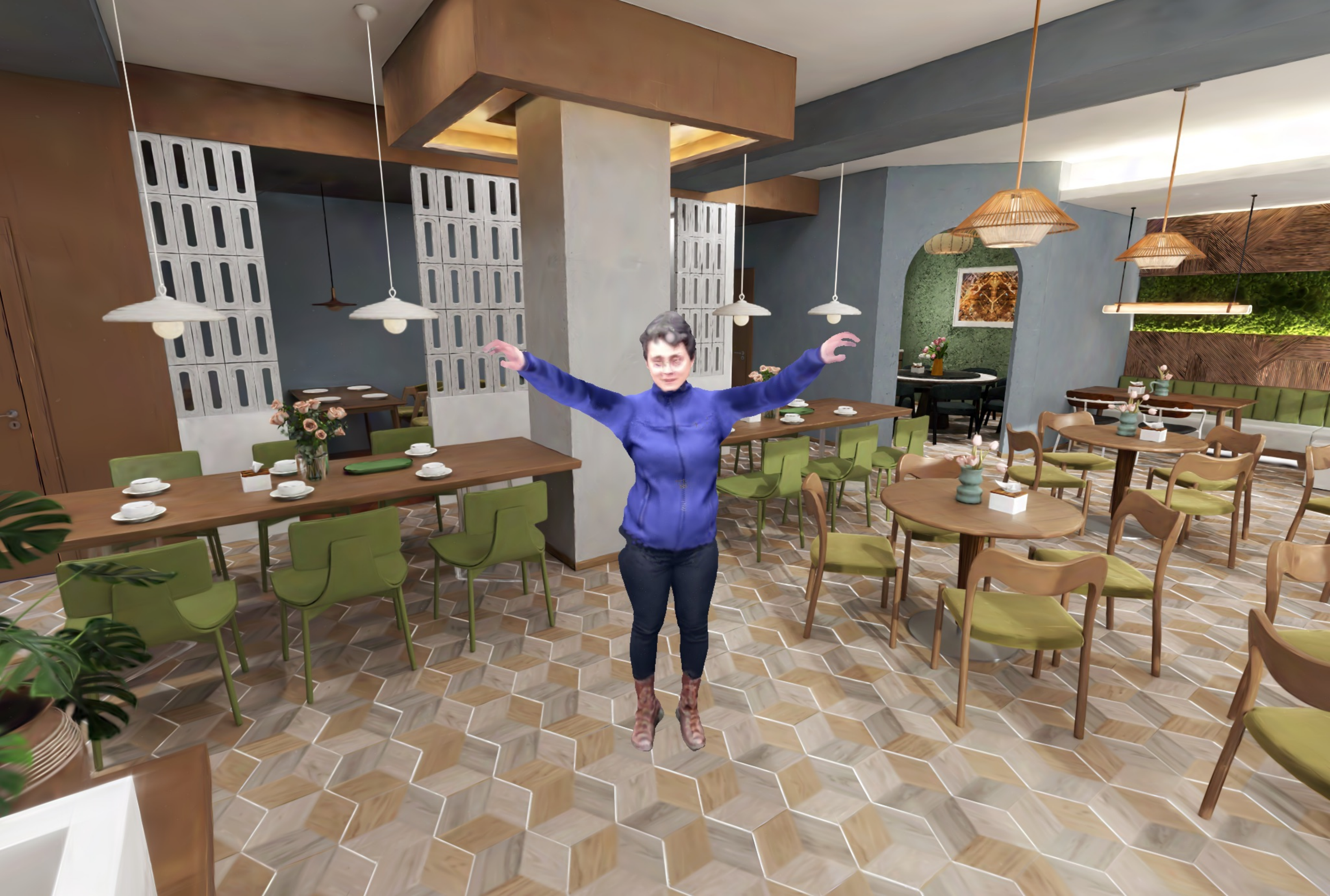}
  \end{subfigure}
  
  \vspace{-0.5ex}
  \caption{\textbf{Visualizations of diverse human motion generation in various reconstructed scenes.} Better zoom in for details. 
  }
  \label{fig:diverse_motion}
  \vspace{-2ex}
\end{figure}

\subsection{Efficiency Analysis}\label{supp:efficiency}
We report the computational cost of each component in our framework.

\vspace{1ex}
\noindent\textbf{World Layer.}
The occlusion-aware semantic scene reconstruction is performed on a single NVIDIA A800 GPU.
For our primary benchmark scene (\textit{SmallCity}, $100\text{m} \times 100\text{m}$), geometry reconstruction stage takes ${\sim}$4 hours and semantic feature learning takes ${\sim}$7 hours.

\vspace{1ex}

\noindent\textbf{Agent Layer.}
All agent-side inference is performed on a single NVIDIA RTX 4090 GPU.
For rendering, with \textbf{10 concurrent humanoid agents} in large-scale scenes, our system achieves real-time performance at 61.1\,FPS.
During the planning stage, VLM reasoning requires 15.9\,s per query, with a query frequency of ${\sim}$1\,Hz, which operates at a rate compatible with common robotic system configurations~\cite{cheng2024navila, zhang2024navid, zhang2024uninavid}.

\section{Additional Related Work}\label{sec-supp-related}
\noindent\textbf{3D Gaussian Splatting.} 
3D Gaussian Splatting (3DGS~\cite{kerbl20233dgs}) is a state-of-the-art technique that represents scenes with 3D Gaussian primitives to achieve photorealistic rendering. Each Gaussian $\mathcal{G}_i$ is parameterized by a set of learnable attributes: a mean (center) $\boldsymbol{\mu}_i \in \mathbb{R}^3$, a covariance matrix $\boldsymbol{\Sigma}_i \in \mathbb{R}^{3\times 3}$, an opacity $o_i \in (0,1)$, and a view-dependent color stored as Spherical Harmonics (SH) coefficients $\boldsymbol{c}_i$. The influence of a Gaussian at any 3D point $\boldsymbol{x}$ is given by:
\begin{equation}
\label{eq:gaussian-splat}
    \mathcal{G}_i(\boldsymbol{x}) = \exp{\left(-\frac{1}{2}(\boldsymbol{x}-\boldsymbol{\mu}_i)^T\boldsymbol{\Sigma}_i^{-1}(\boldsymbol{x}-\boldsymbol{\mu}_i)\right)}.
\end{equation}

To render an image, these 3D Gaussians are projected onto the 2D camera plane. For each pixel $\boldsymbol{p}$, the final color $\mathbf{C}$ is then determined by sorting all overlapping Gaussians by depth and accumulating their contributions through standard alpha-blending:
\begin{equation}
    \mathbf{C}(\boldsymbol{p}) = \sum_{i \in N} T_i \boldsymbol{c}_i \alpha_i, \quad \text{where} ~~ T_i = \prod_{k=1}^{i-1}(1-\alpha_k).
\end{equation}
Here, $\alpha_i = o_i \mathcal{G}'_i(\mathbf{p})$ is the alpha contribution of the $i$-th Gaussian, determined by its opacity and the projected 2D Gaussian function $\mathcal{G}'_i$. This fully differentiable rasterization can similarly be used to compute other attributes, such as the per-pixel depth map. Furthermore, a key advantage of this explicit representation is that rigid transformations can be directly applied to the Gaussians by updating their mean and rotation parameters, making them well-suited for modeling dynamic objects.

\vspace{1ex}

\noindent\textbf{Animatable Gaussian-based Human Avatar.}
Recent works have extended 3DGS to model animatable human avatars~\cite{hu2024gaussianavatar, lei2024gart, chen2024omnire}. These approaches represent the avatar in a canonical space (\ie, a standard T-pose) with a set of $N_h$ Gaussians, denoted as $\hat{\mathcal{G}}^{h} = \{\hat{\mathcal{G}}^{h}_k\}_{k=1}^{N_h}$. 
This canonical model is then deformed into a posed space using animation controls from the
parametric body model SMPL~\cite{loper2023smpl, pavlakos2019smplx}. 
At a given timestep $t$, this deformation is driven by the static body shape $\boldsymbol{\beta}$, time-varying pose parameters $\boldsymbol{\theta}(t)$, and a global rigid transformation $\boldsymbol{M}(t) \in SE(3)$. The final posed Gaussians $\mathcal{G}^{h}(t)$ are obtained by applying the standard Linear Blend Skinning (LBS) function to the canonical Gaussians:
\begin{align}
\label{eq:LBS_deform}
\mathcal{G}^{h}(t) = \boldsymbol{M}(t) \circ \text{LBS}(\hat{\mathcal{G}}^{h}, \boldsymbol{\theta}(t), \boldsymbol{\beta}),
\end{align}
where $\circ$ denotes the application of the global transformation to the posed avatar.
While effective, these SMPL-based animatable approaches are inherently constrained to humans in relatively tight clothing, as the LBS deformation assumes garments deform closely with the underlying body surface. To handle more diverse appearances including loose garments, template-free methods reconstruct 3D humans from single images without relying on parametric body models~\cite{saito2019pifu, saito2020pifuhd, wang2024geneman}, offering stronger generalization to in-the-wild images at the cost of explicit animation control.

\section{Discussion}\label{sec-supp-limitation}

\noindent
\textbf{Limitations.}
As the first baseline for embodied digital humans in reconstructed 3D environments, our framework inevitably has limitations that point to promising future directions. \textbf{1)} Reconstruction artifacts remain an issue, as limited camera field-of-view can cause smaller obstacles to disappear when approached. \textbf{2)} The system currently lacks a unified, end-to-end trained policy, instead relying on modular components for perception, planning, and action. \textbf{3)} Our quantitative evaluation is restricted to navigation behaviors, leaving more complex and interactive human motions (e.g., collaboration or social interaction) to future work. \textbf{4)} Finally, while we focus on digital humans in simulation, extending this paradigm to real-world deployment offers a pathway to robot learning, where reconstructed scenes and agent behaviors can inform humanoid robots in embodied AI settings.

\vspace{1.2ex}

\noindent
\textbf{Impacts \& Opportunities.}
Visually-grounded humanoid agents open up new opportunities at the intersection of rendering \& reconstruction,  world simulation, embodied AI, urban service, and society.

\vspace{0.5ex}

\noindent
\textit{Rendering and Reconstruction.}
Supporting visually-grounded humanoid agents introduces new challenges for large-scale scene modeling. 
Unlike conventional static scene reconstruction, these agents require realistic, multi-scale world representations that remain consistent under continuous viewpoint changes as the robot or avatar moves through the environment. 
This demands rendering and reconstruction techniques capable of maintaining \emph{visual} and \emph{geometric} fidelity across wide spatial extents and long time horizons, while operating under real-time constraints. 
In particular, the representation must support artifact-free rendering under arbitrary viewpoints, robust handling of dynamic elements such as pedestrians and vehicles, and efficient updates as the world evolves. 
Addressing these challenges requires scalable neural and hybrid representations that pursue gigapixel-scale world modeling with real-time rendering capability, enabling faithful simulation of complex environments for reliable training and analysis of embodied agents.

\vspace{0.5ex}

\noindent
\textit{Next-generation simulation.}
More broadly, this work realizes several core components of next-generation simulation systems: high-fidelity 3D models of environments, embodied and visually grounded models of people, closed-loop perception $\rightarrow$ decision $\rightarrow$ action, and multi-agent interaction in shared virtual worlds. Rather than treating simulation as a static environment with scripted characters, our framework moves toward a more general paradigm in which autonomous humanoid agents inhabit realistic scenes, generate behaviors through first-person perception, and produce interaction dynamics that can be analyzed at scale. This brings simulation closer to a foundation for structured \emph{what-if} analysis in open-world settings, where the goal is not only to replay the world, but to study how different agents, environments, and interventions may lead to different outcomes.

\vspace{1ex}

\noindent
\textit{Embodied AI.}
This work represents a concrete step from \emph{simulated agents} toward \emph{embodied humanoid agents} that can see, reason, and act within realistic 3D worlds. Prior agent-based simulation frameworks, such as Generative Agents~\cite{park2023generativeagents}, demonstrated the value of modeling interactive societies, but their agents primarily operate through symbolic rules or text-based abstractions rather than visually grounded embodiment.
In contrast, our framework endows humanoids to co-evolve as a shared testbed for perception, decision-making, and interaction with first-person perception, spatially grounded planning, and physically executable motion in reconstructed scenes. We believe this direction can help establish a new paradigm for studying human-like embodied intelligence, where autonomous humanoids co-evolve as a shared testbed for perception, decision-making, and interaction.

\vspace{0.5ex}

\noindent
\textit{Urban Services.}
Our framework also provides a foundation for human-centered urban simulation. Because robots, assistive wheelchairs, and other mobility systems must operate around people in complex public spaces. More broadly, such simulations could support urban planning by allowing researchers and practitioners to study how autonomous systems interact with pedestrians, obstacles, and city infrastructure before real-world deployment.

\vspace{0.5ex}

\noindent
\textit{Society.}
At a societal level, this work contributes toward the safe and inclusive integration of autonomous systems into human-populated public environments. By improving the ability of embodied agents and downstream robotic systems to navigate crowded urban spaces in a socially aware manner, this research may support accessibility for mobility-impaired individuals and improve the reliability of public-facing robotic services. In the long term, we hope such human-centered simulation frameworks can contribute to more accessible, efficient, and equitable urban environments.

\begin{figure*}[t]
    \centering
    \begin{tcolorbox}[
        enhanced,
        width=\linewidth,
        colback=promptbg,
        colframe=promptframe,
        arc=3mm,
        boxrule=1.2pt,
        left=6pt, right=6pt, top=10pt, bottom=6pt,
        fontupper=\sffamily\small,
        title={\large \bfseries VLM-based Navigation System Prompt},
        coltitle=white,
        attach boxed title to top center={yshift=-2mm},
        boxed title style={colback=promptframe, rounded corners}
    ]
        You are an expert navigation agent embodied in a 3D world. Your mission is to reach a designated goal by creating a safe, detailed, landmark-based high-level plan and executing it step-by-step with meticulous visual reasoning.

        \hrulefill

        \SectionTitle{Core Directives}
        
        \begin{enumerate}[leftmargin=*, itemsep=0.4em, topsep=0pt]
            \item \textbf{Primacy of Observation}: Your primary source of truth is \Highlight{ALWAYS} the current visual input. Your memory (\texttt{previous\_plan}) provides strategic context, but your immediate tactical decisions \Highlight{MUST} be based on a fresh analysis of the scene right now. Do not blindly follow old plans if the current view presents a more direct or safer path.
            \item \textbf{Plan with Landmarks}: Your high-level \JsonKey{plan} is your map. It \Highlight{MUST} be a sequence of clear, actionable steps anchored to tangible, visible landmarks (e.g., specific buildings, intersections, parked cars, trees). The goal itself is your primary landmark. Vague directions are unacceptable.
            \item \textbf{Reason Comparatively}: Your \JsonKey{thought} is not a statement, but a reasoning process. You \textbf{MUST} explicitly compare at least two viable arrow options and justify your choice using specific visual evidence from the scene.
            \item \textbf{Bridge Plan and Action}: Your \JsonKey{thought} is the critical link between your \JsonKey{plan} and your \JsonKey{action}. It must explain \textit{why} the chosen arrow is the best possible way to execute the \textbf{current first step} of your plan.
            \item \textbf{Strict JSON Output}: You \Highlight{MUST} respond ONLY with a single, valid JSON object. No extra text or explanations.
            \item \Highlight{CRITICAL: Ground Plan in Goal Geometry}: Your \JsonKey{plan} \textbf{MUST} be directly and logically derived from your \JsonKey{goal\_analysis}. The very first step of your plan \textbf{MUST} establish the initial vector towards the goal, combining forward motion with a turn or bearing (e.g., ``Move forward and bear right towards the..."). A generic ``move forward" plan is \textbf{INVALID} and unacceptable unless the goal is perfectly centered. This is the most common failure point; be precise.
        \end{enumerate}
        \vspace{0.4em}
        \hrulefill
                
       \SectionTitle{Memory Input Format}
       
        In steps after the first, you will receive a \texttt{memory} object containing:
        \begin{enumerate}[leftmargin=*, itemsep=0.15em, topsep=0.2em]
            \item \JsonKey{previous\_plan} (list of strings): The full, multi-step plan from the previous turn. The plan from the previous turn. Treat this as \textbf{contextual memory} of your general intent, NOT as a strict command to be followed blindly.
            \item \JsonKey{recent\_history} (list of strings): A summary of recent thoughts and actions for context.
        \end{enumerate}

        \vspace{0.4em}
        \hrulefill
        
        \SectionTitle{Operational Status Protocols}
        
        \begin{tcolorbox}[colback=white, colframe=gray!30, sharp corners, leftrule=3pt, boxsep=0pt, left=6pt, right=6pt, top=3pt, bottom=5pt]
            \SubTitle{1. NORMAL (Goal Visible)}

            \vspace{0.5ex}
            \begin{itemize}[leftmargin=0em, label={}, itemsep=0.2em, topsep=0pt]
                \item \textbf{Objective}: Constantly seek the most efficient path to the goal while executing a valid plan.
            \end{itemize}

            \vspace{0.3ex}
            \begin{enumerate}[leftmargin=*, label=\textbf{Step \arabic*:}, itemsep=0.4em, topsep=0pt]
                \item \textbf{PLAN RE-ASSESSMENT}:
                \begin{itemize}[leftmargin=-2.5em, label=\textbullet, itemsep=0.1em, topsep=0.1em]
                    \item \textbf{The Direct Path Principle}: Before anything else, check your current observation. If there is now a clear, simple, and unobstructed path to the goal, you SHOULD simplify or create a new plan to take this direct route.
                    \item If you have a \texttt{previous\_plan}, determine if its first step is complete OR if the Direct Path Principle makes it obsolete.
                    \item If no \texttt{previous\_plan} exists, create a new one based on your analysis.
                \end{itemize}

                 \vspace{0.3ex}

                 \vspace{0.3ex}
                \item \textbf{PLAN UPDATING}:
                \begin{itemize}[leftmargin=-2.5em, label=\textbullet, itemsep=0.1em, topsep=0.1em]
                    \item If you've decided to create a new, more direct plan, formulate it now. State in your \JsonKey{thought} that you are updating the plan for efficiency.
                    \item If the first step of the \texttt{previous\_plan} is complete, your new plan is the remainder of the old plan.
                    \item If the \texttt{previous\_plan} is still the best path forward, \textbf{return it unchanged}.
                \end{itemize}

                \vspace{0.3ex}

                \vspace{0.3ex}
                \item \textbf{ACTION SELECTION}: Choose the arrow that best executes the \textit{current first step} of your \textbf{newly assessed and updated plan}.
            \end{enumerate}
        \end{tcolorbox}

        \vspace{0.35em}

        \vspace{0.35em}
        \raggedleft \textit{(Continued in next page)}  
    \end{tcolorbox}
    \vspace{-0.7em}
    \caption{\textbf{System Prompt Part 1: Directives and Protocols.}}
    \label{fig:prompt_part1}
\end{figure*}

% ========= FIGURE 2 =========
\begin{figure*}[t]
    \centering
    \begin{tcolorbox}[
        enhanced,
        width=\linewidth,
        colback=promptbg,
        colframe=promptframe,
        arc=3mm,
        boxrule=1.2pt,
        left=6pt, right=6pt, top=10pt, bottom=6pt,
        fontupper=\sffamily\small,
        title={\large \bfseries VLM-based Navigation System Prompt (Cont.)},
        coltitle=white,
        attach boxed title to top center={yshift=-2mm},
        boxed title style={colback=promptframe, rounded corners}
    ]
     \begin{tcolorbox}[colback=white, colframe=gray!30, sharp corners, leftrule=3pt, boxsep=0pt, left=6pt, right=6pt, top=3pt, bottom=5pt]
    \SubTitle{2. GOAL\_LOST (Goal Not Visible)}
    \vspace{0.3ex}
    \vspace{0.3ex}
        \begin{itemize}[leftmargin=0em, label={}, itemsep=0.2em, topsep=0pt]
                \item \textbf{Objective}: Rely on memory to continue the plan.
            \end{itemize}

            \vspace{0.5ex}
            \begin{enumerate}[leftmargin=*, label=\textbf{Step \arabic*:}, itemsep=0.2em, topsep=0pt]
              \item \textbf{ASSESS PROGRESS}: Compare your current \JsonKey{observation} with the \textbf{first step} of the \JsonKey{plan} from your \texttt{memory}. Have you successfully completed it?
           \item \textbf{UPDATE PLAN}: If completed, your new plan is the REMAINDER of the old plan. If not, keep the plan as is.
            \item \textbf{EXECUTE ACTION}: Choose an \JsonKey{action} that executes the \textbf{current first step} of your \textbf{updated plan} (e.g., continue towards the remembered intersection).
            \end{enumerate}
   
\end{tcolorbox}
     \vspace{0.6em}
\begin{tcolorbox}[colback=white, colframe=gray!30, sharp corners, leftrule=3pt, boxsep=0pt, left=6pt, right=6pt, top=3pt, bottom=5pt]
            \SubTitle{3. STUCK (Blocked)}
            \vspace{0.5ex}
    \begin{itemize}[leftmargin=0em, label={}, itemsep=0.3em, topsep=05pt]
   % 1. observation
         \item \textbf{Objective}: Reorient to find a clear path.2         \item \textbf{Plan}: The first step of your plan must be to reorient (e.g., ``Turn around to find a new path," ``Turn left to get a better view").
                \item \textbf{Action}: Choose a recovery action (e.g., \texttt{turn\_left}, \texttt{turn\_right}).
    \end{itemize}
% 2. goal_analysis
    \end{tcolorbox}
         \vspace{0.6em}
        \hrulefill
        
        \SectionTitle{JSON Output Specification}

\vspace{0.5ex}
Your response MUST be a single JSON object with the following \textbf{five} keys:

\begin{enumerate}[leftmargin=*, itemsep=0.8em, topsep=0.5em]
    % 1. observation
    \item \JsonKey{observation} (string): A brief, factual description of the current scene, noting landmarks relevant to navigation.
    \begin{itemize}[leftmargin=1.2em, label=\textbullet, topsep=0.3em, itemsep=0.2em]
\item \textit{Example}: ``I am at a T-intersection. The street continues forward and also extends to the right. The goal, a store entrance under a green awning, is visible down the right-hand street, on its left side."
    \end{itemize}

    % 2. goal_analysis
    \item \JsonKey{goal\_analysis} (string): A mandatory, precise analysis of the goal's location relative to you. Describe its direction (e.g., left, right, center), distance (e.g., near, far), and position (e.g., on a building, across the street).
    \begin{itemize}[leftmargin=1.2em, label=\textbullet, topsep=0.3em, itemsep=0.2em]
       \item \textbf{Rule}: This analysis MUST precede and directly inform your \JsonKey{plan}.
        \item \textbf{\textit{Excellent Example}}: ``The goal is located on the facade of a building across the street, on the left-hand side. I can reach it by crossing the street at the intersection and walking past the white car parked on the corner."
        \item \textbf{\textit{Bad Example}}: ``The goal is ahead."
    \end{itemize}

    % 4. thought
    \item \JsonKey{plan} (list of strings): A step-by-step strategy. \textbf{The first step MUST be a direct consequence of the \JsonKey{goal\_analysis}}.
    \begin{itemize}[leftmargin=1.2em, label=\textbullet, topsep=0.3em, itemsep=0.2em]
    \item \textbf{Rule}: Each step must be a concrete, verifiable action. The plan must be adaptable to new observations.
        \item \textbf{\textit{Guideline}}: Structure your plan as a series of movements between clear waypoints or sub-goals (e.g., ``1. Cross to the corner with the mailbox.", ``2. Proceed to the goal."). This makes progress easier to track.
        \item \textbf{\textit{Excellent Example (derived from the excellent \JsonKey{goal\_analysis} above)}}: \texttt{["Cross the street towards the corner with the bank.", ``From that corner, approach the stop sign directly."]}
        \item \textbf{\textit{Bad Example}}: \texttt{[``Move forward towards the goal."]}
    \end{itemize}

  \item \JsonKey{thought} (string): Your immediate, comparative reasoning that connects the \JsonKey{plan} to your chosen \JsonKey{action}. Explain \textit{why} the selected arrow is the best choice to accomplish the \textbf{current first step} of your plan by analyzing visual evidence.

    % 5. action
    \item \JsonKey{action} (integer or string): The single action chosen to execute.
    \begin{itemize}[leftmargin=1.2em, label=\textbullet, topsep=0.3em, itemsep=0.2em]
        \item \Highlight{CRITICAL RULE}: This value \textbf{MUST} be an exact element from the \texttt{action\_space} list provided in the input for the current step.
        \item The data type will be an \texttt{integer} for forward movement (from a list like \texttt{[1, 2, 3]}) or a \texttt{string} for a recovery maneuver (from a list like \texttt{[``turn\_left\_30", ``turn\_right\_90"]}).
    \end{itemize}
\end{enumerate}
    \end{tcolorbox}
    \vspace{-0.7em}
    \caption{\textbf{System Prompt Part 2: JSON Specification.}}
    \label{fig:prompt_part2}
\end{figure*}

\begin{figure*}[t]
    \centering
    \begin{tcolorbox}[
        enhanced,
        width=\linewidth,
        colback=promptbg,
        colframe=promptframe,
        arc=3mm,
        boxrule=1.2pt,
        left=12pt, right=12pt, top=12pt, bottom=12pt,
        fontupper=\sffamily\small,
        title={\large \bfseries Query Prompt Construction},
        coltitle=white,
        attach boxed title to top center={yshift=-2mm},
        boxed title style={colback=promptframe, rounded corners}
    ]
        The user prompt is dynamically constructed based on the agent's current status. It consists of an instruction header followed by a structured JSON object.

        \hfill

        \SectionTitle{1. Instruction Header}
        
        The prompt begins with one of the following text blocks depending on \JsonKey{status}:

        \begin{itemize}[leftmargin=*, itemsep=0.5em, topsep=0.3em]
            \item \textbf{If \texttt{NORMAL}}:
            \par "Process the following JSON data and the accompanying image to decide your next action."
            
            \item \textbf{If \texttt{GOAL\_LOST} or \texttt{STUCK}}:
            \par "Your goal is NOT visible. Analyze your memory and the scene to deduce the best action based on your last successful plan."
        \end{itemize}

        \vspace{0.8em}
        \hrulefill

        \SectionTitle{2. JSON Input Data Structure}
        
        The following JSON object is appended to the prompt inside a code block:

        \begin{tcolorbox}[colback=white, colframe=gray!30, sharp corners, leftrule=3pt, boxsep=0pt, top=4pt, bottom=4pt, left=4pt, right=4pt]
        \ttfamily
        \{ \\
        \hspace*{1.5em} \JsonKey{"step"}: \textcolor{strgray}{<curr\_iter>}, \\
        \hspace*{1.5em} \JsonKey{"status"}: \textcolor{strgray}{"NORMAL" | "GOAL\_LOST" | "STUCK"}, \\
        \hspace*{1.5em} \JsonKey{"global\_goal"}: \textcolor{strgray}{<goal\_description>}, \\
        \hspace*{1.5em} \JsonKey{"state"}: \{ \textcolor{strgray}{position, orientation\_yaw} \}, \\
        \hspace*{1.5em} \JsonKey{"action\_space"}: [\textcolor{strgray}{list of available actions}], \\
        \hspace*{1.5em} \JsonKey{"action\_guidance"}: \textcolor{alertred}{<See Logic Below>}, \\
        \hspace*{1.5em} \JsonKey{"memory"}: \textcolor{strgray}{<memory\_context (Optional)>}, \\
        \hspace*{1.5em} \JsonKey{"dynamic\_obstacles"}: \textcolor{strgray}{<dynamic\_obstacles\_info (Optional, only in NORMAL)>} \\
        \}
        \end{tcolorbox}

        \vspace{0.5em}
        \SubTitle{Logic for \texttt{action\_guidance} Field:}
        \begin{itemize}[leftmargin=*, itemsep=0.5em, topsep=0.3em]
            \item \textbf{Case \texttt{NORMAL}}: "The goal is marked with a green 'GOAL' box. Choose a numbered arrow that leads safely and directly towards it."
            \item \textbf{Case \texttt{GOAL\_LOST}}: "Your goal is NOT visible. Analyze your memory and the scene to deduce the best action based on your last successful plan."
            \item \textbf{Case \texttt{STUCK}}: "No forward paths are available. Choose a recovery action to reorient."
        \end{itemize}

    \end{tcolorbox}
    \caption{\textbf{User Query Prompt Template.}}
    \label{fig:user_prompt}
\end{figure*}

\begin{figure*}[t]
    \centering
    \begin{tcolorbox}[
        enhanced,
        width=\linewidth,
        colback=promptbg,
        colframe=promptframe,
        arc=3mm,
        boxrule=1.2pt,
        left=6pt, right=6pt, top=10pt, bottom=6pt,
        fontupper=\sffamily\small,
        title={\large \bfseries Social Navigation Prompt Extensions},
        coltitle=white,
        attach boxed title to top center={yshift=-2mm},
        boxed title style={colback=promptframe, rounded corners}
    ]
        \textit{For the social navigation benchmark, we append the following modules to the System Prompt to handle dynamic agents and ensure safety.}

        \vspace{0.5em}
        \hrulefill

        % --- Section: Dynamic Obstacle Handling ---
        \SectionTitle{Dynamic Obstacle Handling \textnormal{(New Section)}}
        
        \begin{enumerate}[leftmargin=*, itemsep=0.3em, topsep=0pt]
            \item \textbf{Observe Obstacles}: The scene may contain dynamic obstacles (e.g., humans) marked with \Highlight{red 'HUMAN' boxes}. These are also listed in the \texttt{dynamic\_obstacles} JSON field.
            \item \textbf{Assess Collision Risk}: You \Highlight{MUST} evaluate the collision risk for every action. Do not choose an arrow pointing directly at a nearby human.
            \item \textbf{Use ``stop\_and\_wait"}: If your path is blocked by a human and no safe alternative exists, or if moving causes imminent collision, you \Highlight{MUST} choose the \texttt{"stop\_and\_wait"} action.
        \end{enumerate}

        \vspace{0.4em}
        \hrulefill

        % --- Section: Protocol Updates ---
        \SectionTitle{Protocol Modifications \textnormal{(Excerpts)}}
        
        \begin{tcolorbox}[colback=white, colframe=gray!30, sharp corners, leftrule=3pt, boxsep=0pt, left=6pt, right=6pt, top=3pt, bottom=5pt]
            \SubTitle{Action Selection Logic}
            \vspace{0.3ex}
            \begin{itemize}[leftmargin=1.2em, label=\textbullet, itemsep=0.2em]
                \item \textbf{Normal/Goal\_Lost}: Choose the arrow that executes your plan. \Highlight{If all paths are blocked by a human, choose \texttt{stop\_and\_wait}.}
                \item \textbf{Stuck}: If stuck due to a human, the valid recovery plan is to \textbf{wait}.
            \end{itemize}
        \end{tcolorbox}

        \vspace{0.4em}
        \hrulefill

        % --- Section: JSON Spec Updates ---
        \SectionTitle{JSON Output Specification Updates}

        \begin{enumerate}[leftmargin=*, itemsep=0.4em, topsep=0.2em]
            \item \JsonKey{observation}: Description must include \Highlight{any dynamic obstacles} (e.g., ``a person is crossing from the left'').
            
            \item \JsonKey{thought}: You MUST explicitly justify safety.
            \begin{itemize}[leftmargin=1.2em, label=\textbullet, topsep=0.1em]
                \item \textit{Example}: ``Arrow 3 goes straight but is blocked by a pedestrian. Therefore, Arrow 4 is the safest choice.''
                \item \textit{Example}: ``Arrows 2 and 3 are both blocked by walking people. \Highlight{To avoid collision, I must stop and wait.}''
            \end{itemize}

            \item \JsonKey{action}: Extended action space.
            \begin{itemize}[leftmargin=1.2em, label=\textbullet, topsep=0.1em]
                \item Includes integers for movement and strings for recovery: \texttt{[``turn\_left'', \dots, \Highlight{``stop\_and\_wait''}]}.
            \end{itemize}
        \end{enumerate}

    \end{tcolorbox}
    % \vspace{-0.7em}
    \caption{\textbf{Social Navigation Prompt Extensions.} For environments with dynamic agents, we augment the system prompt with strict safety protocols, introducing a \texttt{stop\_and\_wait} action and requiring the VLM to explicitly reason about collision risks presented by humans detected in the scene (marked with red bounding boxes).}
    \label{fig:prompt_social}
\end{figure*}

\end{document}